\documentclass{article}

\usepackage{wrapfig}
\usepackage{amsmath}

\usepackage[preprint,nonatbib]{neurips_2021}
\usepackage[breaklinks=true,citecolor=blue,linkcolor=red]{hyperref}

\usepackage[utf8]{inputenc} 
\usepackage[T1]{fontenc}    

\usepackage{url}            
\usepackage{booktabs}       
\usepackage{amsfonts}       
\usepackage{nicefrac}       
\usepackage{microtype}      
\usepackage{xcolor}         
\usepackage{microtype}
\usepackage{graphicx}
\usepackage{booktabs} 
\usepackage{subcaption}

\usepackage{amssymb,amsmath, amsthm, latexsym}

\usepackage{enumitem}
\setlist[itemize]{topsep=-3pt}
\usepackage{amssymb,mathtools}

\usepackage{thmtools}

\usepackage{cleveref}

\newcommand{\A}{\mathbf{A}}

\newcommand{\x}{\mathbf{x}}
\newcommand{\y}{\mathbf{y}}

\newcommand{\p}{\mathbf{p}}

\newcommand{\X}{\mathbf{X}}

\newtheorem{remark}{Remark}

\newtheorem{prop}{Proposition}

\title{Rethinking positional encoding}

\author{%
  Jianqiao Zheng\thanks{The authors contributed equally.}\\
  \texttt{jianqiao.zheng@adelaide.edu.au} \\
  \And
  Sameera Ramasinghe$^{\ast}$\\
  \texttt{sameera.ramasinghe@adelaide.edu.au} \\
  \And
  Simon Lucey \\
  \texttt{simon.lucey@adelaide.edu.au} \\
   University of Adelaide\\
}

\begin{document}

\maketitle

\begin{abstract}
It is well noted that coordinate based MLPs benefit greatly -- in terms of preserving high-frequency information -- through the encoding of coordinate positions as an array of Fourier features. Hitherto, the rationale for the effectiveness of these \emph{positional encodings} has been solely studied through a Fourier lens. In this paper, we strive to broaden this understanding by showing that alternative non-Fourier embedding functions can indeed be used for positional encoding. Moreover, we show that their performance is entirely determined by a trade-off between the stable rank of the embedded matrix and the distance preservation between embedded coordinates. We further establish that the now ubiquitous Fourier feature mapping of position is a special case that fulfills these conditions.  Consequently, we present a more general theory to analyze positional encoding in terms of shifted basis functions. To this end, we develop the necessary theoretical formulae and empirically verify that our theoretical claims hold in practice. Codes available at \url{https://github.com/osiriszjq/Rethinking-positional-encoding}.
\end{abstract}

\section{Introduction}
Positional encoding is an umbrella term used for representing the coordinates of a structured object as a finite-dimensional embedding. Such embeddings are fast becoming critical instruments in modern language models \cite{gehring2017convolutional, vaswani2017attention, devlin2018bert, lewis2019bart, yang2019xlnet, bao2020unilmv2} and vision tasks that involve encoding a signal (e.g. 2D image, 3D object, etc.) as weights of a neural network \cite{mildenhall2020nerf, zhong2019reconstructing, park2020deformable,li2020neural, ost2020neural, gafni2020dynamic, martin2020nerf, barron2021mip}. Of specific interest in this paper is the use of positional encodings when being used to enhance the performance of \emph{coordinate-MLPs}. Coordinate-MLPs are fully connected networks, trained to learn the structure of an object as a continuous function, with coordinates as inputs. However, the major drawback of training coordinate-MLPs with raw input coordinates is their sub-optimal performance in learning high-frequency content \cite{rahaman2019spectral}. 

As a remedy, recent studies empirically confirmed that projecting the coordinates to a higher dimensional space using sine and cosine functions of different frequencies (\textit{i.e.}, Fourier frequency mapping) allows coordinate-MLPs to learn high-frequency information more effectively~\cite{mildenhall2020nerf, zhong2019reconstructing}. This observation was recently characterized theoretically by Tancik \textit{et al.}~\cite{tancik2020fourier}, showing that the above projection permits tuning the spectrum of the neural tangent kernel (NTK) of the corresponding MLP, thereby enabling the network to learn high-frequency information. Despite impressive empirical results, encoding position through Fourier frequency mapping entails some unenviable attributes. First, prior research substantiates the belief that the performance of the Fourier feature mapping is sensitive to the choice of frequencies. Leading methods for frequency selection, however, employ a stochastic strategy (i.e. random sampling) which can become volatile as one attempts to keep to a minimum the number of sampled frequencies. Second, viewing positional encoding solely through a Fourier lens obfuscates some of the fundamental principles behind its effectiveness. These concerns have heightened the need for an extended analysis of positional encoding.

This paper aims to overcome the aforesaid limitations by developing an alternative and more comprehensive understanding of positional encoding. The foremost benefit of our work is allowing non-Fourier embedding functions to be used in positional encoding. Specifically, we show that positional encoding can be accomplished via systematic sampling of shifted continuous basis functions, where the shifts are determined by the coordinate positions. In comparison to the ambiguous frequency sampling in Fourier feature mapping, we derive a more interpretable relationship between the sampling density and the behavior of the embedding scheme. In particular, we discover that the effectiveness of the proposed embedding scheme primarily relies on two factors: (i) the approximate matrix rank of the embedded representation across positions, and (ii) the distance preservation between the embedded coordinates. Distance preservation measures the extent to which the inner product between the shifted functions correlates with the Euclidean distance between the corresponding coordinates. Intuitively, a higher approximate matrix rank causes better memorization of the training data, while the distance preservation correlates with generalization. Remarkably, we establish that any given continuous function can be used for positional encoding -- as performance is simply determined by the trade-off between the aforementioned two factors. Further, we assert that the effectiveness and shortcomings of Fourier feature mapping can also be analyzed in the context of this newly developed framework. In summary, the contribution of this paper is three-fold:

\begin{itemize}
    \item We expand the current understanding of positional encoding and show that it can be formulated as a systematic sampling scheme of shifted continuous basis functions. Compared to the popular Fourier frequency mapping, our formulation is more interpretative in nature and less restrictive. 
    \item We develop theoretical formulae to show that the performance of the encoding is governed by the approximate rank of the embedding matrix (sampled at different positions) and the distance preservation between the embedded coordinates. We further solidify this new insight using empirical evaluations. 
    \item As a practical example, we employ a Gaussian signal as the embedding function and show that it can deliver on-par performance with the Fourier frequency mapping. Most importantly, we demonstrate that the Gaussian embedding is more efficient in terms of the embedding dimension while being less volatile. Impressive empirical reconstuction performance is obtained on both 1D and 2D signals using our proposed embedding function in conjunction with coordinate MLPs. 
\end{itemize}



\begin{figure}[ht]
    \centering
    \subfloat[\centering]{{\includegraphics[width=0.46\textwidth]{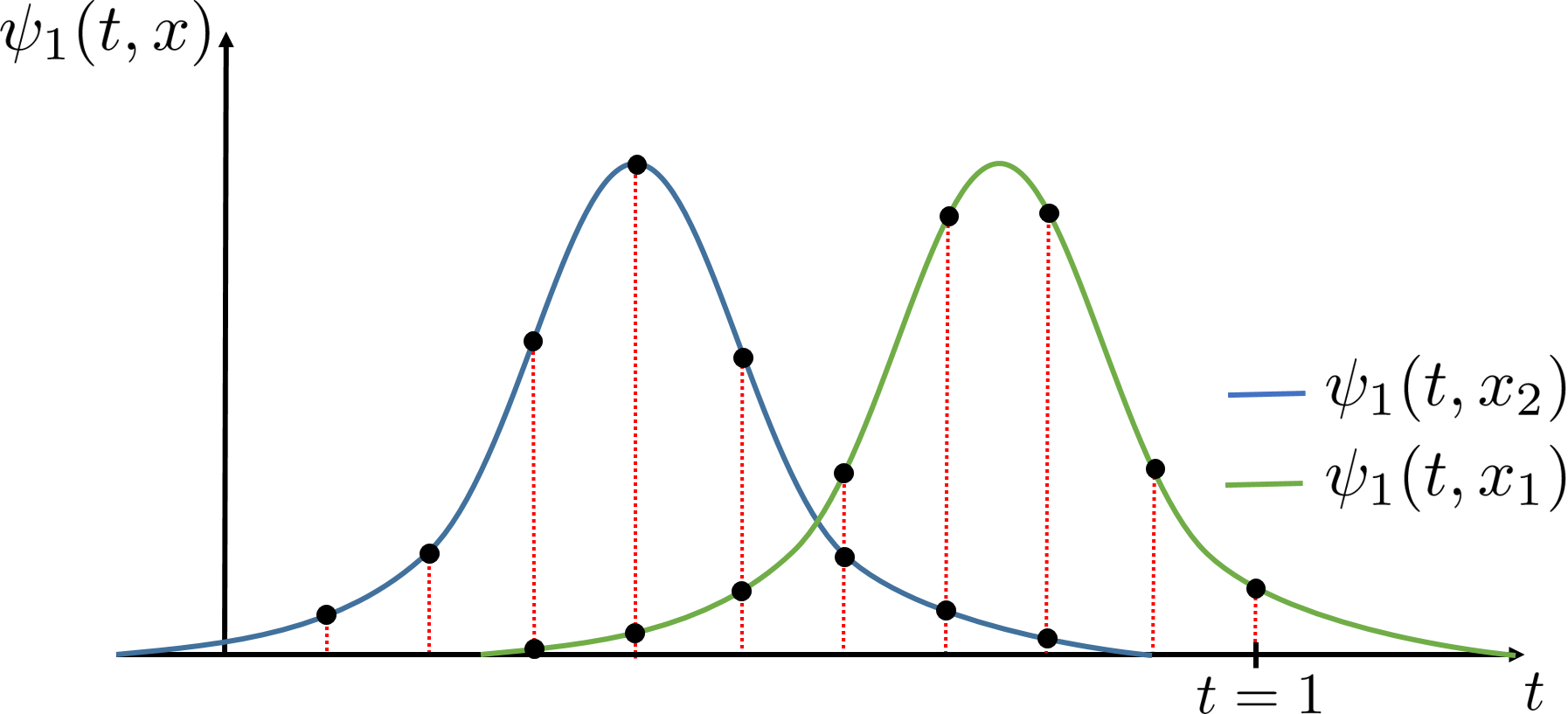} }}%
    \qquad
    \subfloat[\centering]{{\includegraphics[width=0.46\textwidth]{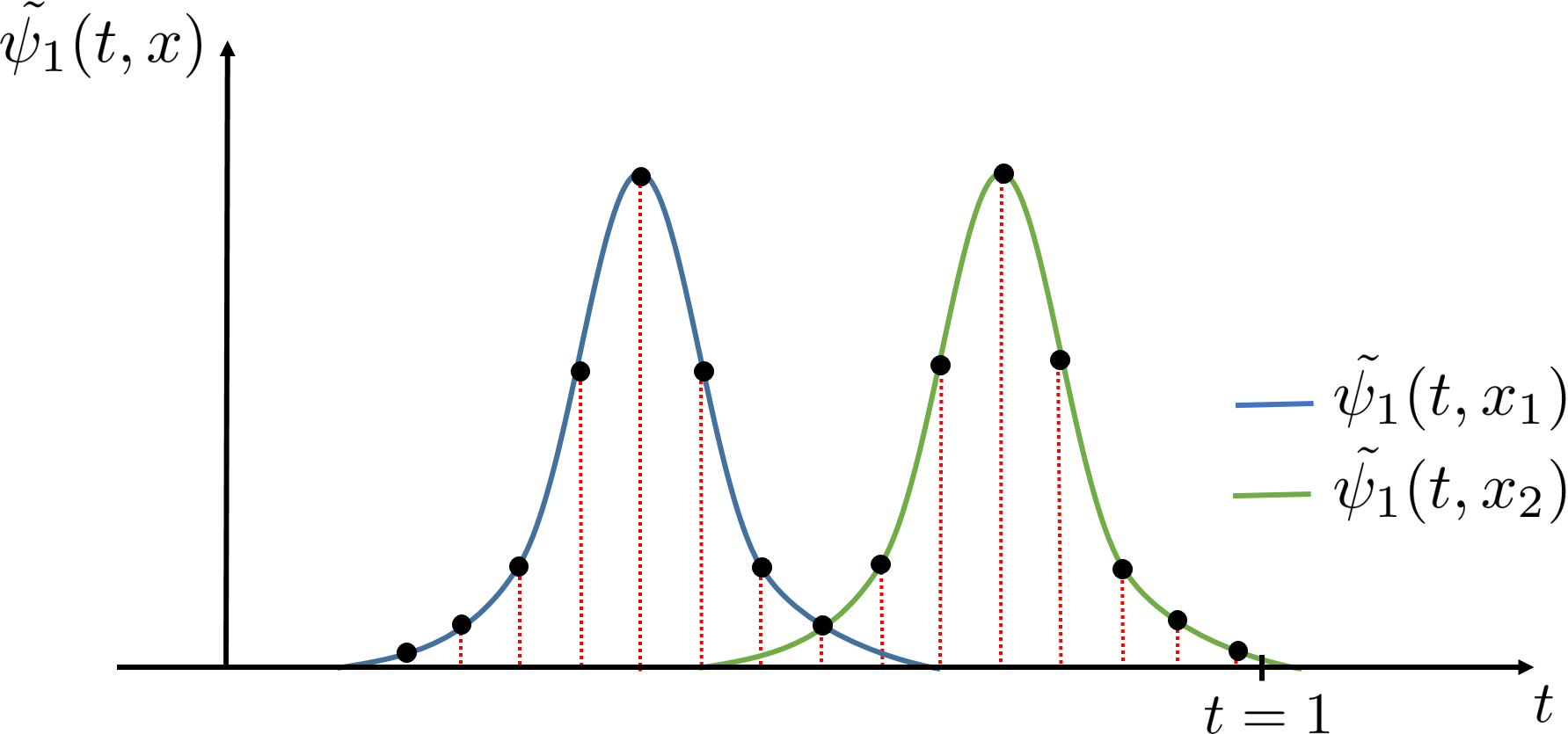} }}%
    \qquad
    \subfloat[\centering]{{\includegraphics[width=0.46\textwidth]{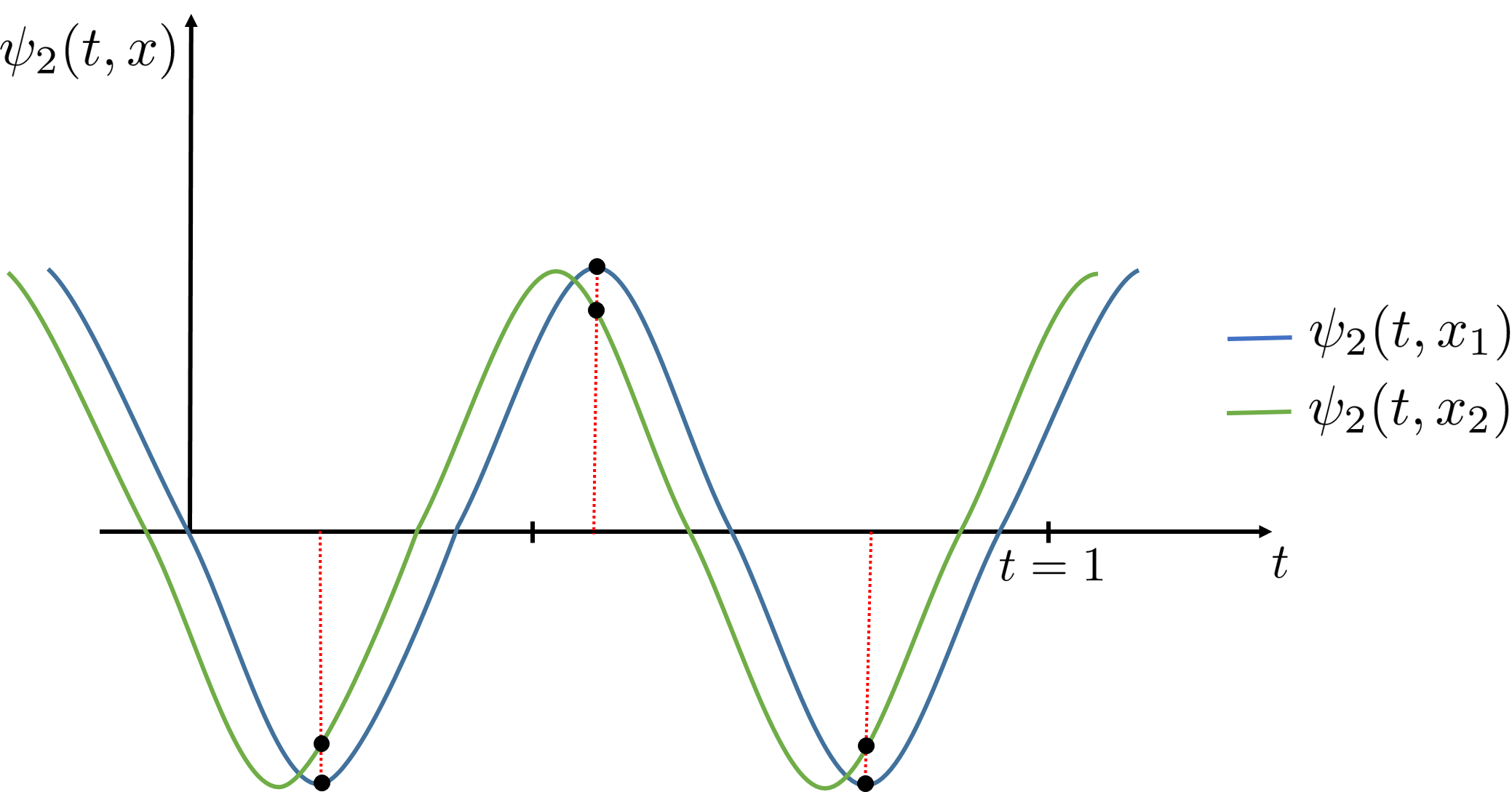} }}%
     \qquad
    \subfloat[\centering]{{\includegraphics[width=0.46\textwidth]{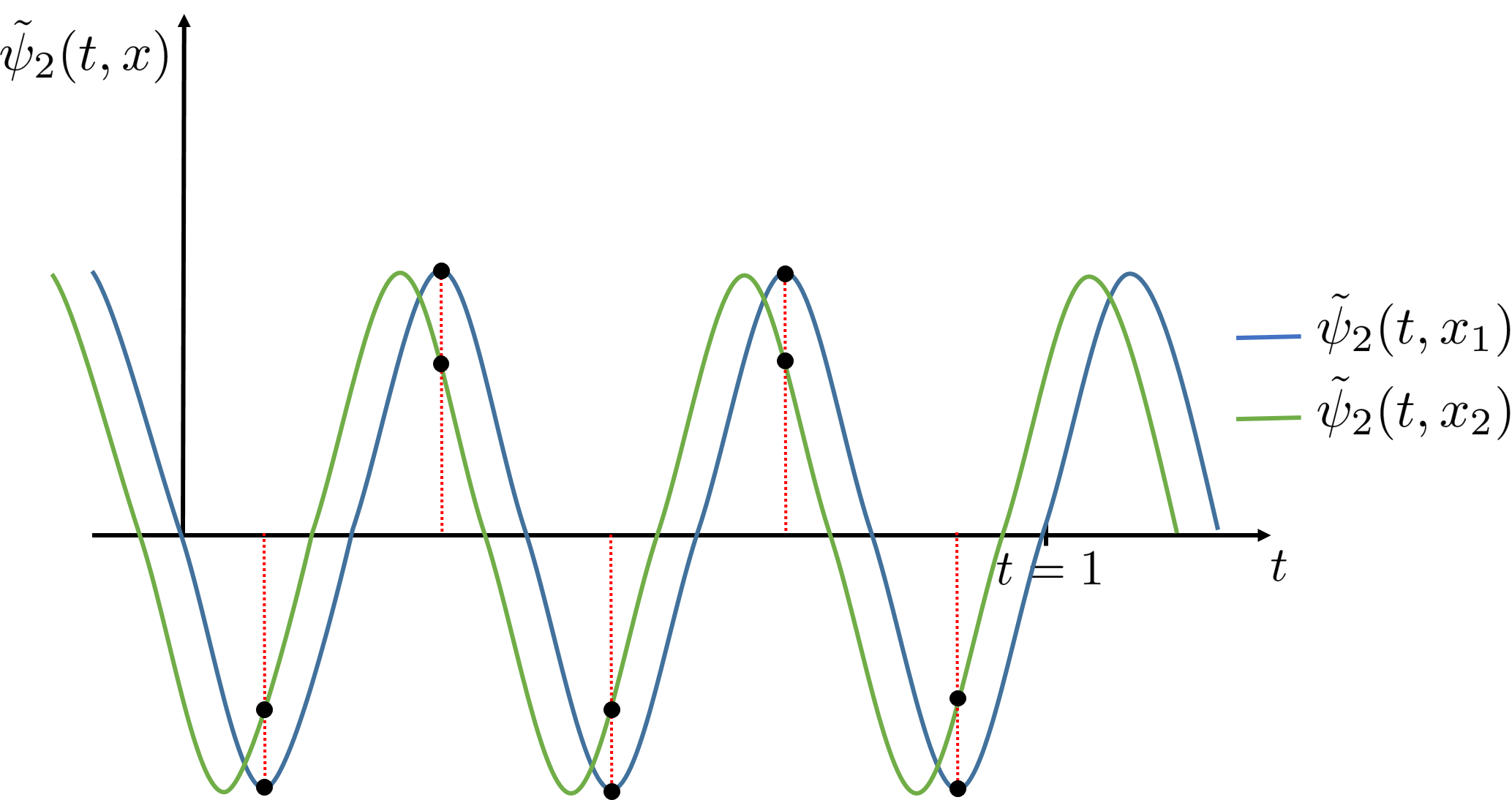} }}%
    \caption{Overview of the proposed positional encoding scheme. Positions are encoded as equidistant samples from shifted basis functions (embedders). The shifts are determined by the corresponding coordinate positions we are wanting to embed. In (a) and (b),  $x_1$ and $x_2$ are encoded as samples from shifted Gaussians with a higher and a lower standard deviation, respectively. Note that we need a higher number of samples for (b) due to higher bandwidth (see Sec.~\ref{sec:methodology}). In (c) and (d), $x_1$ and $x_2$ are encoded with sinusoidals with a different frequencies. Note that although different sampling rates are employed for (c) and (d), the same two values are repeated across the samples. Hence, sampling more than twice is redundant.}%
    \label{fig:example}%
\end{figure}

\section{Related works}
\label{sec:related_work}
Positional encoding became a popular topic among the machine learning community after the seminal work on Transformers by Viswani \textit{et al.} \cite{vaswani2017attention}  Since the attention mechanism used in the Transformers is position-insensitive, they employed a sinusoidal signal to encode the positions before feeding them to the higher blocks.  A contemporary work by Gehring \textit{et al.} \cite{gehring2017convolutional} also proposed a convolutional seq2seq model, adapting a positional encoding mechanism. Since then, using positional encoding in language models became a common trend \cite{shaw2018self, dai1901attentive, raffel2019exploring, he2020deberta, kitaev2018constituency}. Notably, Wang \textit{et al.} \cite{wang2019encoding} extended the embedding space from real numbers to complex values. Another critical aspect of their work is replacing the pre-defined encoding mechanism with a learnable one. There have also been other exciting attempts to improve positional encoding, such as extending the sequential positional encoding to tree-based positional encoding \cite{shiv2019novel}, untying the correlations between words and positions while embedding coordinates \cite{ke2020rethinking}, and modeling positional encoding using dynamical systems \cite{liu2020learning}. 

In parallel, positional encoding is also gaining attention in computer vision, specifically with coordinate-MLPs. Coordinate-MLPs provide an efficient method to encode objects such as images \cite{nguyen2015deep, stanley2007compositional}  and 3D scenes \cite{niemeyer2020differentiable, saito2019pifu, sitzmann2019scene} as their weights.  Remarkably, Mildenhall \textit{et al.} \cite{mildenhall2020nerf} and Zhong \textit{et al.} \cite{zhong2019reconstructing} found that encoding coordinates with sinusoidals allow coordinate-MLPs to learn high frequency content better.  One of the earliest roots of this approach can perhaps be traced to the work by Rahimi \textit{et al.} \cite{rahimi2007random}, where they used random Fourier features to approximate an arbitrary stationary kernel function by applying
Bochner’s theorem. More recently, Tancik \textit{et al.} \cite{tancik2020fourier}, leveraging the NTK theory \cite{arora2019fine, bietti2019inductive, du2018gradient, jacot2018neural,lee2019wide},  recently added theoretical rigor to this particular practice by showing that such embeddings enable tuning the spectrum of the NTK of the corresponding MLP. In contrast, the goal of this paper is to show that one does not have to be limited to the Fourier embedding for positional encoding. We demonstrate that alternative functions can be used for positional encoding while gaining similar or better performance compared to Fourier embedding. 

\section{Positional embedding: a theoretical walk-through}
\label{sec:methodology}

This section contains an exposition of the machinery and fundamentals necessary to understand the proposed framework. We
begin by considering a simple linear learner and show that, in theory, the capacity to memorize a given set of training data entirely depends on the approximate rank of the embedding matrix. Next, we establish that for generalization, the rank should be upper-bounded against the number of coordinates, \textit{i.e.}, the embedding function should be bandlimited. We incur a crucial insight here that positional encoding essentially portrays a trade-off between memorization and generalization. Afterward, we discuss the importance of distance preservation between embedded coordinates and its relationship to bandlimited embedding functions. Finally, we consider several possible embedder functions and analyze their behavior using the developed tools.

\subsection{Rank of the embedded representation}
\label{sec:rank}
Let $\x = [x_1, x_2, \cdots, x_N]^T$  be a vector of 1-D coordinates where $x_i \in [0,C]$ and $\y = [y_1, y_2, \cdots, y_n]^T$ be the corresponding outputs of a function $f:\mathbb{R} \to \mathbb{R}$. Our goal is to find a $d$ dimensional embedding $\Psi:\mathbb{R} \to \mathbb{R}^d$ for these positions, so that a linear model can be employed to learn the mapping $f$ as,

\begin{equation}
\label{equ:linear_learner}
    \textbf{w}^T\Psi(\x) + b \approx f(\cdot),
\end{equation}

where $\textbf{w} \in \mathbb{R}^{d}$ and $b \in \mathbb{R}$,   are the learnable weights and the bias, respectively. Then, considering the worst case scenario, where $\y$ is a random vector with no redundancy, it is straightforward to show that for perfect reconstruction of $\y$ using Eq.~\ref{equ:linear_learner} the following condition should be satisfied: 

\begin{equation}
\label{equ:rank_condition}
    \mathrm{Rank}\{[\Psi(x_1) \, \Psi(x_2) \, \dots \, \Psi(x_N) ]\}= N.
\end{equation}

Thus, we establish the following Proposition:
\begin{prop}
Consider a set of coordinates $\x = [x_1, x_2, \cdots, x_N]^T$, corresponding outputs $\y = [y_1, y_2, \cdots, y_N]^T$, and a $d$ dimensional embedding $\Psi:\mathbb{R} \to \mathbb{R}^d$. Assuming perfect convergence, the necessary and sufficient condition for a linear model to perfect memorize of the mapping between $\x$ and $\y$ is for $\X = [\Psi(x_1), \Psi(x_2), \dots, \Psi(x_N)]^T$ to have full rank.
\end{prop}

\subsection{Bandlimited embedders}

One possible way of enforcing the condition in Eq.~\ref{equ:rank_condition} is to define an embedding scheme where the rank of the embedded matrix strictly monotonically increases with $N$ (for a sufficiently large $d$). As depicted in Sec.~\ref{sec:rank}, this would ensure that the model can memorize the training data and therefore perfectly reconstruct $\y$. However, memorization alone does not yield a good model. On the contrary, we also need our model to be generalizable to unseen coordinates.

To this end, let us define elements of $\Psi(\cdot)$ as sampled values from a function $\psi:\mathbb{R}^2 \to \mathbb{R}$ such that for a given $x$,

\begin{equation}
\label{equ:embedding_matrix}
    \Psi(x) = [\psi(0,x), \psi(s,x), \dots, \psi((d-1)s,x)  ]^T,
\end{equation}

where $s = Cd^{-1}$ is the sampling interval. We shall refer to $\psi(\cdot)$ as the \emph{embedder}. As discussed above, for better generalization, we need, 
\begin{equation}
\label{equ:band_limited}
    \psi(t,x) \approx \sum_{b = 0}^{B} \alpha_{b} \beta_{b}(t)
\end{equation}
where~$\alpha$ and~$\beta(t)$ are linear weights and basis functions that can approximately estimate~$\psi(t,x)$ at any arbitrary position~$x$. We refer to such embedders as \emph{bandlimited embedders} with a bandwidth $B$. This is equivalent to saying that the embedding matrix has a bounded rank, i.e., the rank cannot increase arbitrarily with $N$.  The intuition here is that if $B$ is too small, the model wll demonstrate poor memorization and overly smooth generalization. On the other hand, if $B$ is extremely high, the model is capable of perfect memorization but poor generalization. Therefore we conclude that for ideal performance, the embedder should be chosen carefully, such that it is both bandlimited and has a sufficient rank. As we shall discuss the bandwidth~$B$ can also act as a guide for the minimal value of~$d$. 

\subsection{Distance preservation}

Intuitively, the embedded coordinates should preserve the distance between the original coordinates, irrespective of the absolute position. The embedded distance (or similarity) $\mathrm{D(\cdot,\cdot)}$ between two coordinates $(x_1,x_2)$ can be measured via the inner product $\mathrm{D}(x_1,x_2) = \int_{0}^{1} \psi(t,x_1)\psi(t,x_2)dt$. For ideal distance preservation we need,

\begin{equation}
   \lVert {x_1 - x_2} \rVert \propto \mathrm{D}(x_1,x_2).
\end{equation}

Interestingly, this property is also implicitly related to the limited bandwidth requirement. Note that in practice, we employ sampled embedders to construct $\Psi$ as shown in Eq.~\ref{equ:embedding_matrix}. Hence, the dot product between the sampled $\psi(t, x_1)$ and $\psi(t, x_2)$ should be able to approximate $\mathrm{D}$ as,

\begin{equation}
\label{equ:distance_preservation}
    \mathrm{D}(x_1,x_2) = \int_{0}^{C} \psi(t,x_1)\psi(t,x_2)dt \approx \sum_{d = 0}^{d-1}\psi(s\cdot d,x_1)\psi(s \cdot d,x_2),
\end{equation}

which is possible if, and only if, $\psi$ is bandlimited. In that case, $d = B$ is sufficient where $B$ is the bandwidth of $\psi$ (by Nyquist sampling theory). In practical implementations, we choose $C=1$.

\begin{remark}
The embedder should be bandlimited for better generalization (equivalently, the rank of the embedded matrix should be upper-bounded). Further, the ideal embedder should essentially face a trade-off between memorization and generalization. Here, memorization correlates with the rank of the embedded matrix, while generalization relates to the distance preservation between the embedded coordinates.
\end{remark}

\section{Analysis of possible embedders}
\label{sec:analysis}
Although our derivations in Sec.~\ref{sec:methodology} are generic, it is imperative to carefully choose a specific form of $\psi(\cdot, \cdot)$, such that properties of candidate embedders can be conveniently analyzed. Hence, we define embedders in terms of shifted basis functions, \textit{i.e.}, $\psi(t,x) = \psi(t-x)$. Such a definition permits us to examine embedders in a unified manner, as we shall see below.

Moreover, the rank of a matrix can be extremely noisy in practice. Typically, we need to heuristically set an appropriate threshold to the singular values, leading to  unstable calculations. Therefore, following \cite{sanyal2019stable}, we use the stable rank instead of the rank in all our experiments. In particular, the stable rank is a more stable surrogate for the rank, and is defined as  $\frac{\|\A\|_{F}^2}{\|\A\|_{2}^2}$, where $\A$ is the matrix, $\| \cdot \|_F$ is the Frobenius norm, and $\| \cdot \|_2$ is the matrix norm. From here onwards, we will use the terms rank, approximate rank, and stable rank interchangeably.

\textbf{Impulse embedder.} One simple way to satisfy the condition~\ref{equ:rank_condition} for an arbitrary large $N$ is to define $\psi(t,x) = \delta (t-x)$, where $\delta(\cdot)$ is the impulse function. Note that using an impulse embedder essentially converts the embedding matrix to a set of one-hot encodings. With the impulse embedder, we can perfectly memorize a given set of data points, as then the embedded matrix has full rank. The obvious drawback, however, is that the bandwidth of the impulse embedder is infinite, \emph{i.e.}, assuming a continuous domain, $d$ needs to reach infinity to learn outputs for all possible positions. Hence, the distance preservation is hampered, and consequently, the learned model lacks generalization.

\textbf{Sine embedder.} Consider $\psi(t,x) = \mathrm{sin}(f(t-x))$ for an arbitrary fixed $f$. Since $\mathrm{sin}(f(t-x)) = \mathrm{sin}( ft)\mathrm{cos}(fx) - \mathrm{cos}(ft)\mathrm{sin}(fx)$, elements of any row of the embedding matrix can be written as a linear combination of the corresponding $\mathrm{sin}(ft)$ and $\mathrm{cos}(ft)$. Thus, the rank of the embedding matrix is upper-bounded at $2$. Consequently, the expressiveness of the encoding is limited, leading to poor memorization and overly smooth generalization (interpolation) at unseen coordinates.

\begin{figure}[ht]
    \centering
    \subfloat[\centering]{\includegraphics[width=0.46\textwidth]{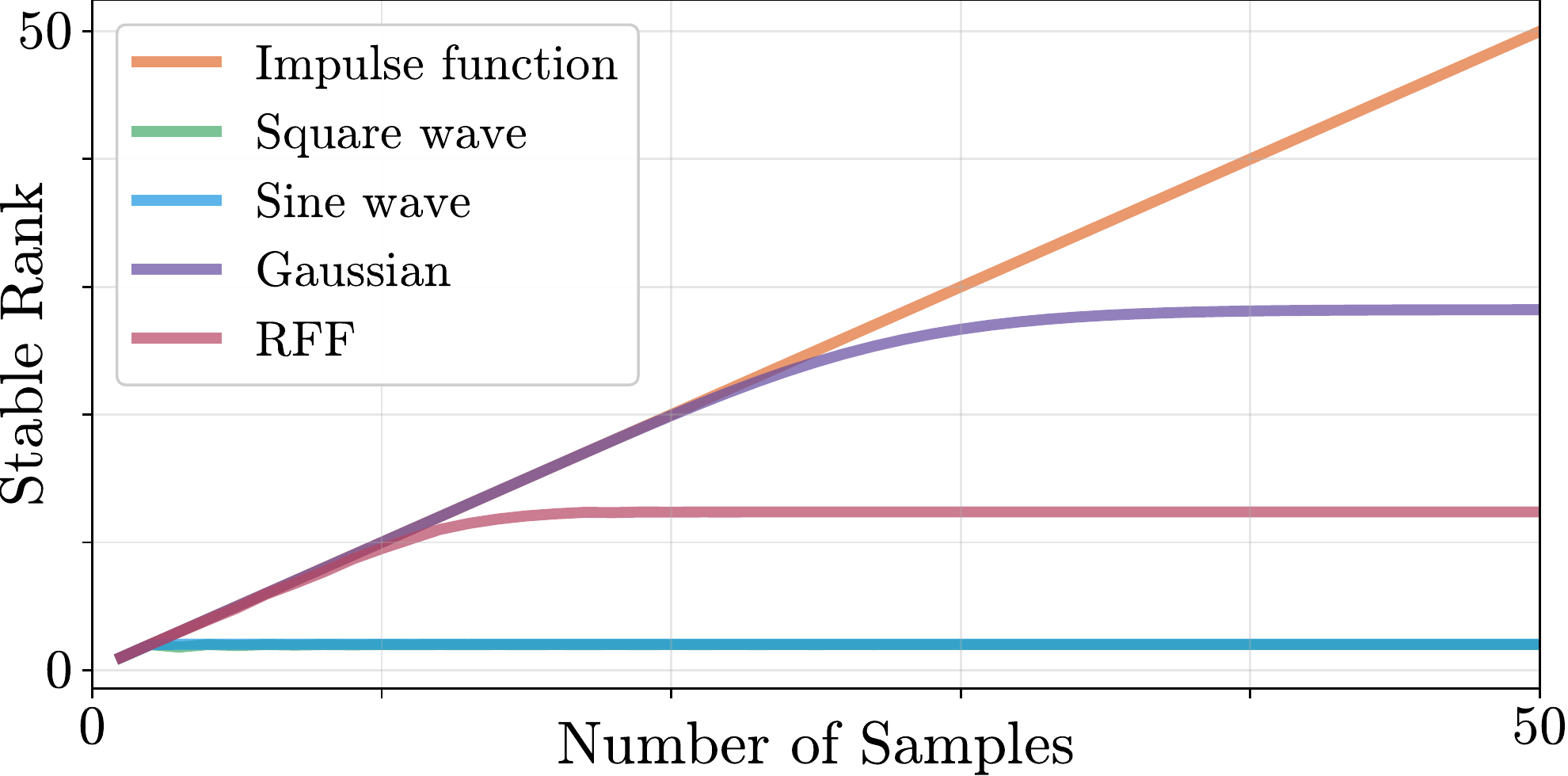}}%
    \qquad
    \subfloat[\centering]{\includegraphics[width=0.46\textwidth]{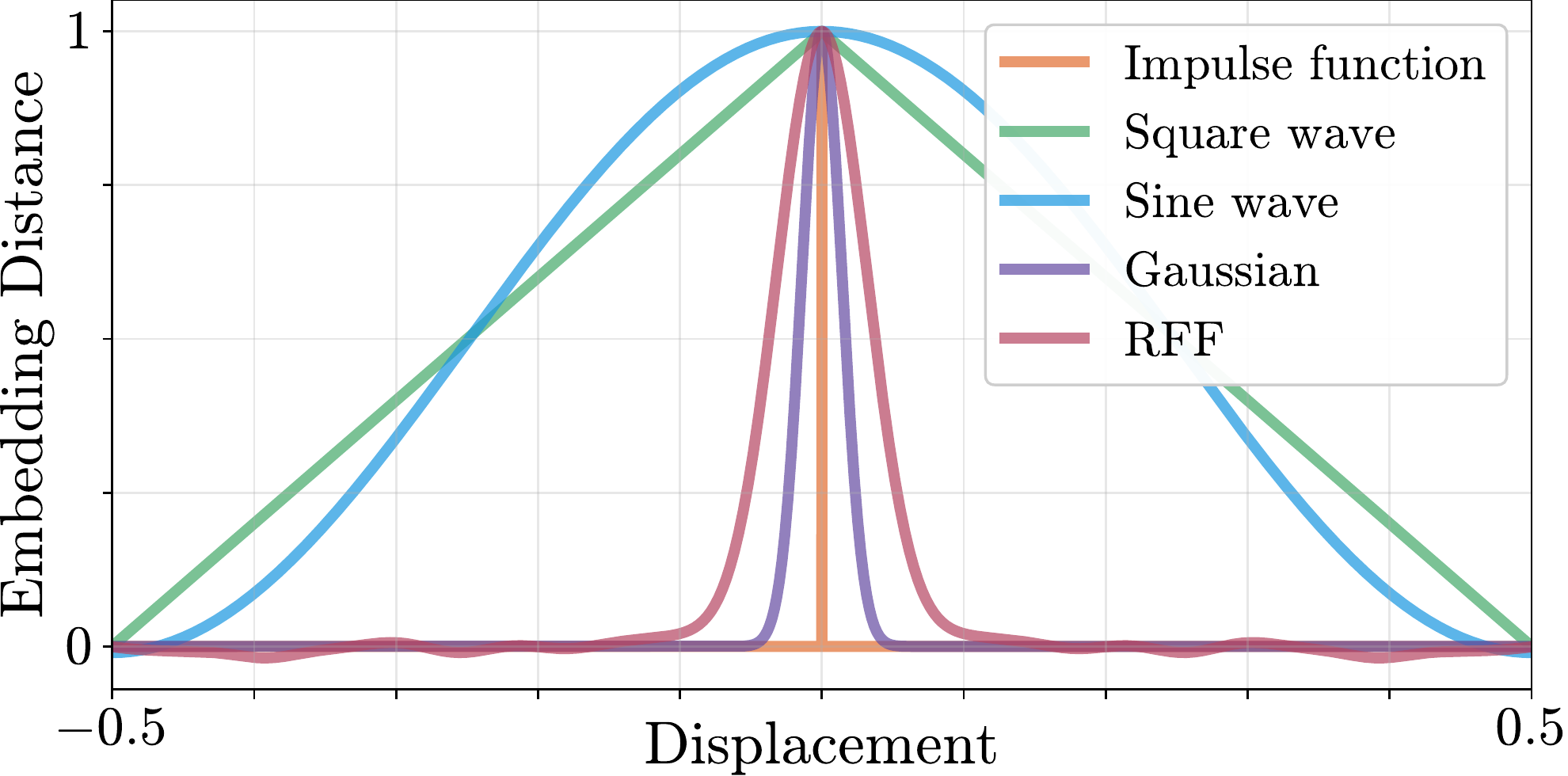}}%
    \caption{Quantitative comparison of  (a) the stable rank and (b) the distance preservation of different embedders and random Fourier features. As expected, the stable rank of the impulse embedder strictly increases with the number of sampled points, causing poor distance preservation. The stable rank of the sine embedder is upper-bounded at $2$. Note that as predicted in theory, the stable ranks of the square embedder and the sine embedder almost overlap. However, if the sample numbers are extremely high (not shown in the figure), their stable ranks begin to deviate. Similarly, the square embedder demonstrates perfect distance preservation, and the sine embedder is a close competitor. In contrast, the Gaussian embedder and the RFF showcase mid-range upper bounds for the stable rank and adequate distance preservation, advocating a much better trade-off between memorization and generalization.  }%
    \label{fig:embeddercomparison}%
\end{figure}

\textbf{Square embedder.} Let us denote a square wave with unit amplitude and period $2\pi$ as $\mathrm{sgn}(\mathrm{sin}(t))$, where $\mathrm{sgn}$ is the sign function. Then, define $\psi(t,x) = \mathrm{sgn}(\mathrm{sin}(t-x))$. It is easy to deduce that the embedded distance $D(x_1, x_2) = 1 - 2\lVert x_1 - x_2 \rVert, \forall |x| \leq 1$ which implies perfect distance preservation. The drawback, however, is that the square wave is not bandlimited. Thus, it cannot approximate the inner product $\int \psi(t,x)\psi(t,x')$ using a finite set of samples  as in Eq.~\ref{equ:distance_preservation}. However, an interesting attribute of the square wave is that it can be decomposed into a series of sine waves with odd-integer harmonic frequencies as $\mathrm{sgn}(\mathrm{sin}(t)) = \frac{4}{\pi} \big[ \mathrm{sin}(t) + \frac{1}{3}\mathrm{sin}(3t) + \frac{1}{5}\mathrm{sin}(5t) + \frac{1}{7}\mathrm{sin}(7t) + \dots \big]$. In other words, its highest energy (from a signal processing perspective) is contained in a sinusoidal with the same frequency. Thus, the square wave can be \emph{almost} approximated by a sinusoidal signal. In fact, the square wave and the sinusoidal demonstrates similar properties in terms of the stable rank and the distance preservation (see Fig.~\ref{fig:embeddercomparison}).

\begin{wrapfigure}[]{R}{0.5\textwidth}
\centering
\includegraphics[width=0.5\textwidth]{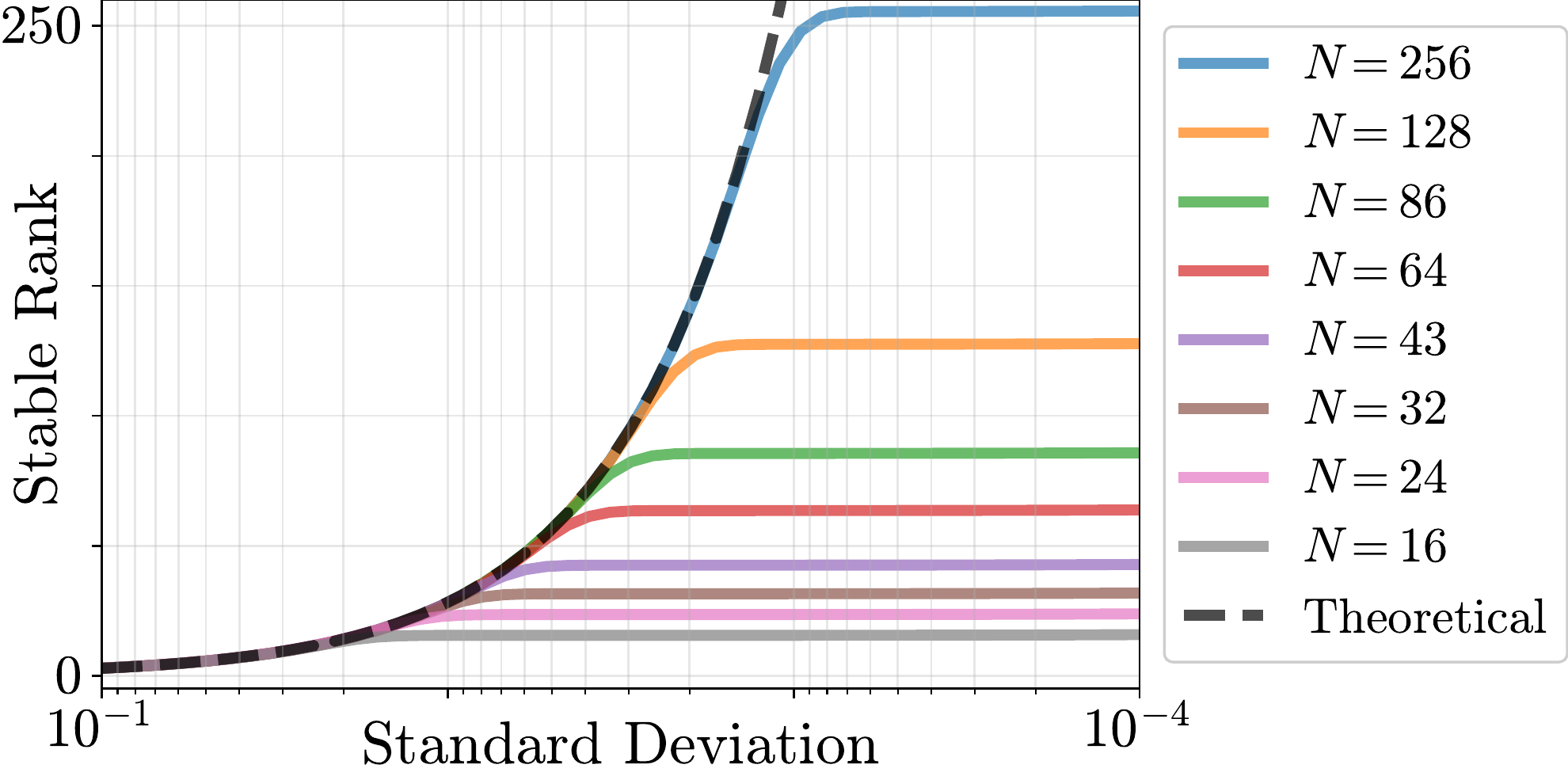}
\caption{\label{fig:gaussian_sr}Stable rank of the Gaussian embedder against the standard deviation for different number of samples. The dash line is the theoretical stable rank $\frac{1}{2\sqrt{\pi}\sigma}$}
\end{wrapfigure}

\textbf{Gaussian embedder.} We define the Gaussian embedder as  $\psi(t,x) = \exp (-\frac{\|t-x\|^2}{2\sigma^2})$ where $\sigma$ is the standard deviation. The Gaussian embedder is also approximately bandlimited like the square embedder. However, the Gaussian embedder has a higher upper bound for the stable rank that can be controlled by $\sigma$. More precisely, when the embedding dimension is large enough, the stable rank of the Gaussian embedding matrix and the embedded distance between coordinates can be obtained analytically as shown below.

\begin{prop}
\label{prop:gaussian}
Let the Gaussian embedder be denoted as $\psi(t,x) = \exp (-\frac{\|t-x\|^2}{2\sigma^2})$. With a sufficient embedding dimension, the stable rank of the embedding matrix obtained using the Gaussian embedder is $\min(N, \frac{1}{2\sqrt{\pi}\sigma})$ where $N$ is the number of embedded coordinates. Under the same conditions, the embedded distance between two coordinates $x_1$ and $x_2$ is $\mathrm{D}(x_1,x_2) = \exp (-\frac{\|x_1-x_2\|^2}{4\sigma^2})$.
\end{prop}

 (see Fig.~\ref{fig:gaussian_sr} for an experimental illustration). It is clear from Proposition \ref{prop:gaussian} that as the number of sampled positions goes up, the stable rank of the Gaussian embedding matrix will linearly increase until it reaches its upper bound. 
 Finally, Fig.~\ref{fig:embeddercomparison} empirically validates the theoretically discussed properties of different embedders.


\subsection{Connection to the Random Fourier Features}
\label{sec:connectiontorff}
The prominent way of employing Fourier frequency mapping is via Random Fourier Features (RFF) mapping, where the frequencies are randomly sampled from a Gaussian distribution with a certain standard deviation $\sigma$. In this Section, we show that RFF mapping can be analyzed through the lens of our theoretical framework discussed thus far. To this end, we first establish the following proposition:

\begin{prop}
\label{prop:rff}
Let the RFF embedding be denoted as $\gamma(x)=[\cos(2\pi\mathbf{b}x),\sin(2\pi\mathbf{b}x)]$, where $\mathbf{b}$ are sampled from a Gaussian distribution. When the embedding dimension is large enough, the stable rank of RFF will be $\min(N, \sqrt{2\pi}\sigma)$, where $N$ is the numnber of embedded coordinates. Under the same conditions, the embedded distance between two coordinates $x_1$ and $x_2$ is $\mathrm{D}(x_1,x_2) = \sum_j \cos{2\pi b_j(x_1-x_2)}$.
\end{prop}

\begin{figure}[ht]
\captionsetup{size=small}
\centering 
\begin{subfigure}{0.015\textwidth}\hspace{-2mm}
\rotatebox[origin=c]{90}{\tiny{Gaussian embedder}}
\end{subfigure}
\begin{subfigure}{0.45\textwidth}
\includegraphics[width=\linewidth]{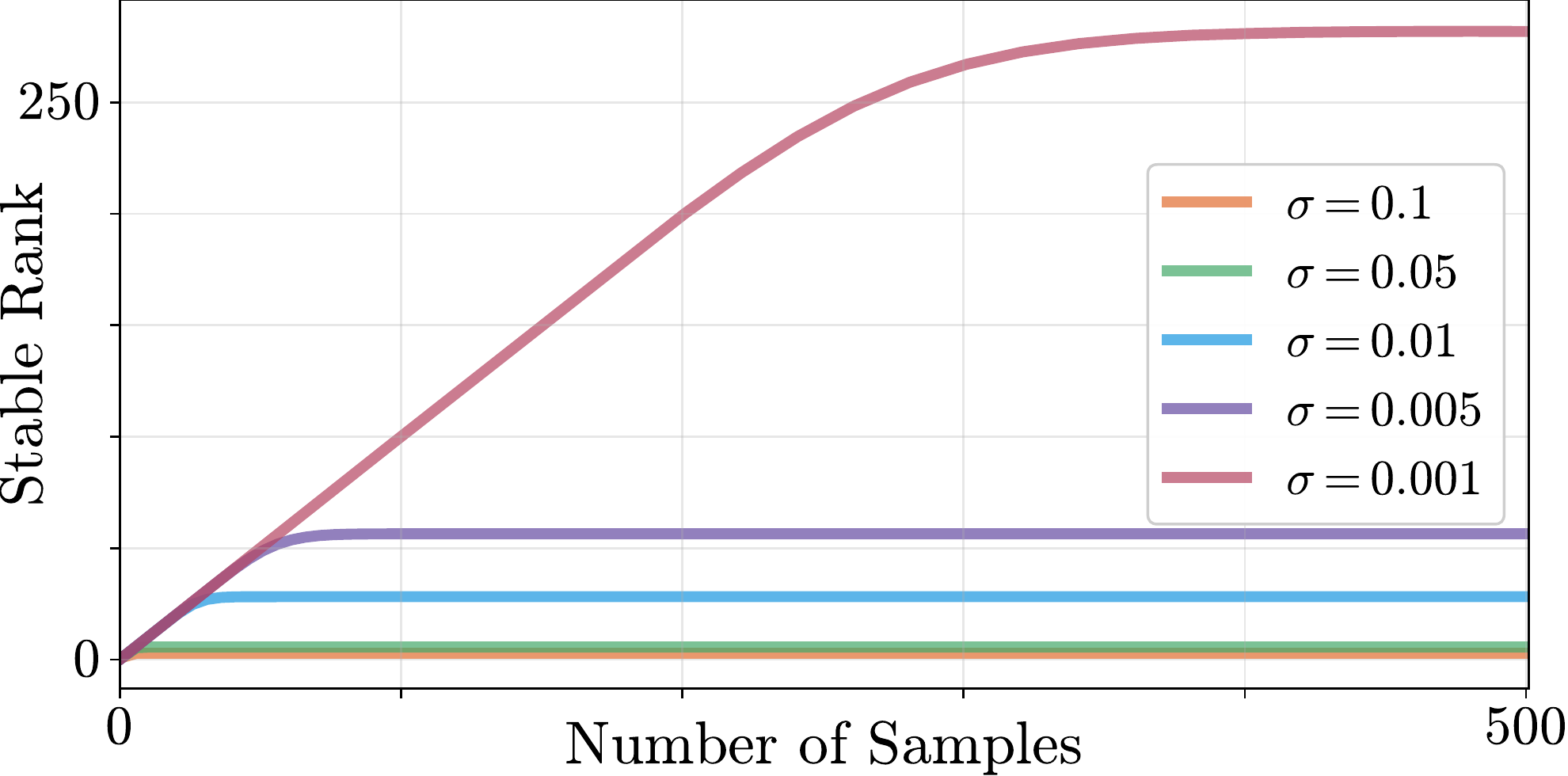} 
\end{subfigure} 
\begin{subfigure}{0.45\textwidth}
\includegraphics[width=\linewidth]{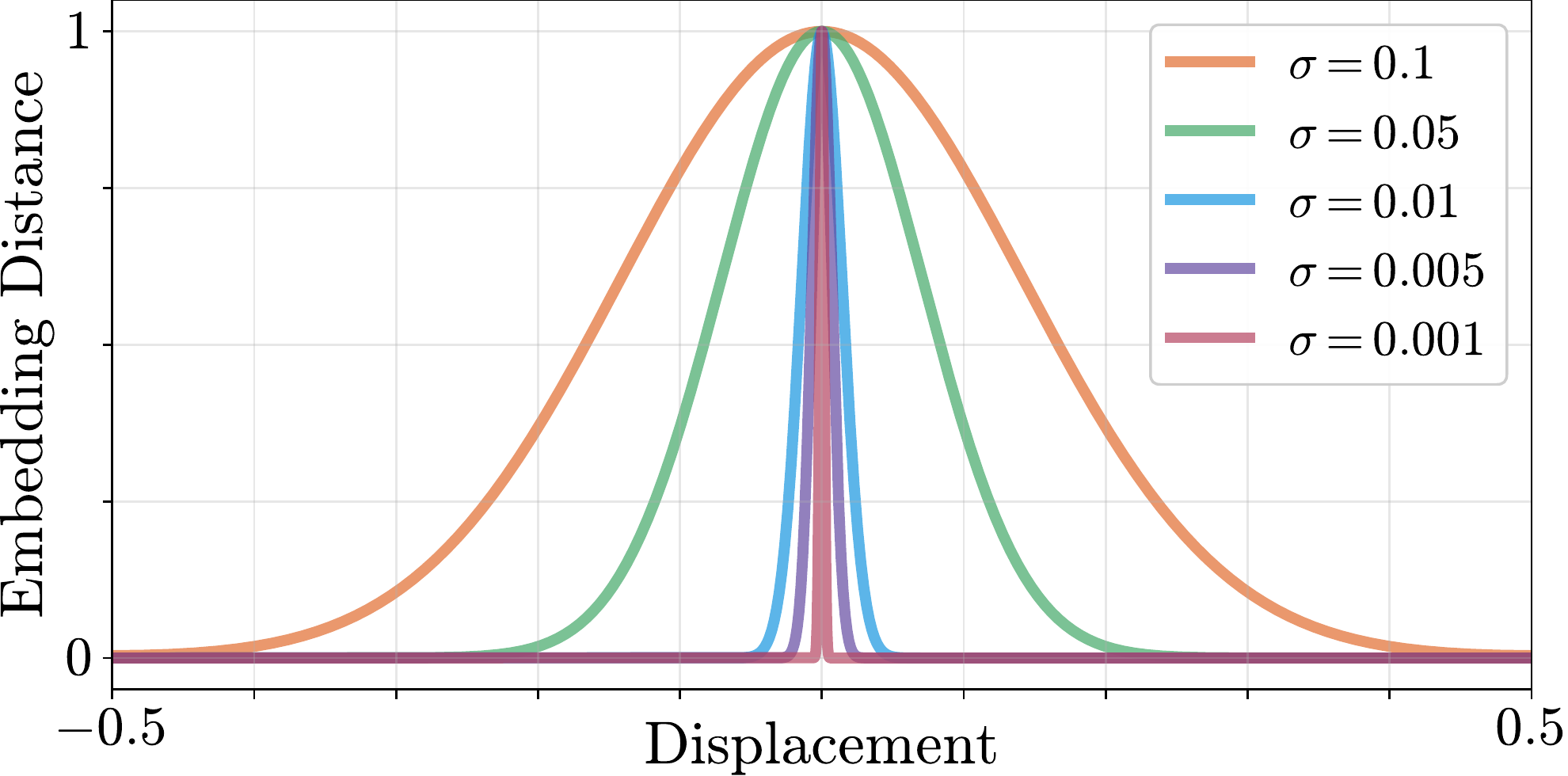} 
\end{subfigure}

\begin{subfigure}{0.015\textwidth}\hspace{-2mm}
\rotatebox[origin=c]{90}{\tiny{RFF}}
\end{subfigure}
\begin{subfigure}{0.45\textwidth}
\includegraphics[width=\linewidth]{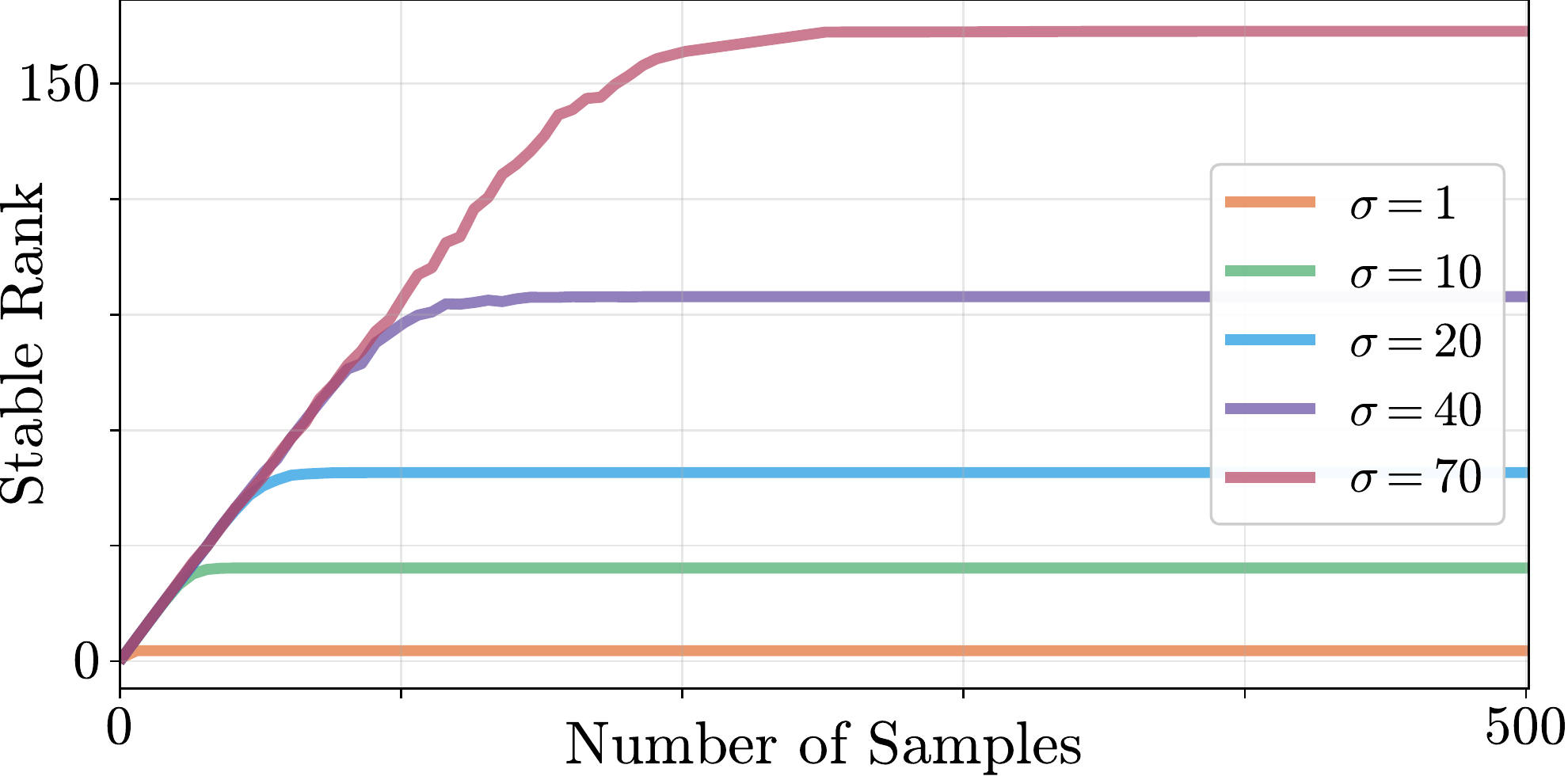} 
\caption{Stable rank vs number of samples}
\end{subfigure} 
\begin{subfigure}{0.45\textwidth}
\includegraphics[width=\linewidth]{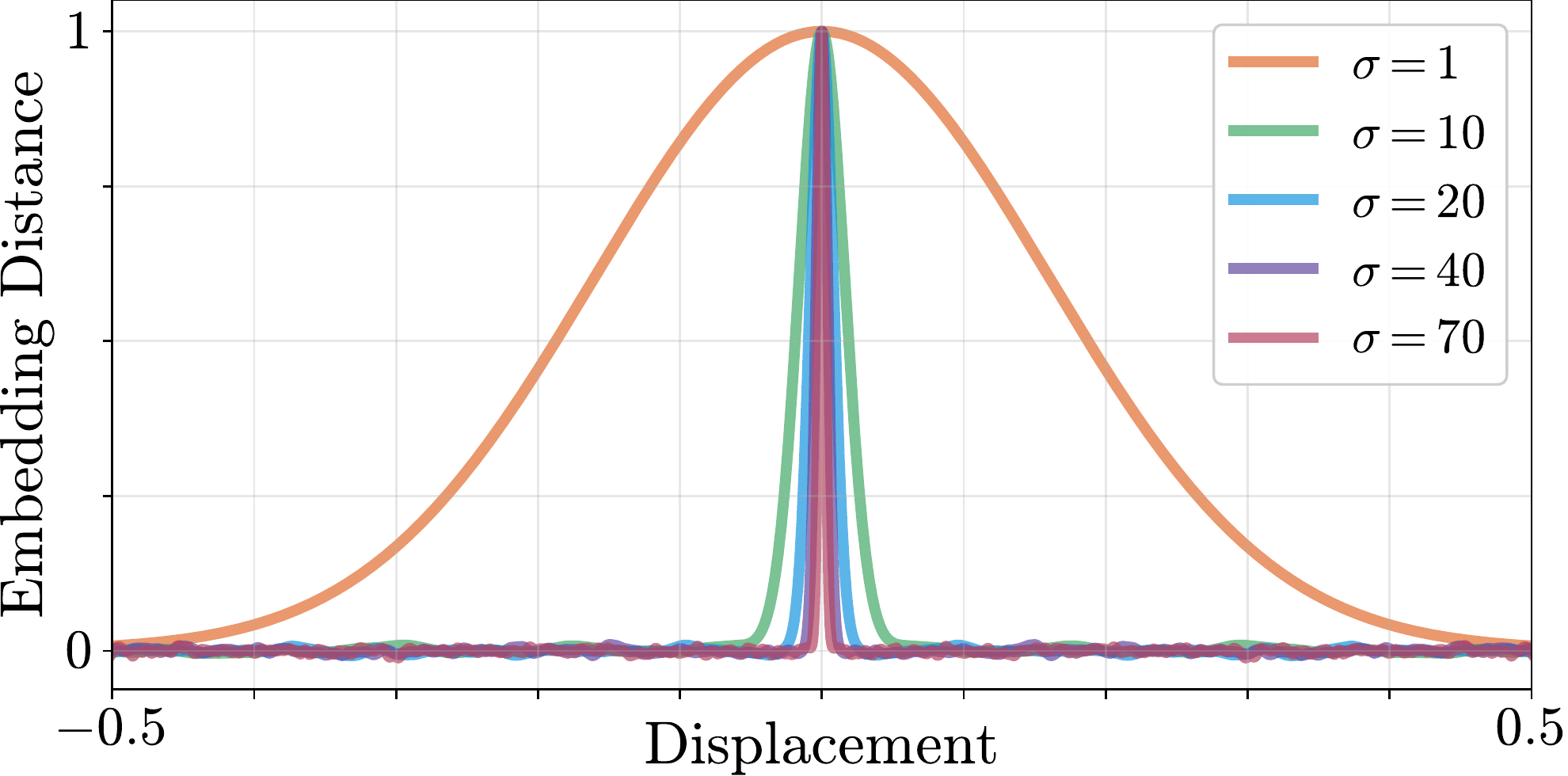} 
\caption{Distance preservation}
\end{subfigure} 
\caption{The stable rank and distance preservation of the Gaussian embedder and RFF across different standard deviations.}
\label{fig:Gauss_vs_RFF}
\end{figure}

As shown in Fig.~\ref{fig:Gauss_vs_RFF}, the stable rank of RFF inceases linearly with the number of samples until it gets saturated at $\sqrt{2\pi}\sigma$. This gives us a relationship between RFF and the Gaussian embedder: Let $\sigma_g$ and $\sigma_f$ be the standard deviations of the Gaussian embedder and RFF, respectively. When their stable ranks are equal, $\frac{1}{2\sqrt{\pi}\sigma_g}=\sqrt{2\pi}\sigma_f$ (from Proposition \ref{prop:gaussian} and \ref{prop:rff}). This implies that when $\sigma_g\sigma_f=\frac{1}{2\sqrt{2}\pi}$, these two embeders are equivalent in terms of the stable rank and distance preservation (observe Fig.~\ref{fig:Gauss_vs_RFF} when $\sigma_g=0.01$ and $\sigma_f=0.1$).

Also, a common observation with RFFs is that when $\sigma_f$ is too low, the reconstruction results are overly smooth and if $\sigma_f$ is too high, it gives noisy interpolations \cite{tancik2020fourier}. This observation directly correlates with our theory. See in Fig.~\ref{fig:Gauss_vs_RFF} that when the standard deviation increases, the stable rank increases and distance preservation decreases. Similarly, When the standard deviation is too low, the stable rank decreases while distance preservation increases.

\begin{figure}[ht]
    \centering
    \subfloat{\includegraphics[width=0.46\textwidth]{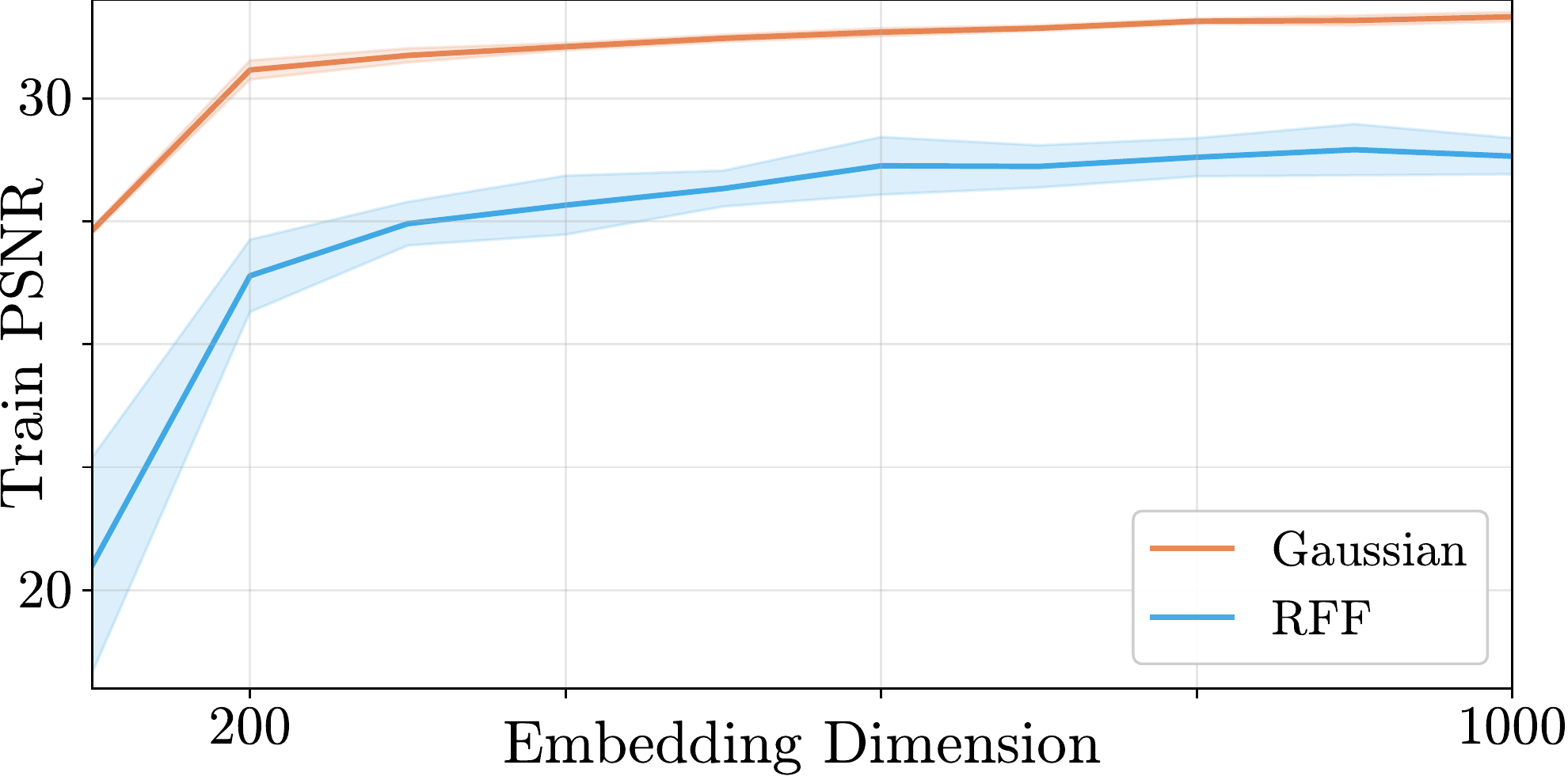}}%
    \qquad
    \subfloat{\includegraphics[width=0.46\textwidth]{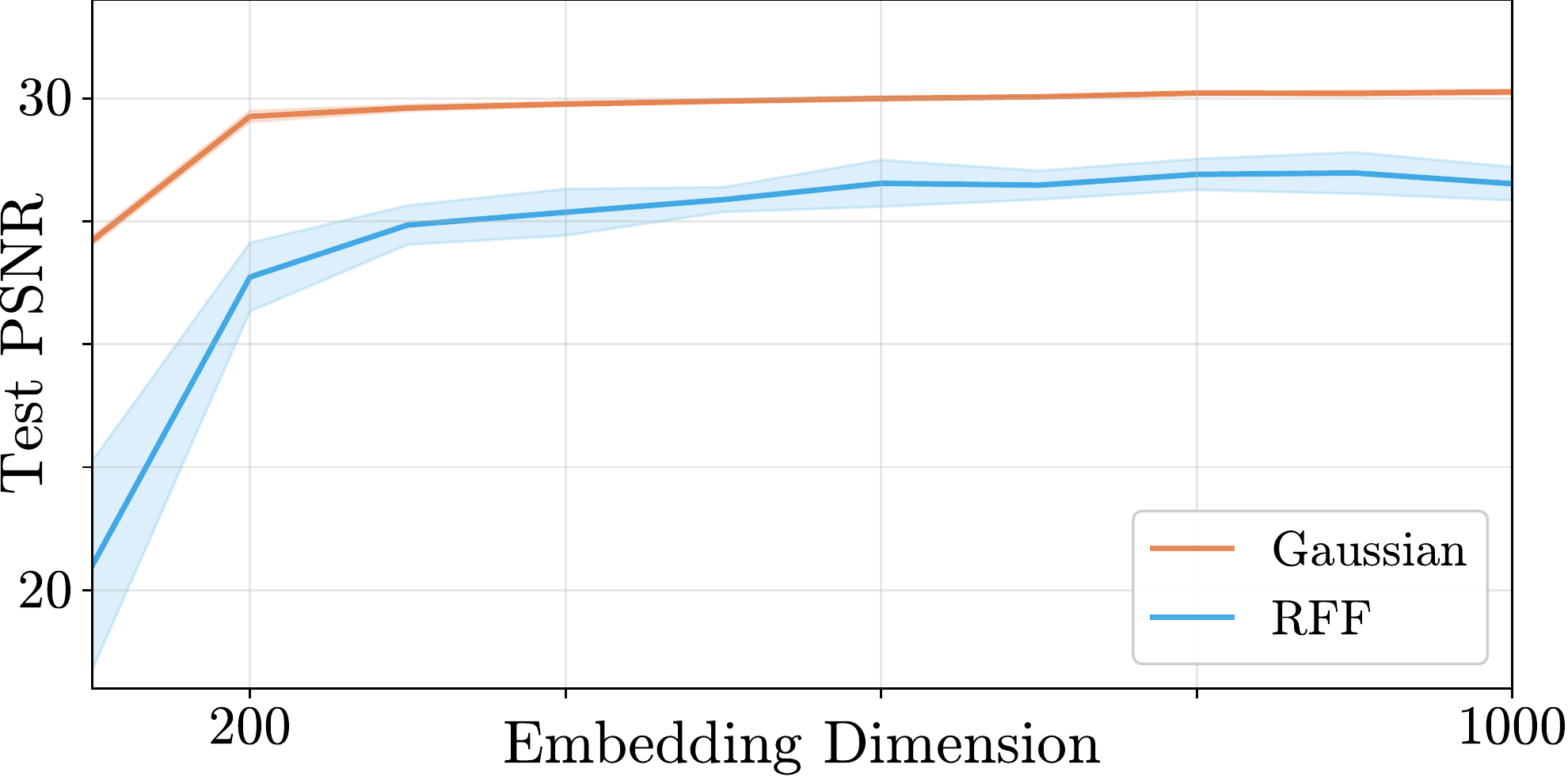}}%
    \caption{Stability of the performance at different embedding dimensions when encoding a 1-D signal. Shaded areas correspond to two standard deviations across $10$ experiments.  As illustrated, the Gaussian embedder demonstrates much stable performance, especially at lower embedding dimensions. }%
    \label{fig:G_vs_RFF}%
\end{figure}

\section{Experiments}

In this Section, we empirically confirm the advantages of using the proposed embedding procedure and verify that the theoretically predicted properties in the previous Sections hold in practice.

\subsection{The Gaussian embedder vs RFF}
The proposed positional embedding scheme is a deterministic process, \textit{i.e.}, there is no ambiguity in performance against hyper-parameters. In comparison, RFF samples frequencies randomly, leading to unstable performance, especially when the embedding dimension is low. In order to verify this, we use a 1D signal (a stripe from the popular Pepper image) as the target signal and train a linear network with a single layer with both embeddings. Following Sec.~\ref{sec:connectiontorff}, we choose the standard deviations $\sigma_g=0.005$ and $\sigma_f=\frac{1}{2\sqrt{2}\pi\sigma_g}$. We run the experiment $10$ times with random initializations and obtain the means and the standard deviations for different embedding dimensions. As reported in Fig.~\ref{fig:G_vs_RFF}, the variance of performance in the Gaussian embedder is much smaller compared to RFF. Note that the performance variance of the RFF decreases as the embedding dimension increases. In particular, this means that the ambiguity of performance in RFF can be reduced with a sufficiently large embedding dimension. However, at lower embedding dimensions, the Gaussian embedder demonstrates less ambiguity and better performance.


\subsection{Performance of different  embedders}
We empirically compare the performance of different embedders and verify that the theoretically predicted properties hold in practice. We use a single-layer linear MLP for this experiment and choose the embedding dimension to be $10000$. Further, we pick ten random rows from the Pepper image as our targets and measure the average performance.  We train the model with each embedder for $4000$ epochs using the Adam optimizer with a learning rate of $1e^{-4}$. 

\begin{figure}[ht]
    \centering
    \subfloat{\includegraphics[width=0.4\textwidth]{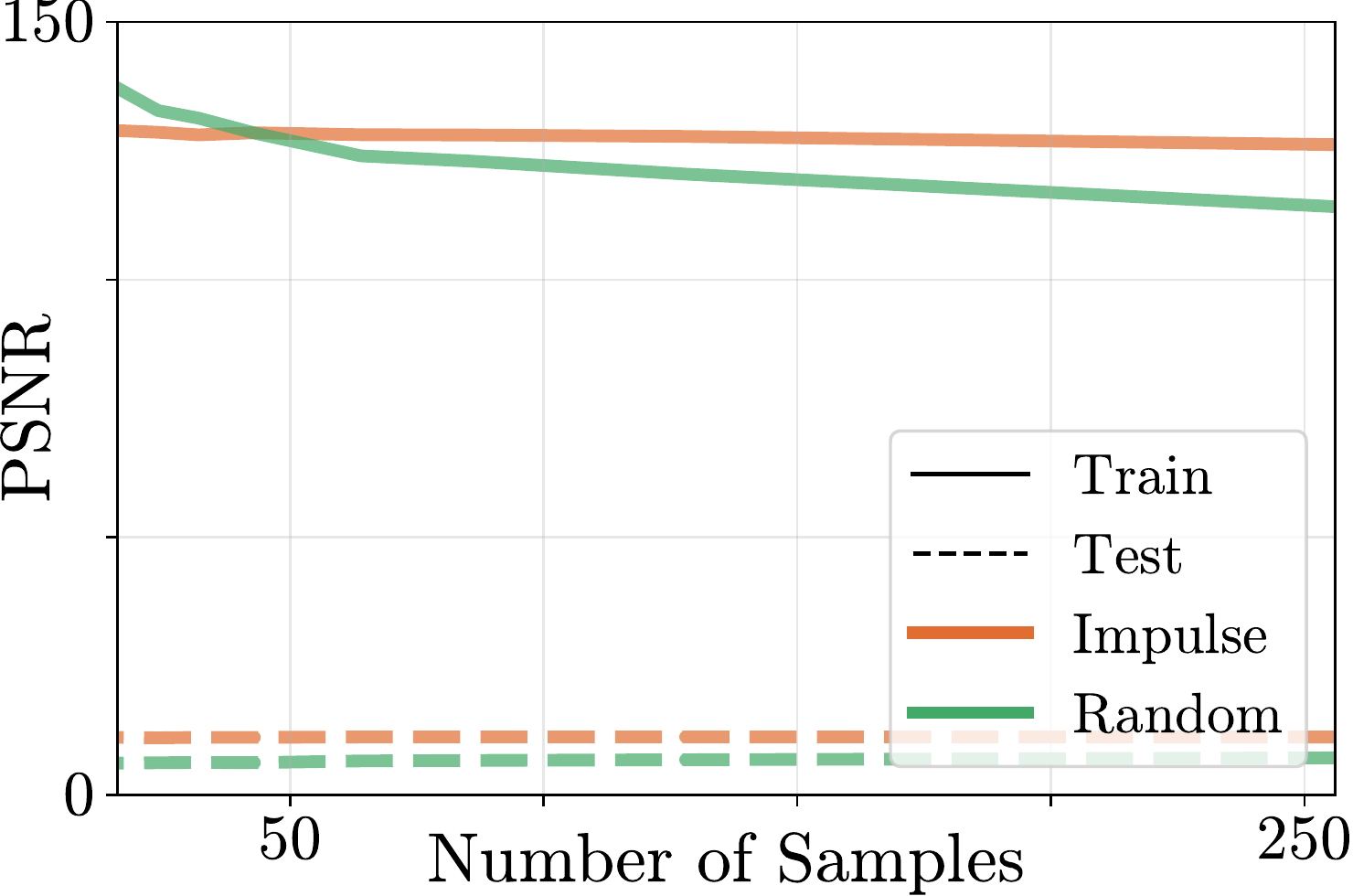}}%
    \subfloat{\includegraphics[width=0.4\textwidth]{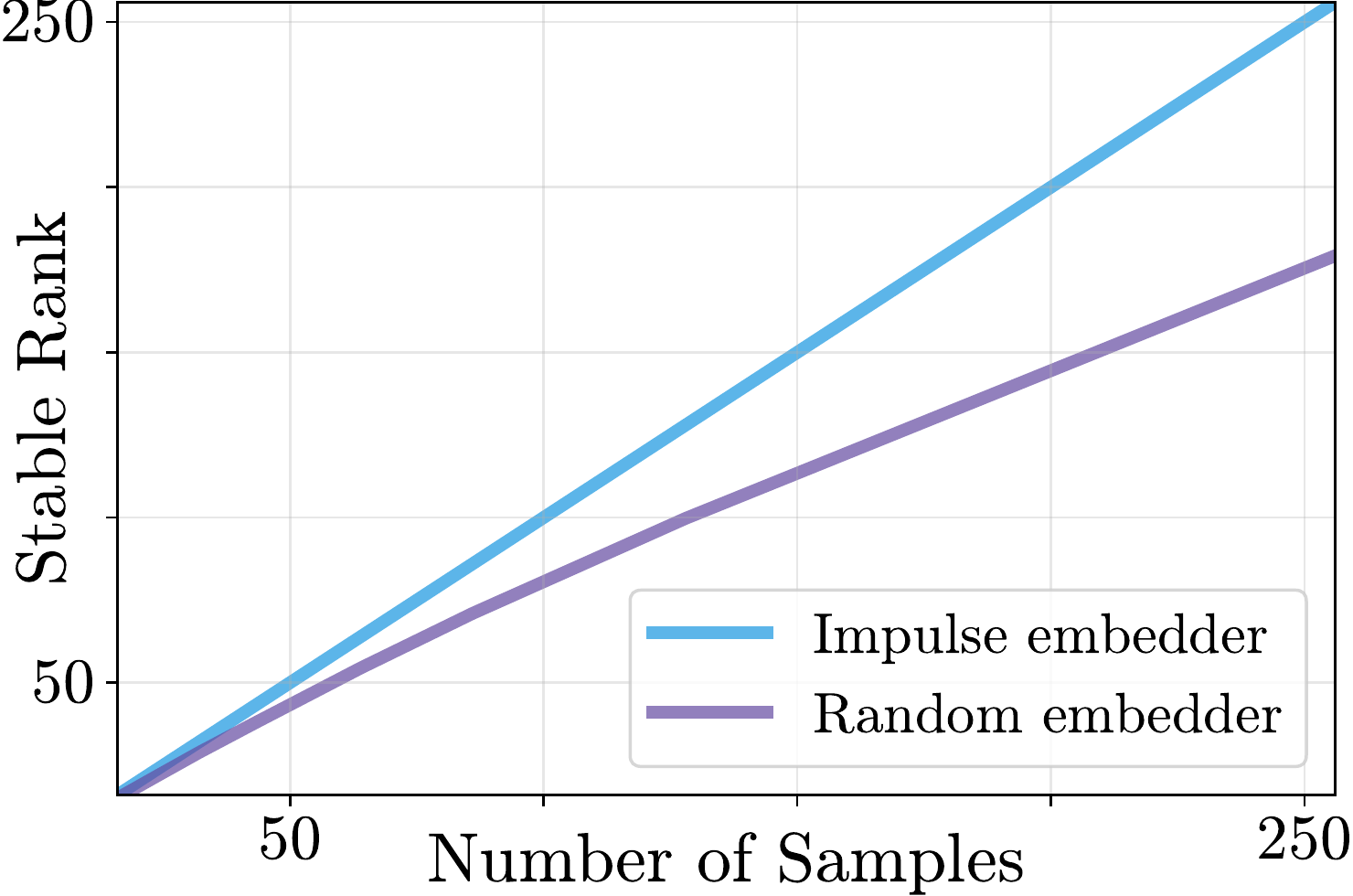}}%
      \qquad
    \subfloat{\includegraphics[width=0.4\textwidth]{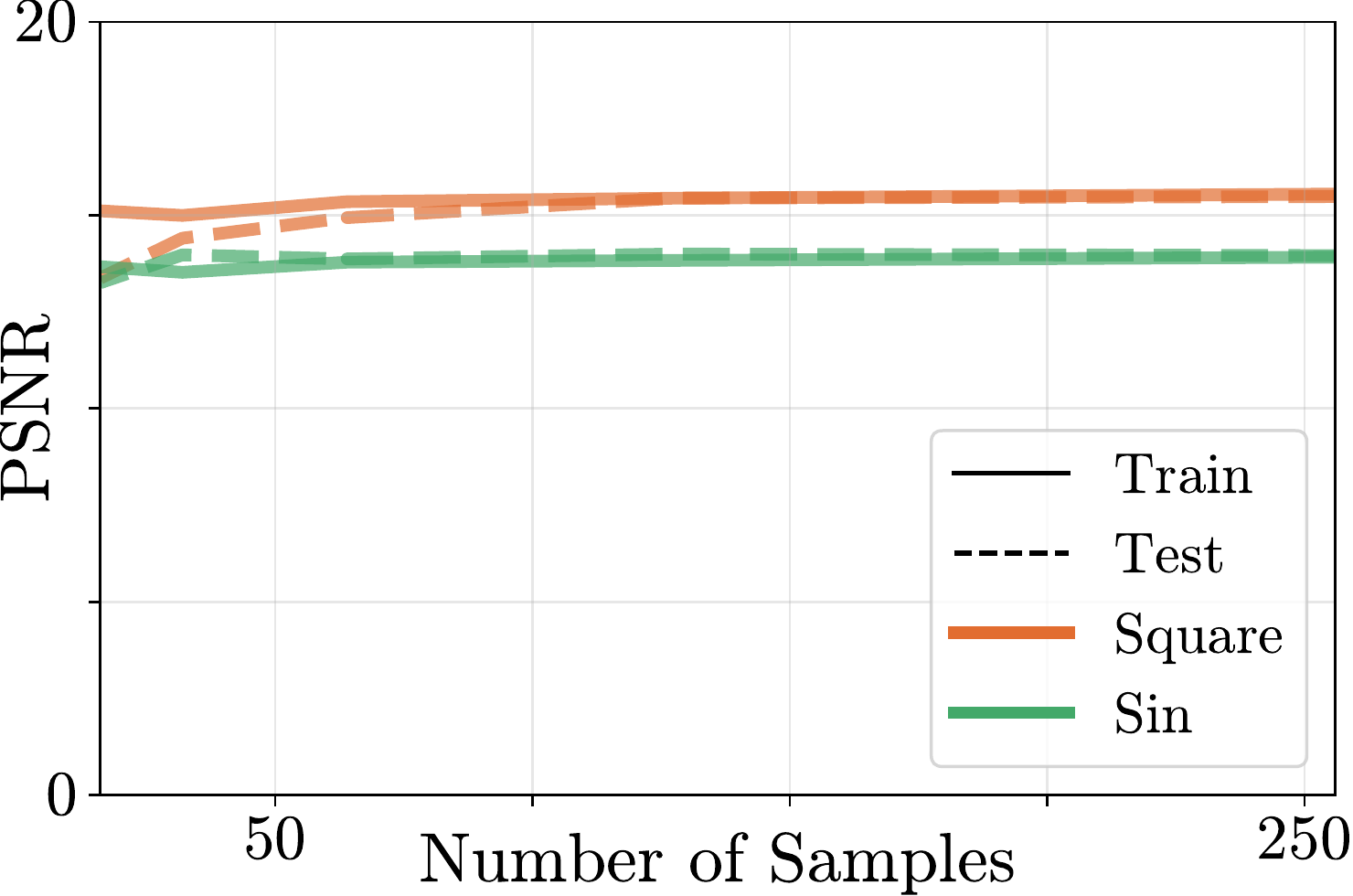}}%
    \subfloat{\includegraphics[width=0.4\textwidth]{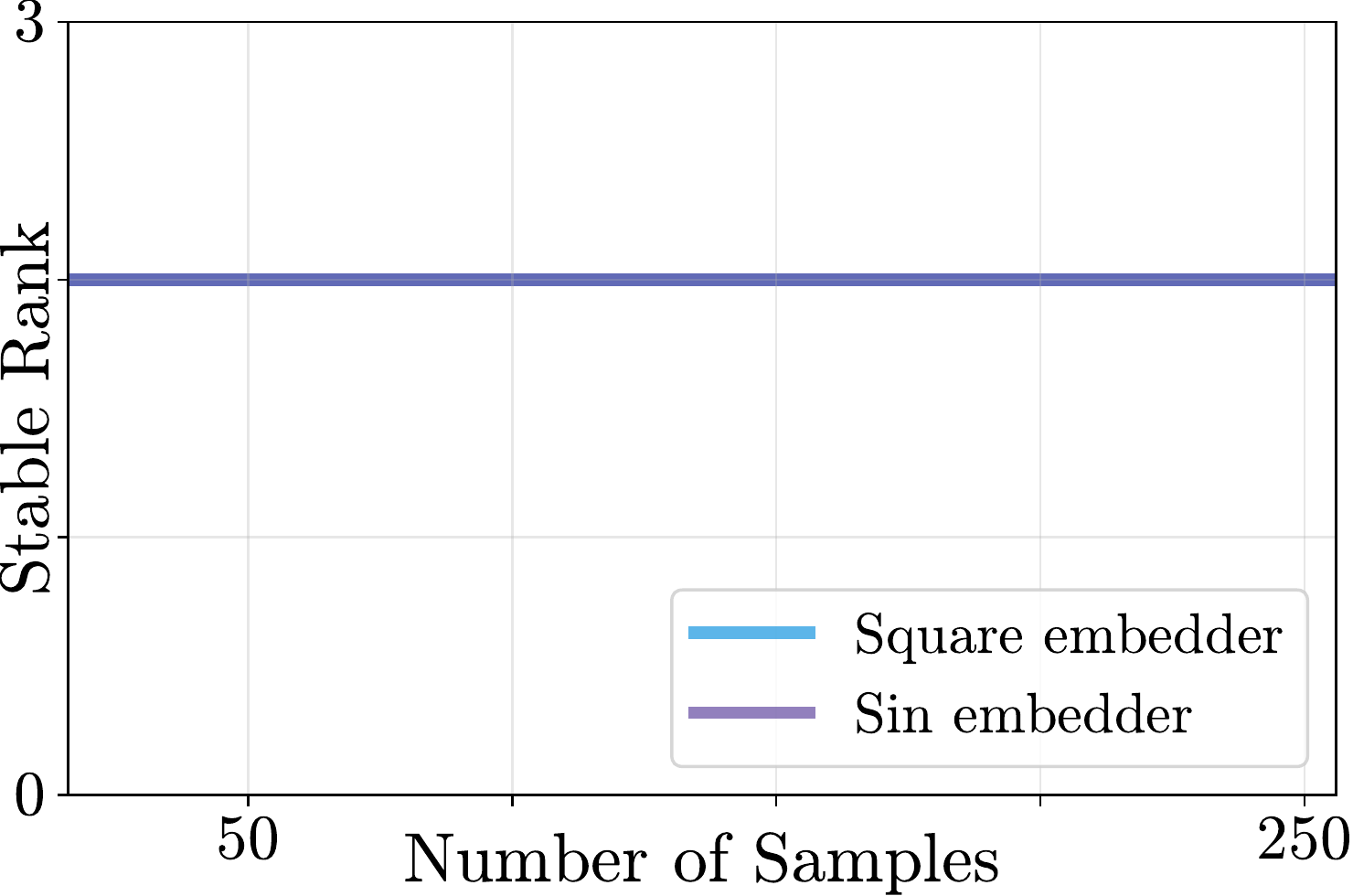}}%
    \caption{ Performance and stable ranks of different embedders against the number of sampled points. }%
    \label{fig:embedders}%
\end{figure}

\textbf{Impulse \& Random embedder:} We implement the impulse embedder as a one-hot encoder. The results are shown in Fig.~\ref{fig:embedders}. It is clear from the theory due to the unbounded stable rank (or equivalently, the unlimited bandwidth) of the impulse embedder, the train PSNR should be high and the test PSNR should be poor. To further validate our theory, we employ a random noise as the embedder and compare its performance. Note that the bandwidth of a random noise is extremely high and as expected, its behavior is similar to the impulse embedder.

\textbf{Square \& Sin embedder:} As discussed in Sec.~\ref{sec:analysis}, these two embedders contain low stable ranks that are approximately upper-bounded at $2$.  Consequently, we can observe a low training PSNR in Fig.~\ref{fig:embedders}. Interestingly, we also observe a lower test PSNR, since extremely small stable ranks cause overly smooth generalizations.

\begin{figure}[ht]
\captionsetup{size=small}
\centering 
\begin{subfigure}{0.42\textwidth}
\includegraphics[width=\linewidth]{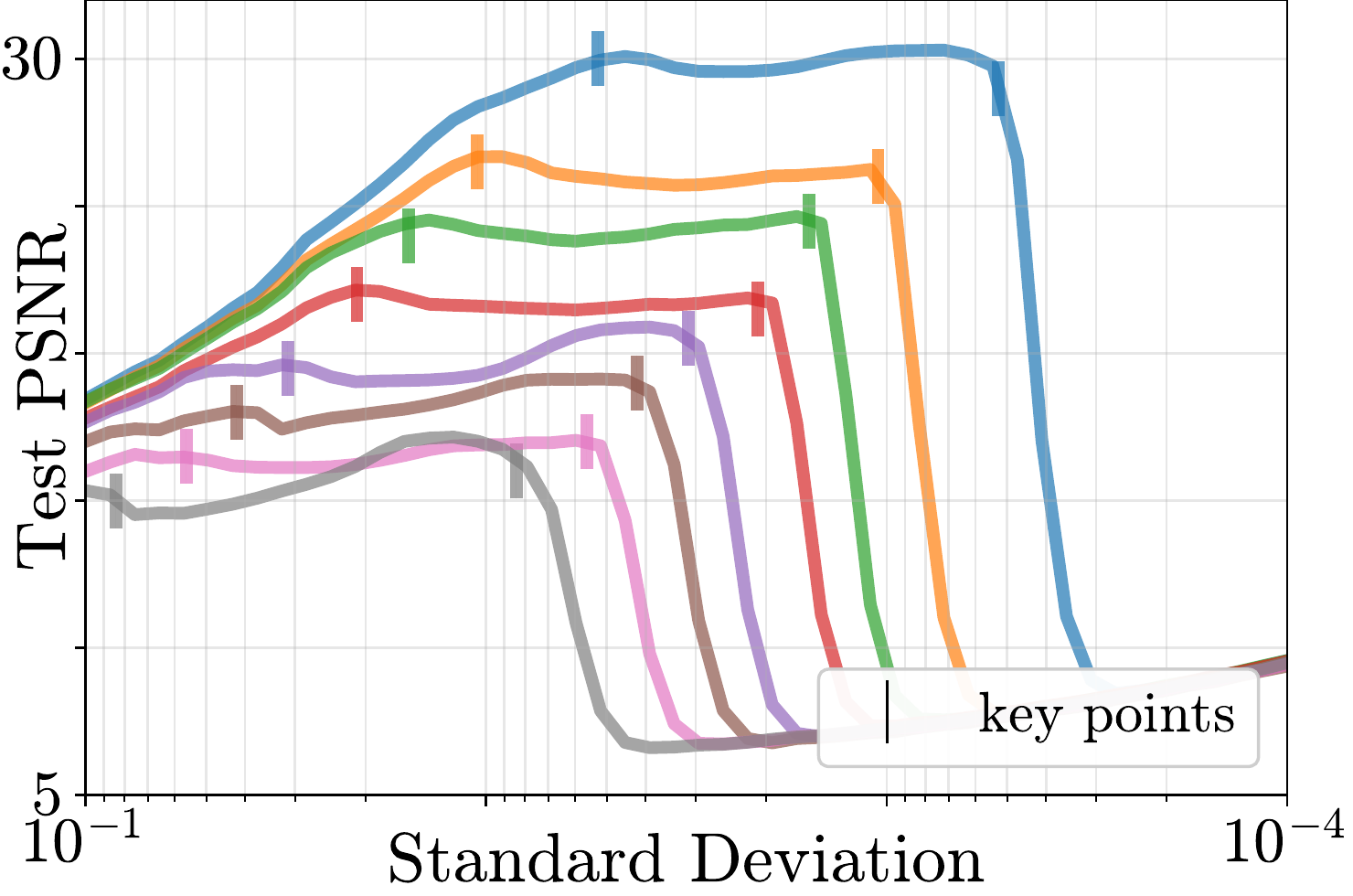} 
\end{subfigure} 
\begin{subfigure}{0.56\textwidth}
\includegraphics[width=\linewidth]{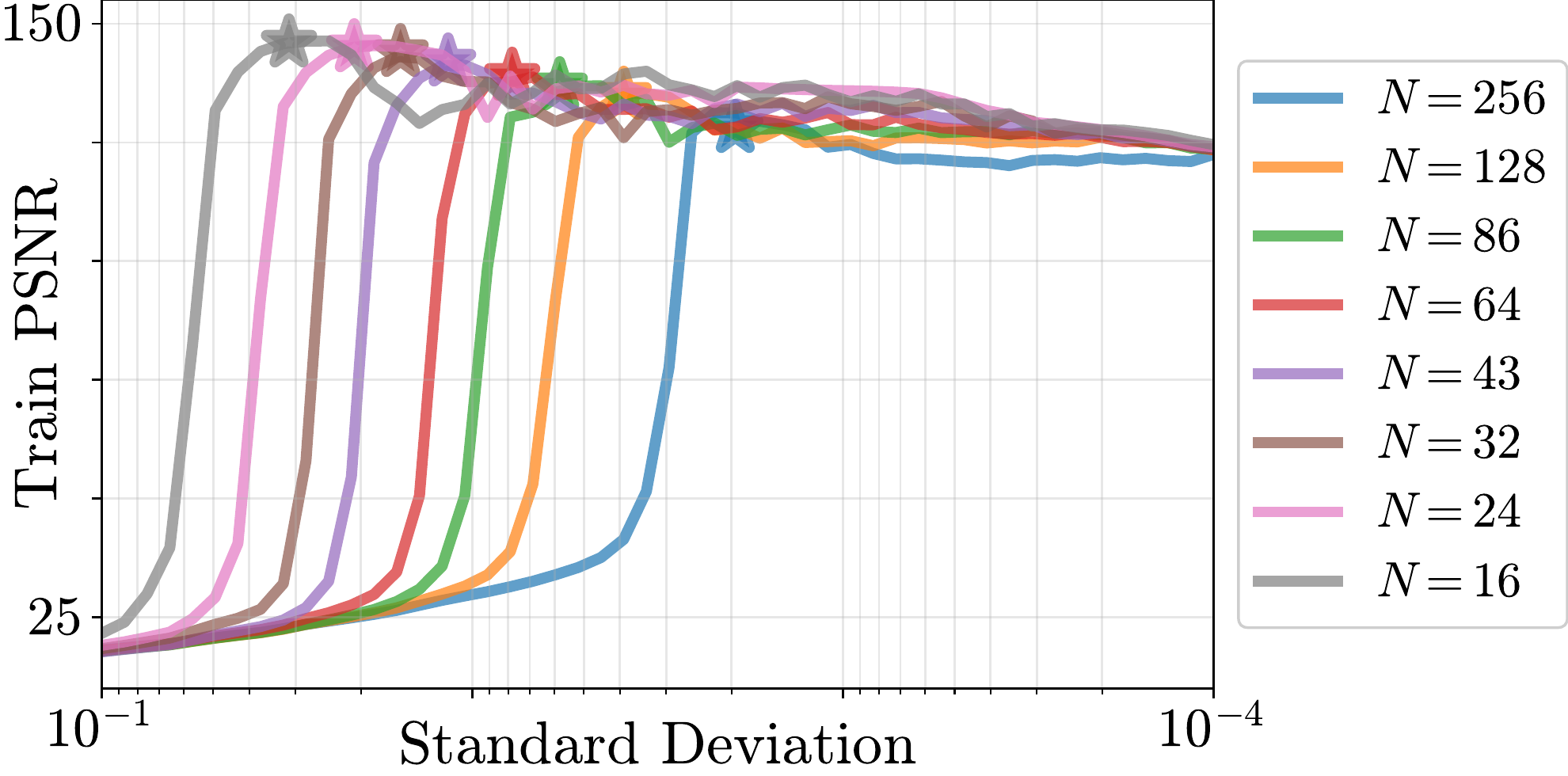} 
\end{subfigure} 
 \caption{Performance of the Gaussian embedder against the standard deviation over different number of input samples.}%
    \label{fig:per_gau}%
\end{figure}

\textbf{Gaussian embedder:} The experimental results for the Gaussian embedder is depicted in Fig.~\ref{fig:per_gau}. As per Proposition \ref{prop:gaussian}, the standard deviation $\sigma$ of the Gaussian embedder can tune the trade-off between the rank and distance preservation. For each number of samples $N$, $50$ samples of $\sigma$ are chose log-linearly from $[10^{-4},10^{-1}]$. Recall that the stable rank of the Gaussian embedding matrix is $\min(N, \frac{1}{2\sqrt{\pi}\sigma})$, and a higher rank can achieve good memorization (a higher training PSNR). However, from the experimental results, it is evident that we do not need full rank to have a good enough PSNR on real signals since real signals contain redundancies. Experimentally,  training PSNR generally hits a peak near $\sigma = \frac{1}{2N}$ (indicated by star key points). Similarly, the test PSNR depends on distance preservation. The experiment results show that the test PSNR decreases as $\sigma$ get very small. This is intuitive, since when $\sigma$ is too small, the output is overly smoothened. Then, for a small interval of $\sigma$, the test PSNR keeps constant before it begins to drop drastically. The reason is obvious, as when $\sigma$ is very high, the distance preservation is hampered.  To avoid this, we need the distance to be preserved \emph{at least} between the two nearest sampled points from the input signal. Therefore, for $N$ equally spaced training samples, we need distance to be preserved in an interval $l \approx  \frac{1}{2N}$. From the Proposition \ref{prop:gaussian} we know that $\sigma = \frac{1}{4N\sqrt{k\ln{10}}}$ where $k$ is an empirically chosen threshold. For our experiment we choose $k=1.6$, thus, $\sigma = \frac{1}{4N\sqrt{1.6\times\ln{10}}}$. This corresponds well with the rightmost key points on the test PSNR plot. The leftmost key points on test PSNR plot is $\sigma = \frac{10}{4N\sqrt{1.6\times\ln{10}}}$, which means the distance is preserved over an interval of five nearest points on either side.

\subsection{The role of stable rank in training an MLP}
A bulk of our derivations stem from the fact that the stable rank plays a key role in determining the performance of the positional embedding. Therefore, it is compelling to observe the connection of the stable rank to the performance and behavior of a network during training. 

\begin{figure}[ht]
\captionsetup{size=small}
\centering 
\begin{subfigure}{0.015\textwidth}\hspace{-2mm}
\rotatebox[origin=c]{90}{\tiny{Model performance}}
\end{subfigure}
\begin{subfigure}{0.24\textwidth}
\includegraphics[width=\linewidth]{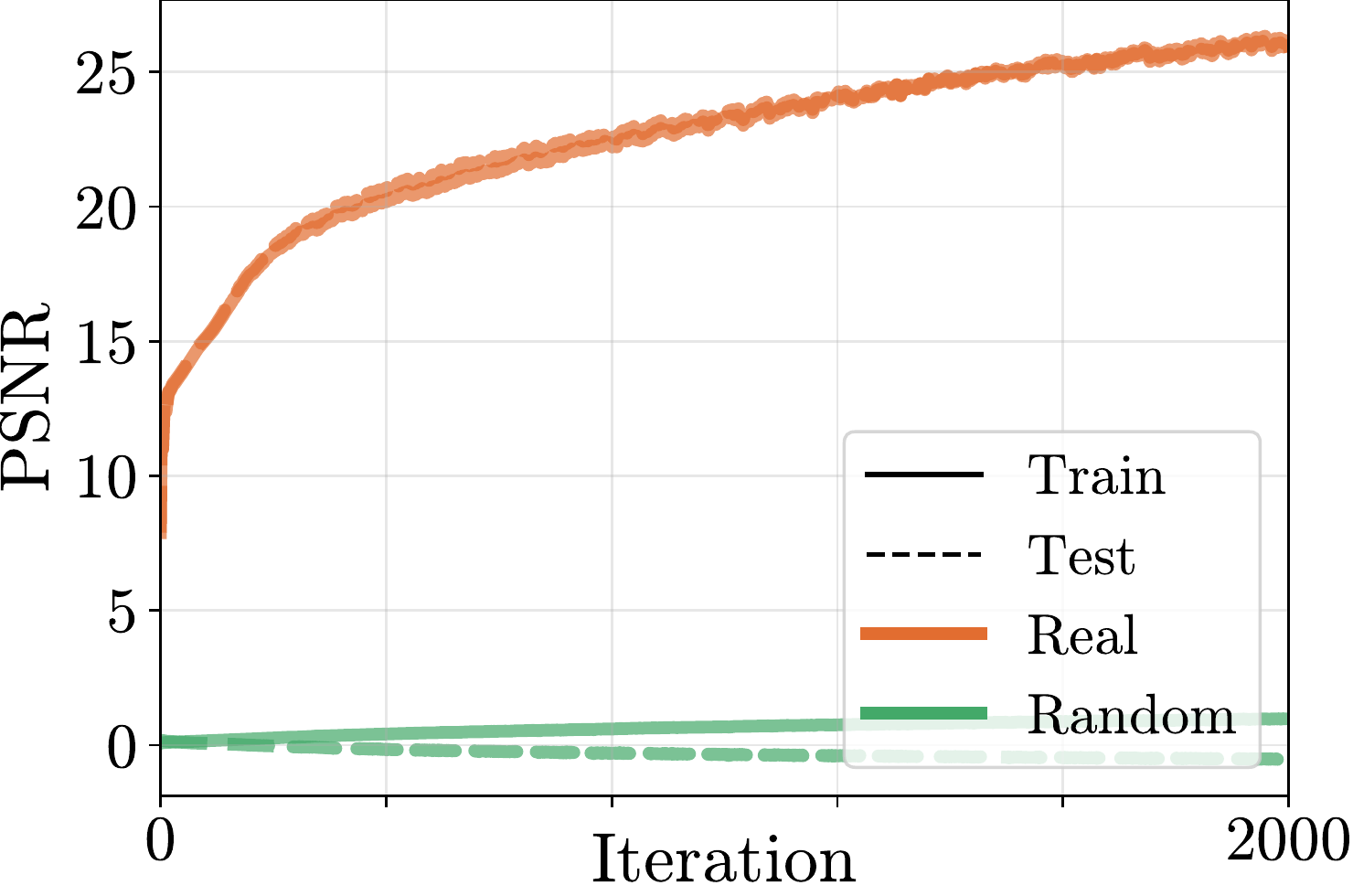} 
\end{subfigure} 
\begin{subfigure}{0.24\textwidth}
\includegraphics[width=\linewidth]{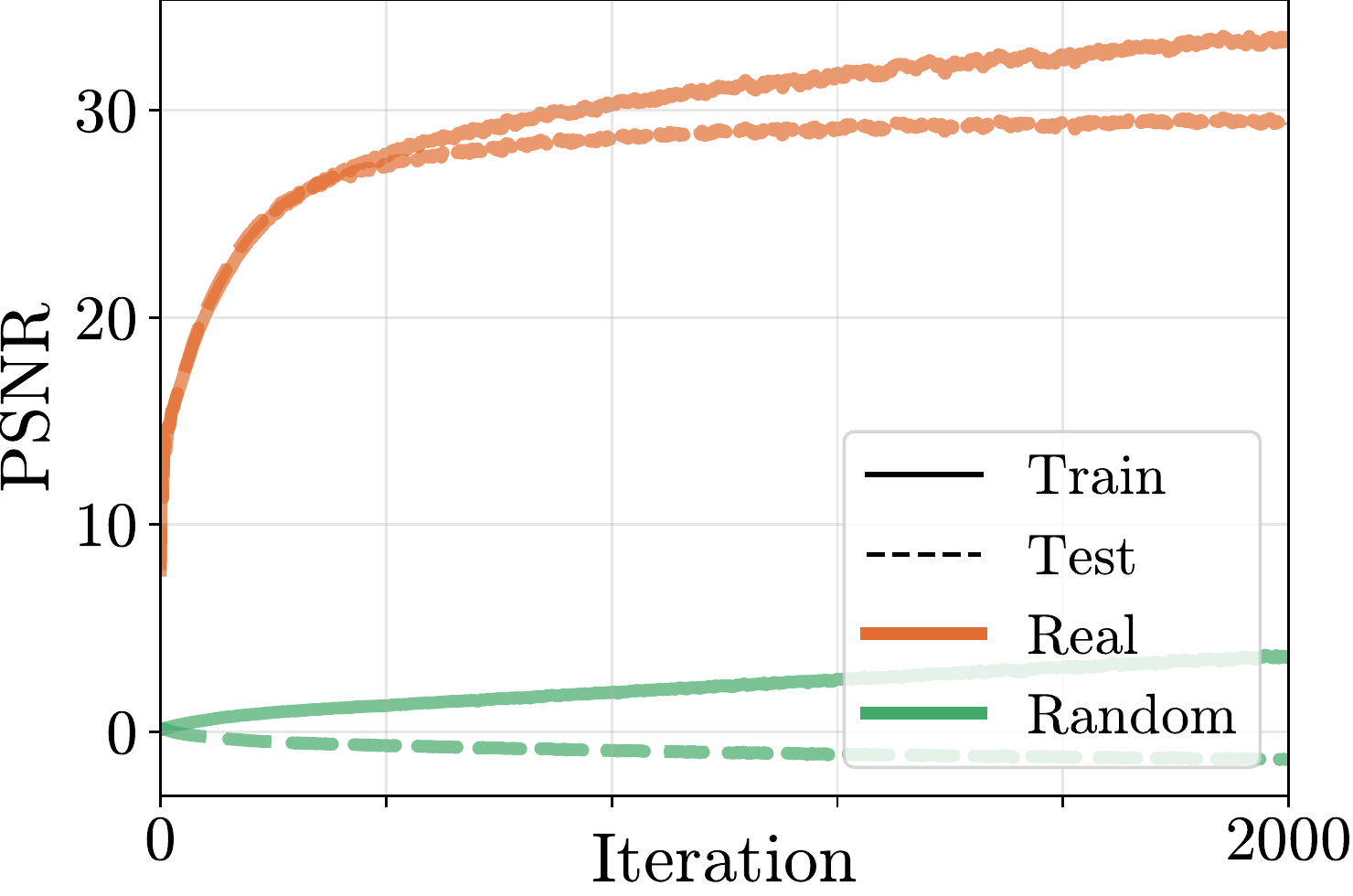} 
\end{subfigure} 
\begin{subfigure}{0.24\textwidth}
\includegraphics[width=\linewidth]{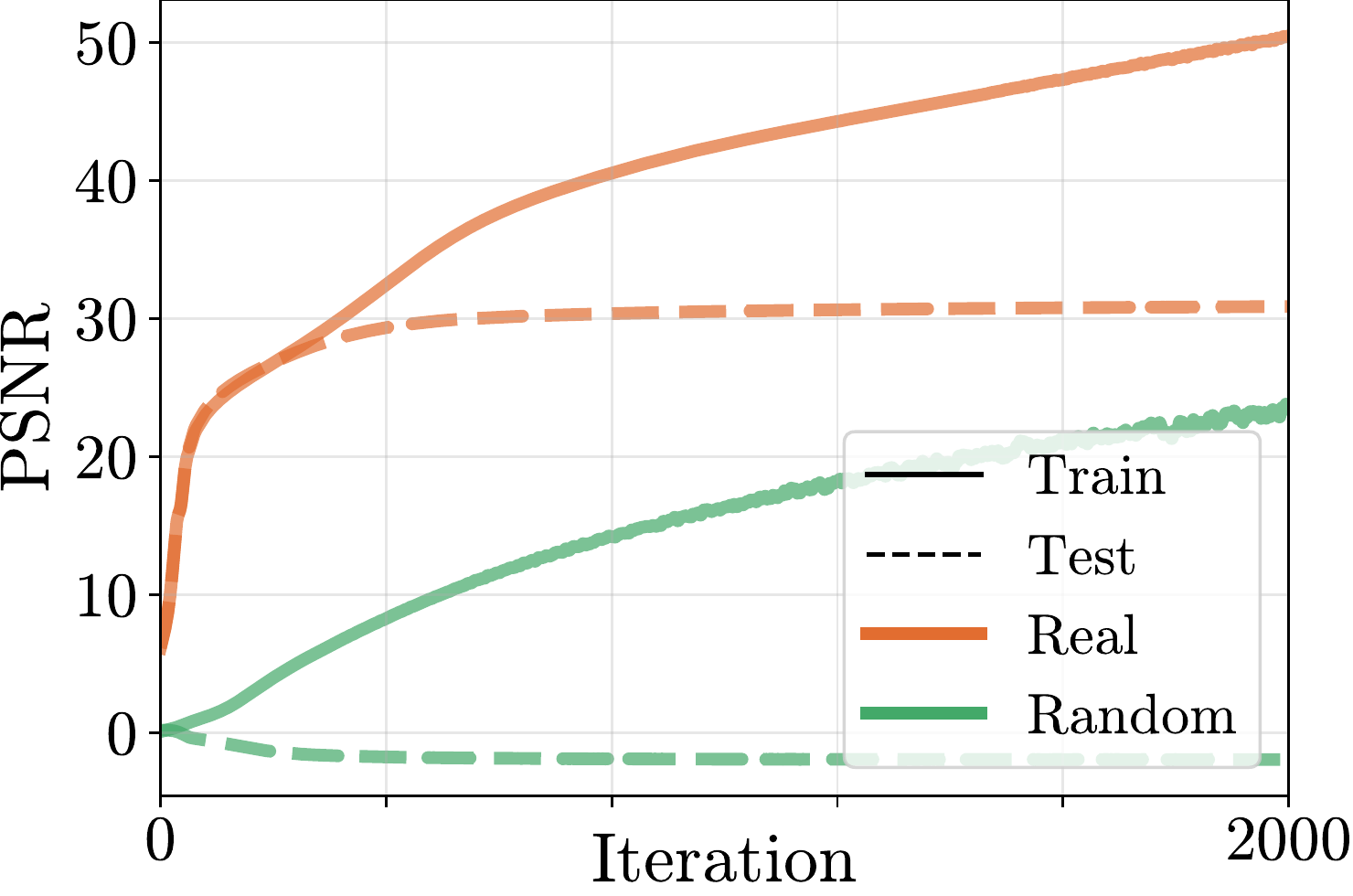} 
\end{subfigure} 
\begin{subfigure}{0.24\textwidth}
\includegraphics[width=\linewidth]{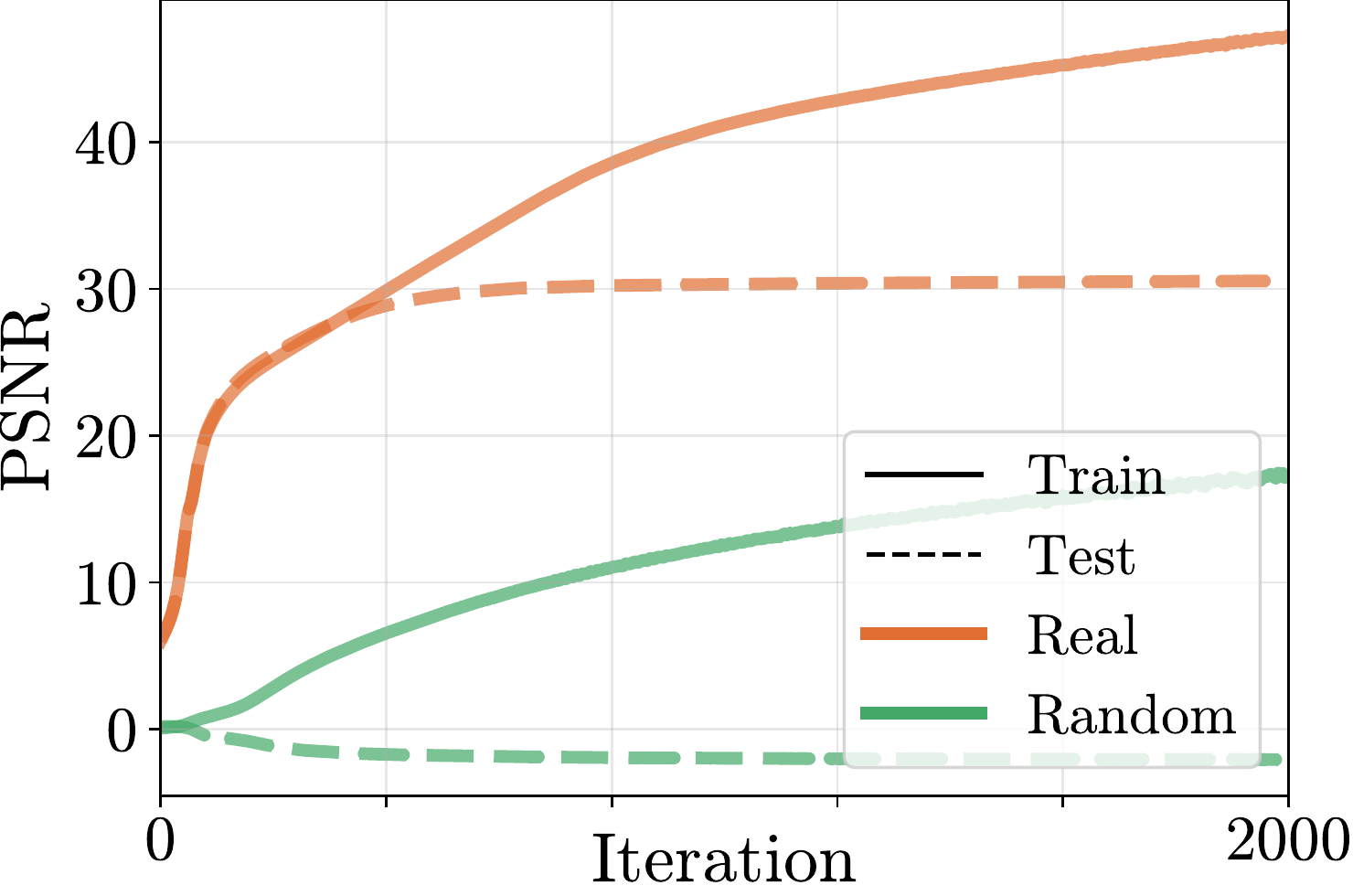} 
\end{subfigure} 

\begin{subfigure}{0.015\textwidth}\hspace{-2mm}
\rotatebox[origin=c]{90}{\tiny{Random signal}}
\end{subfigure}
\begin{subfigure}{0.24\textwidth}
\includegraphics[width=\linewidth]{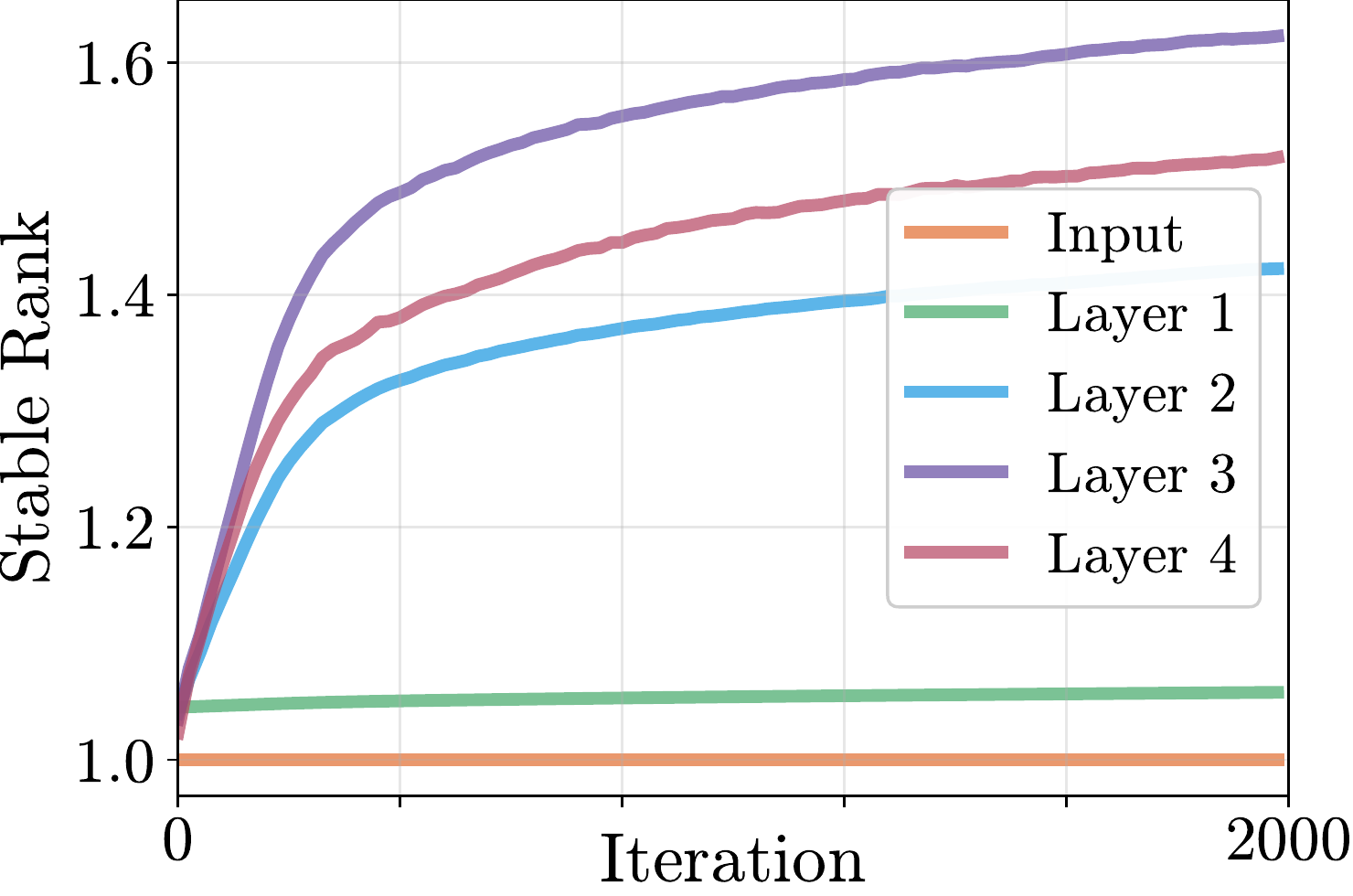} 
\end{subfigure} 
\begin{subfigure}{0.24\textwidth}
\includegraphics[width=\linewidth]{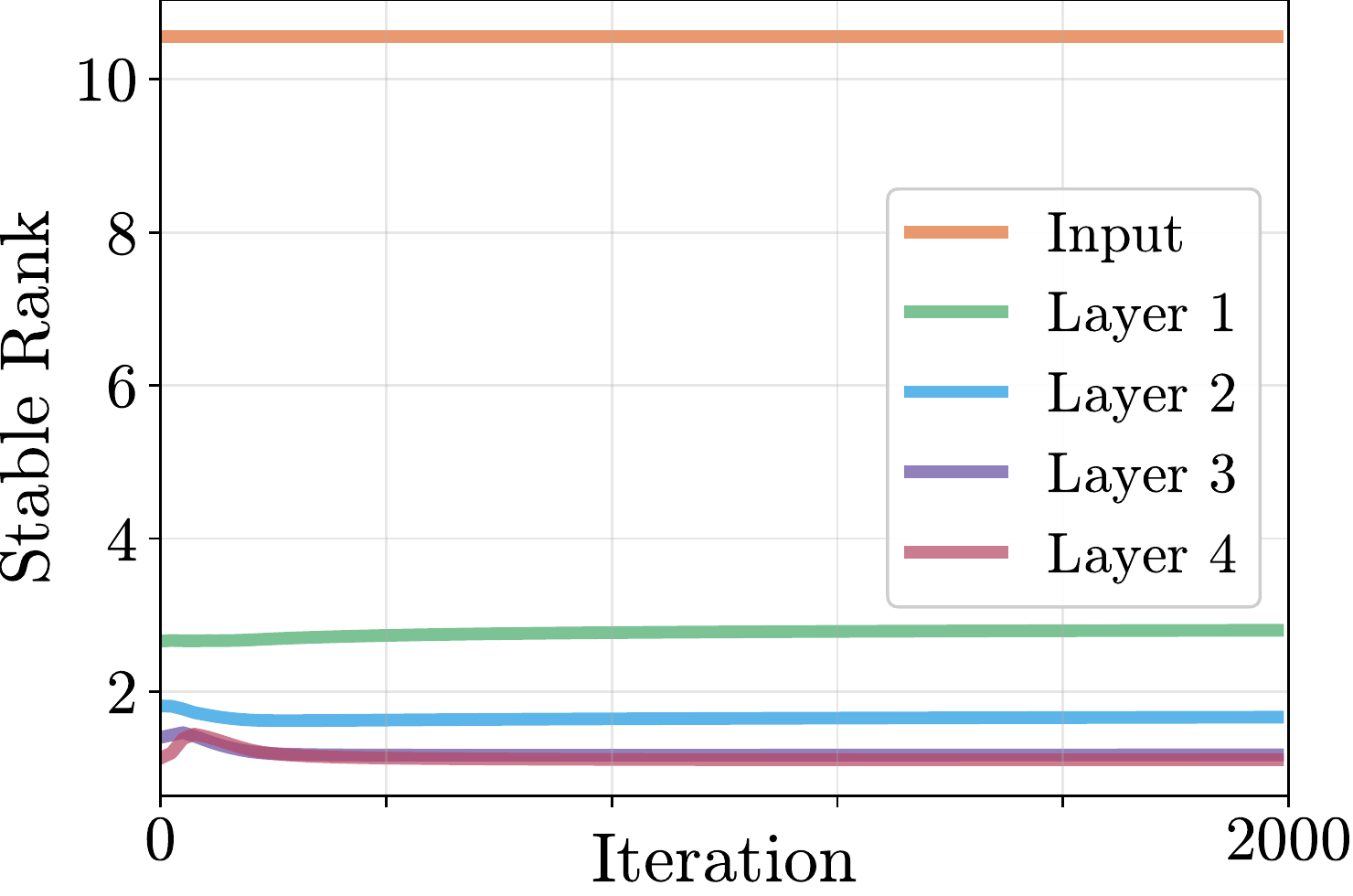} 
\end{subfigure} 
\begin{subfigure}{0.24\textwidth}
\includegraphics[width=\linewidth]{images/First_4_layer/First_rff_SR_noise.pdf} 
\end{subfigure} 
\begin{subfigure}{0.24\textwidth}
\includegraphics[width=\linewidth]{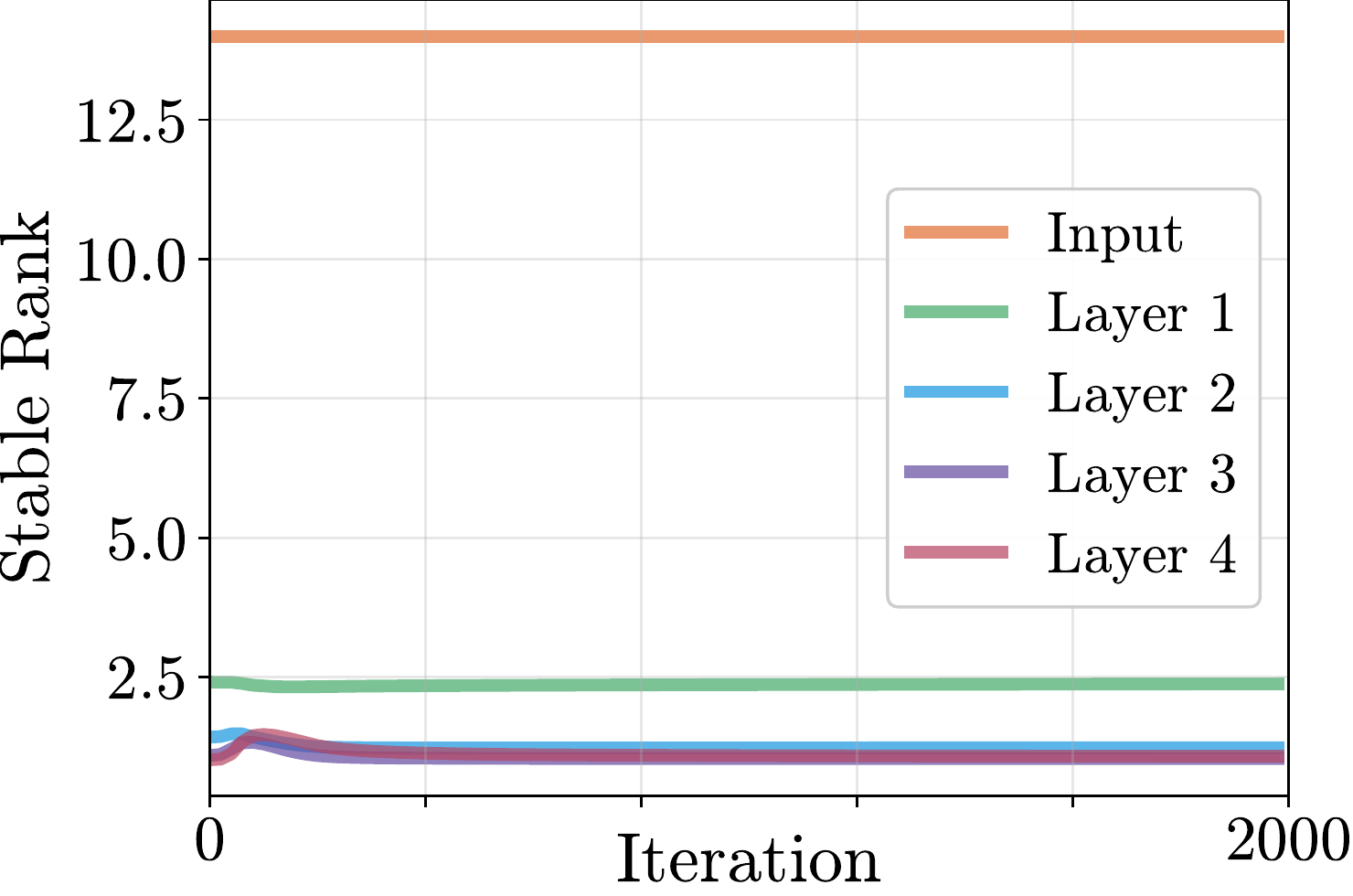} 
\end{subfigure} 

\begin{subfigure}{0.015\textwidth}\hspace{-2mm}
\rotatebox[origin=c]{90}{\tiny{Real signal}}
\end{subfigure}
\begin{subfigure}{0.24\textwidth}
\includegraphics[width=\linewidth]{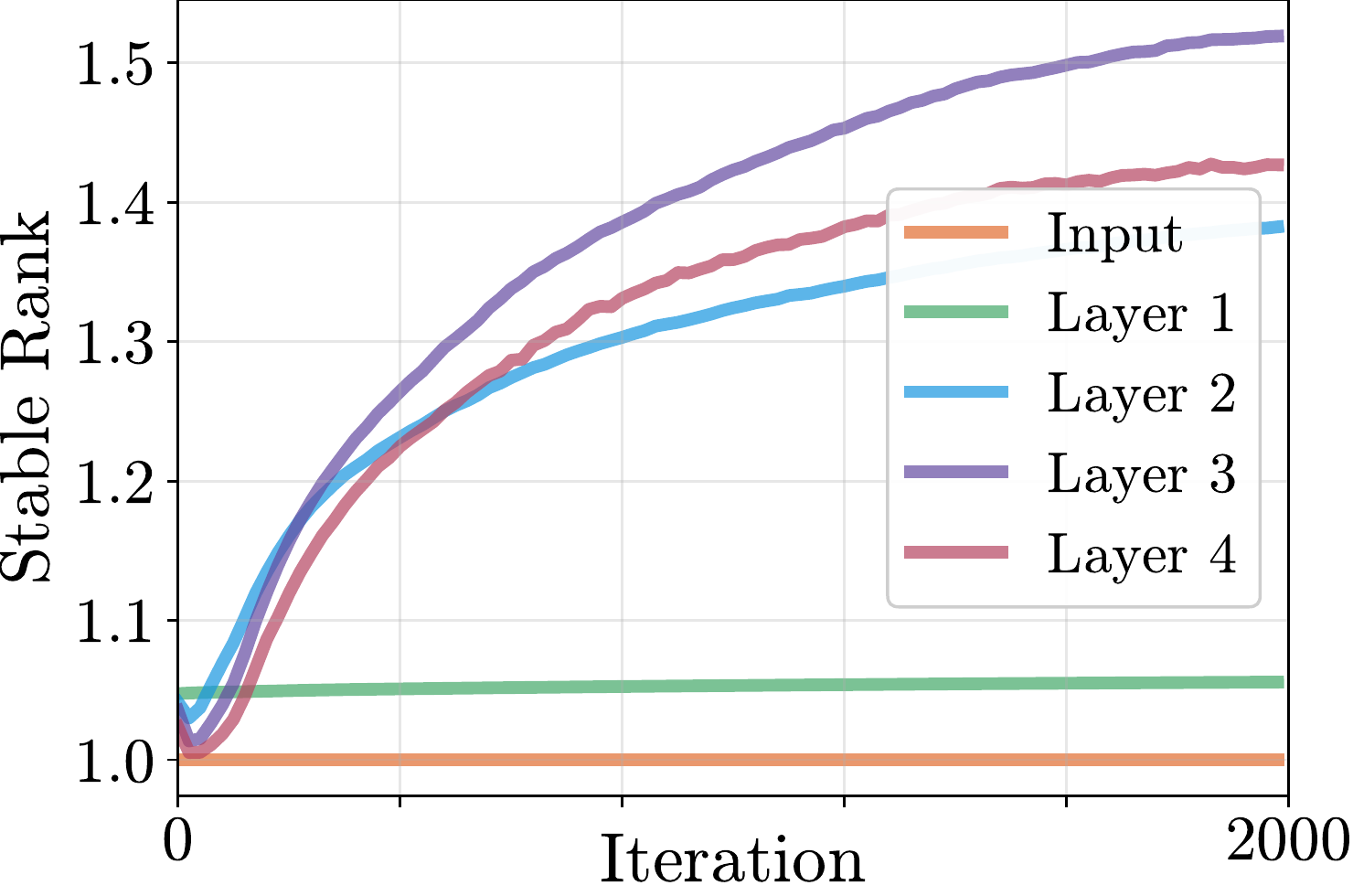} 
\caption{No embedder}
\end{subfigure} 
\begin{subfigure}{0.24\textwidth}
\includegraphics[width=\linewidth]{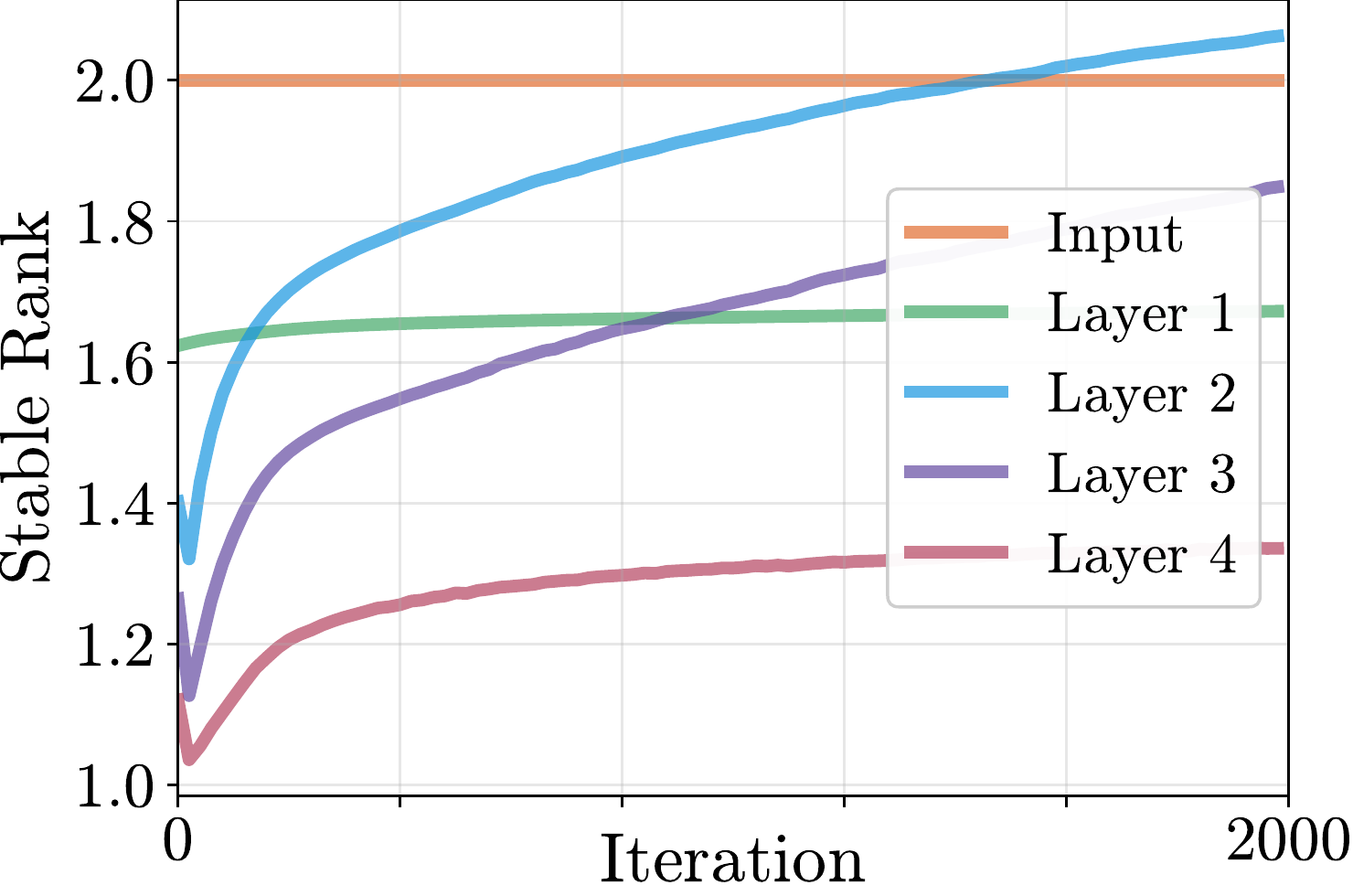} 
\caption{Sinusoidal embedder}
\end{subfigure} 
\begin{subfigure}{0.24\textwidth}
\includegraphics[width=\linewidth]{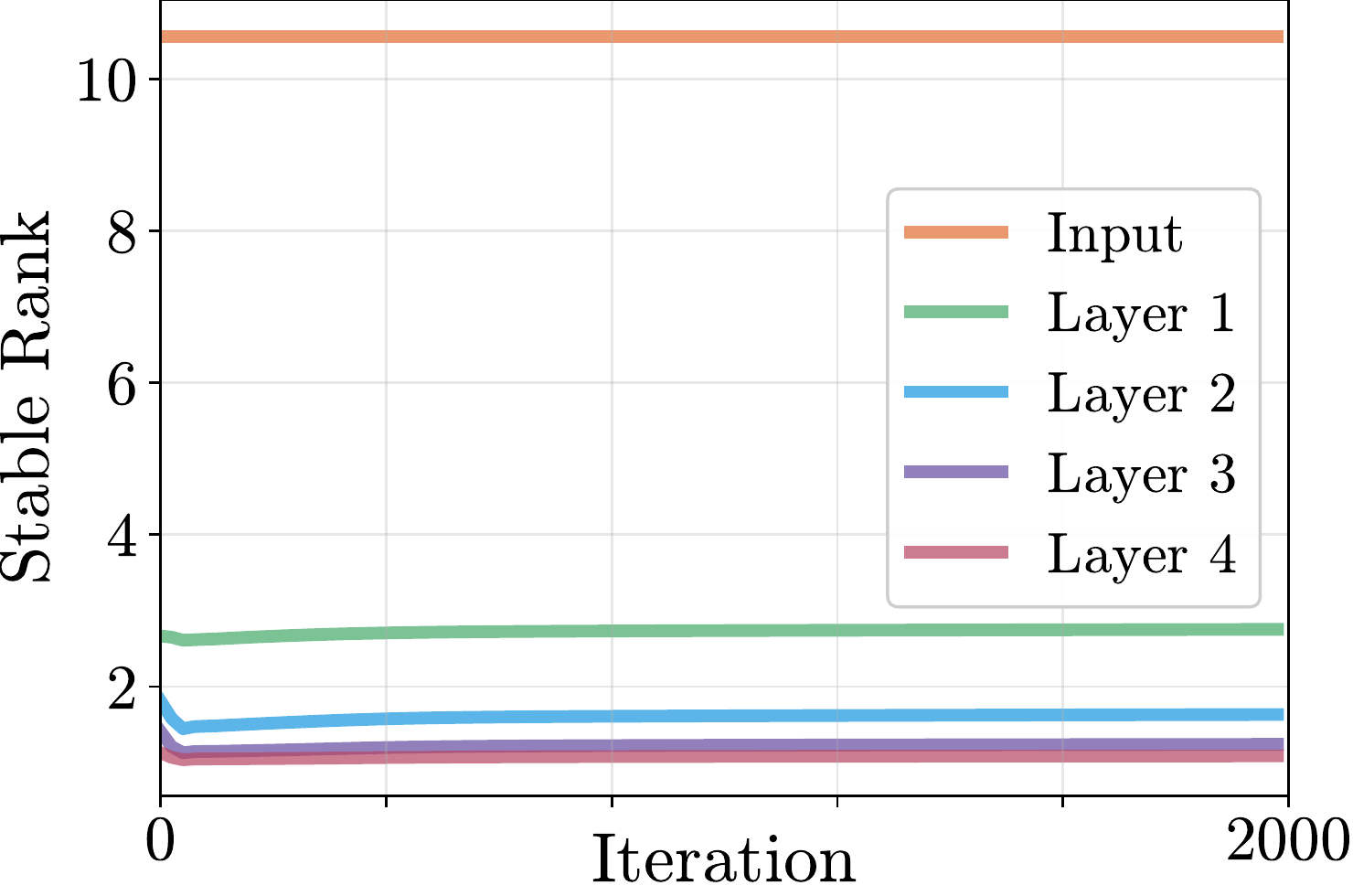} 
\caption{Gaussian embedder}
\end{subfigure} 
\begin{subfigure}{0.24\textwidth}
\includegraphics[width=\linewidth]{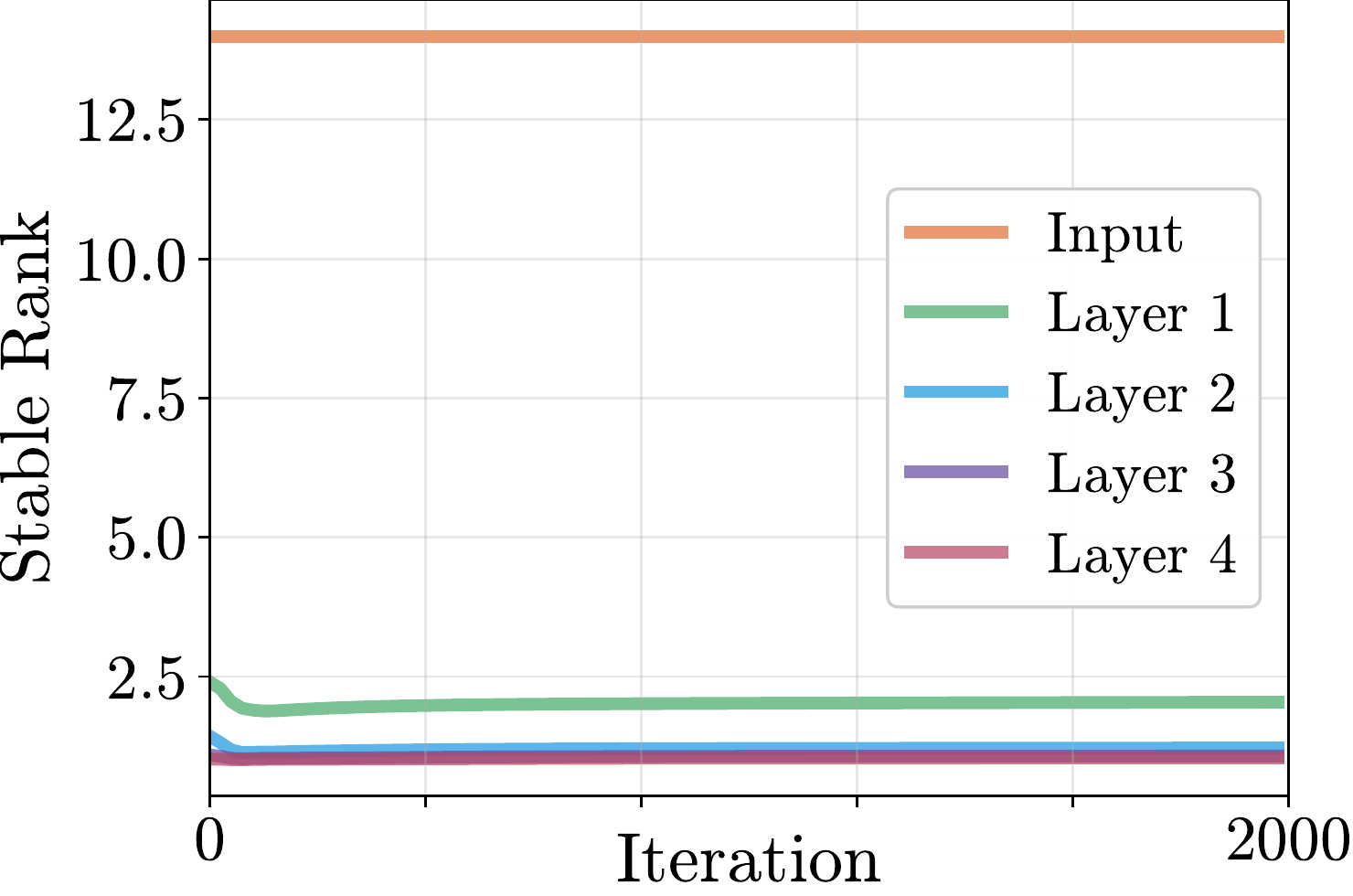} 
\caption{RFF}
\end{subfigure} 
\caption{Connection of the stable rank to the performance and behavior of an MLP. As depicted, if the stable rank of the inputs is too low, the network attempts to increase the stable rank of the hidden layers, causing poor performance and longer training time. As the stable rank of the positional embedding increases, the network demonstrates a lower stable rank at hidden layers, leading to better performance and rapid convergence. }
\label{fig:MLP_performance}
\end{figure}

In this experiment, we consider \emph{real} signals and \emph{random} signals as the targets. For the real signals, we pick $10$ random rows of the Pepper image. Similarly, we obtain $10$ random signals from a Gaussian distribution. We train a 4-layer MLP with ReLU activation for 2000 epochs with Adam optimizer. The learning rate for original coordinates is $1e^{-3}$, and for others, it is $1e^{-4}$. The averaged results over each of the $10$ rows are reported in Fig.~\ref{fig:MLP_performance}. As illustrated, when no encoding mechanism is used, the network attempts to increase the stable rank in each layer, before mapping down to the target, to obtain better performance during the training process. This leads to a considerably higher training time. This trend is common to both real and random signals, although for the random signal, stable rank increases more rapidly. However, since the random signal has no redundancy, the network cannot achieve the desired stable rank, leading to poor performance. In the case of sinusoidal embedder, performance is slightly better since the rank of the input is slightly higher. In contrast, with the RFF and the Gaussian embedder the inputs have sufficient rank, and the network does not have to put effort into increasing the rank. This results in more rapid convergence and better performance. However, the random signals still demand a very high stable rank, causing the networks to demonstrate sub-par performance. Further, observe that the memorization of the training data becomes better as the stable rank of the embedding scheme increase, as predicted in theory.

\subsection{Reconstruction of 1-D signals}
\begin{figure}[ht]
\captionsetup{size=small}
\centering 
\begin{subfigure}{0.24\textwidth}
\includegraphics[width=\linewidth]{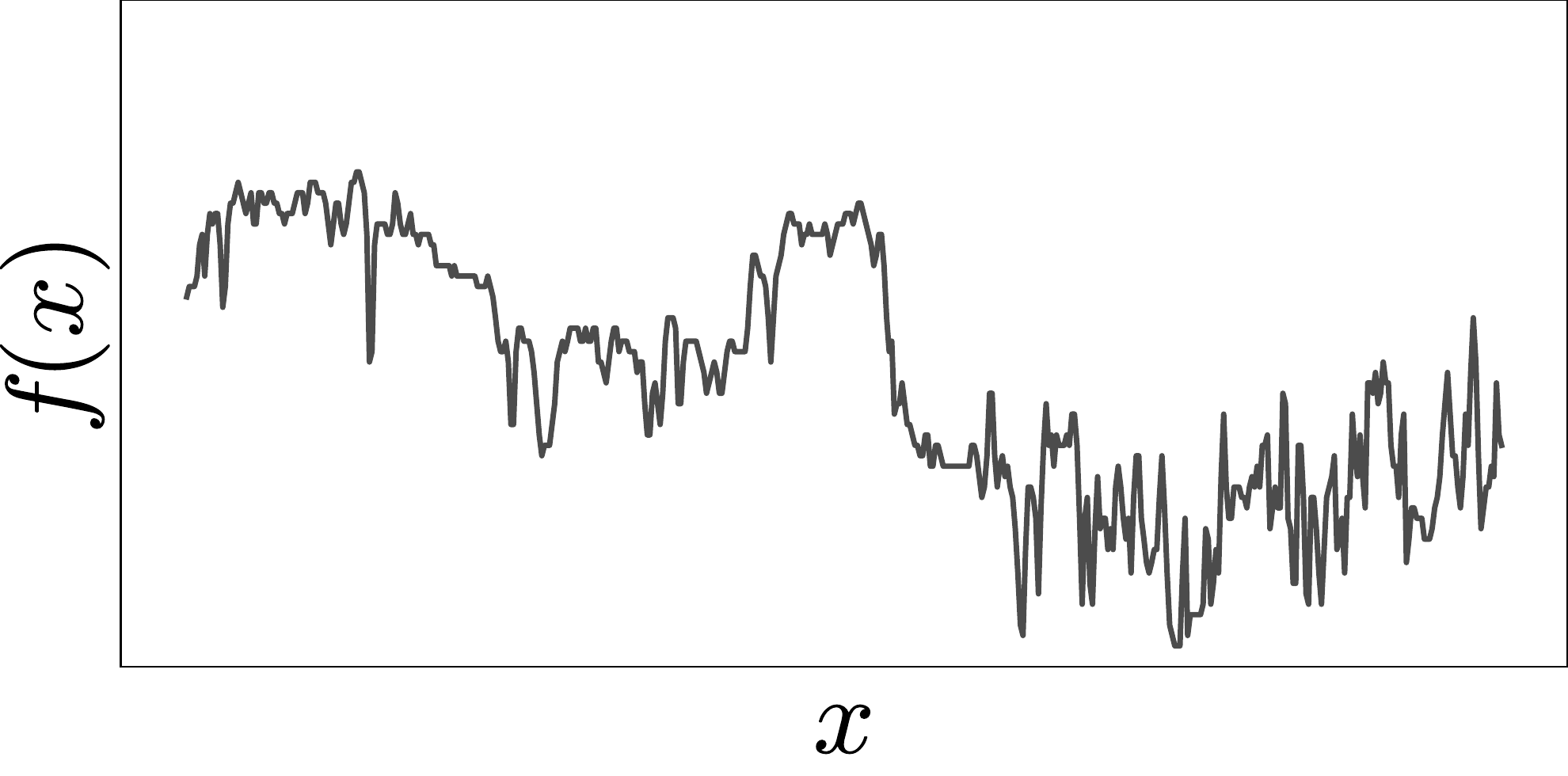} 
\end{subfigure} 
\begin{subfigure}{0.24\textwidth}
\includegraphics[width=\linewidth]{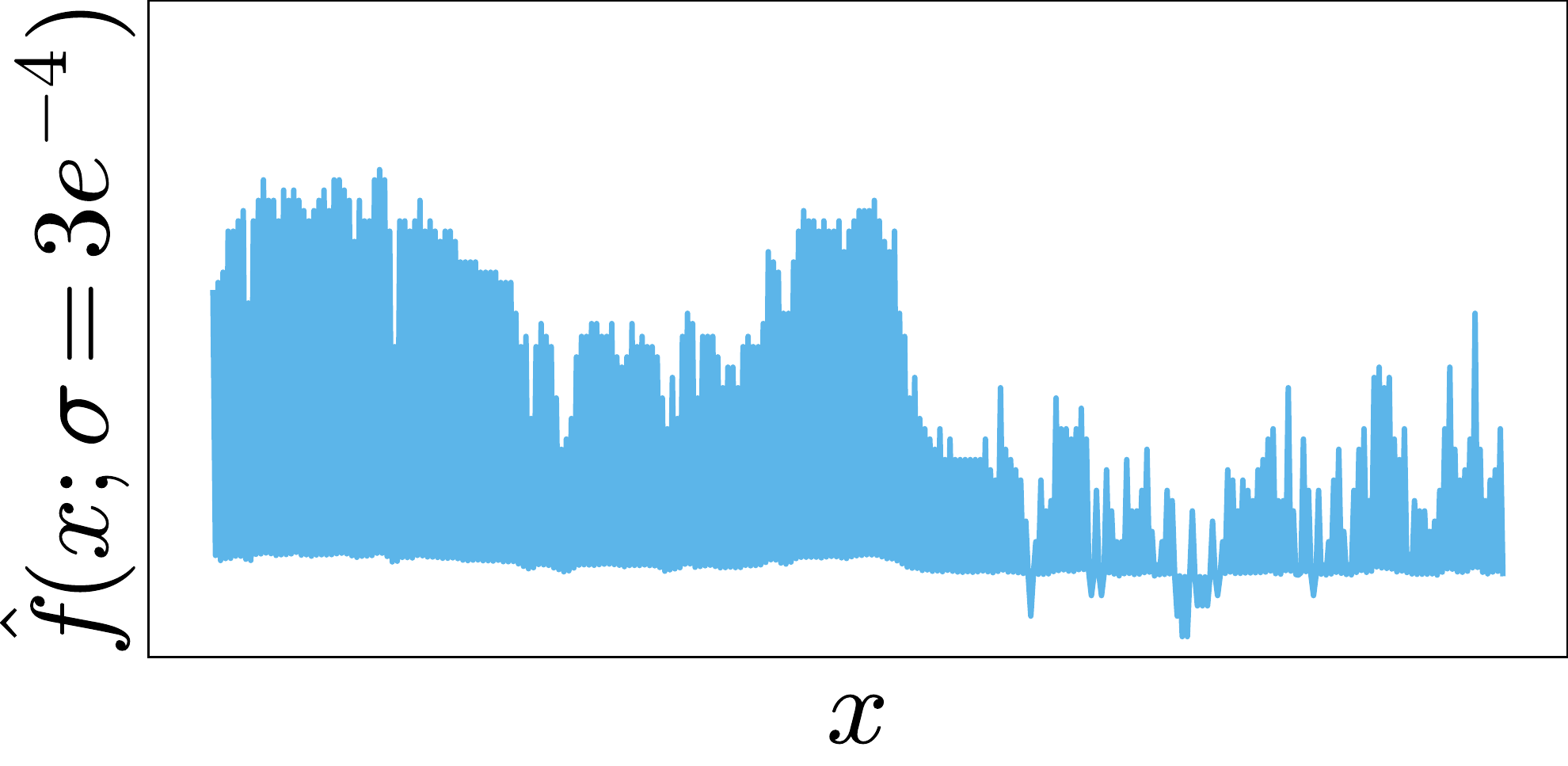} 
\end{subfigure} 
\begin{subfigure}{0.24\textwidth}
\includegraphics[width=\linewidth]{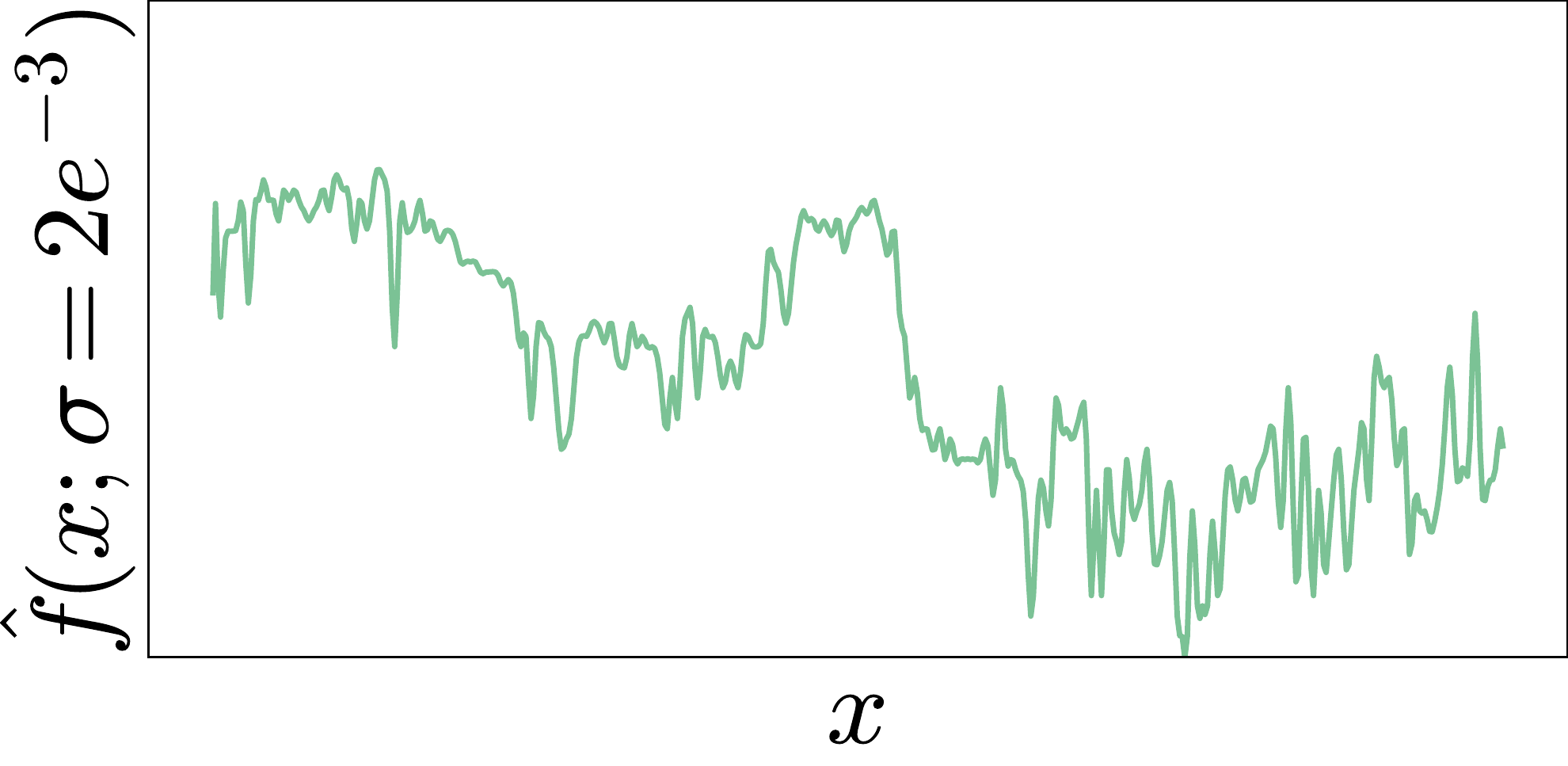} 
\end{subfigure} 
\begin{subfigure}{0.24\textwidth}
\includegraphics[width=\linewidth]{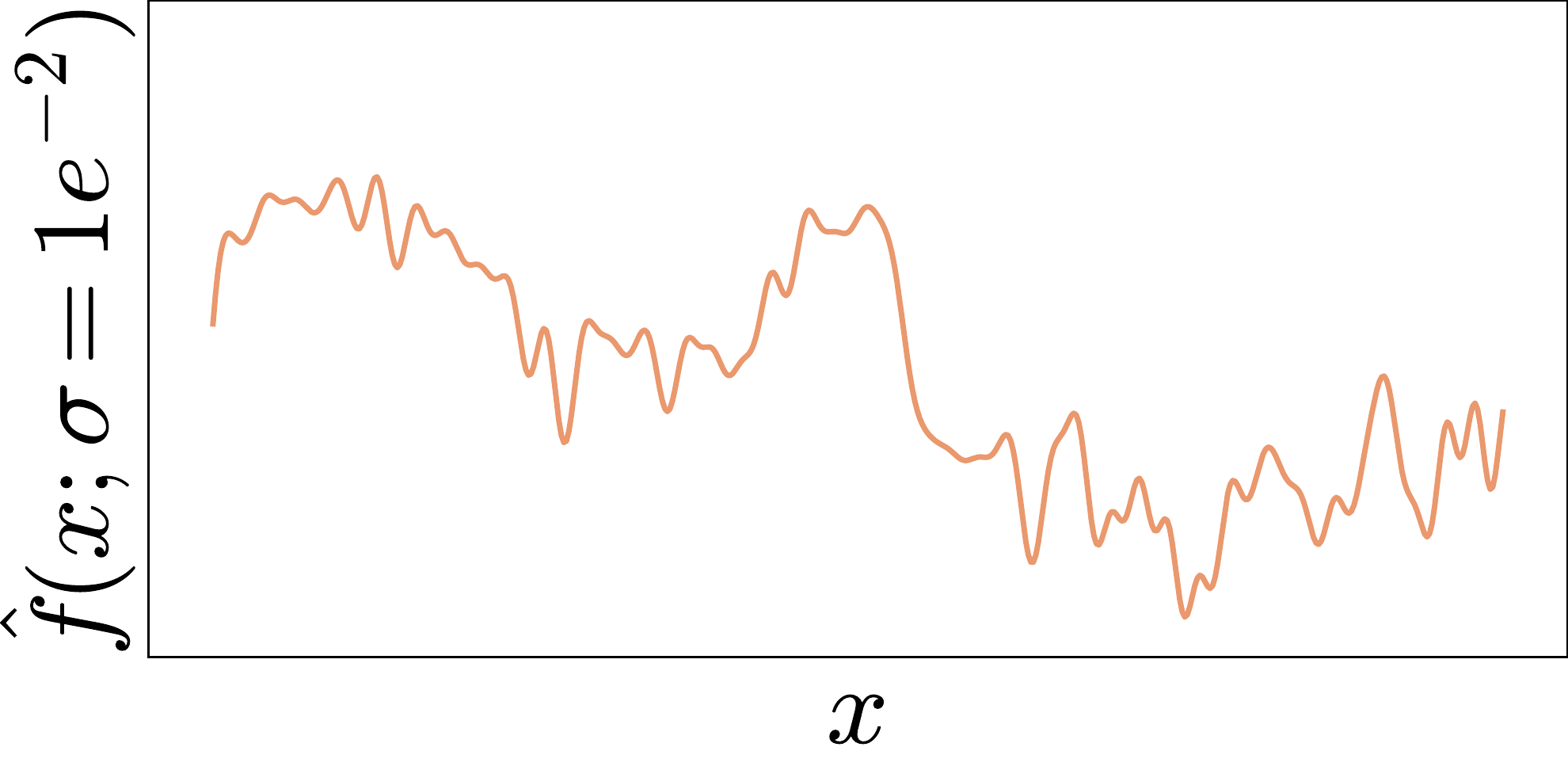} 
\end{subfigure} 

\begin{subfigure}{0.24\textwidth}
\includegraphics[width=\linewidth]{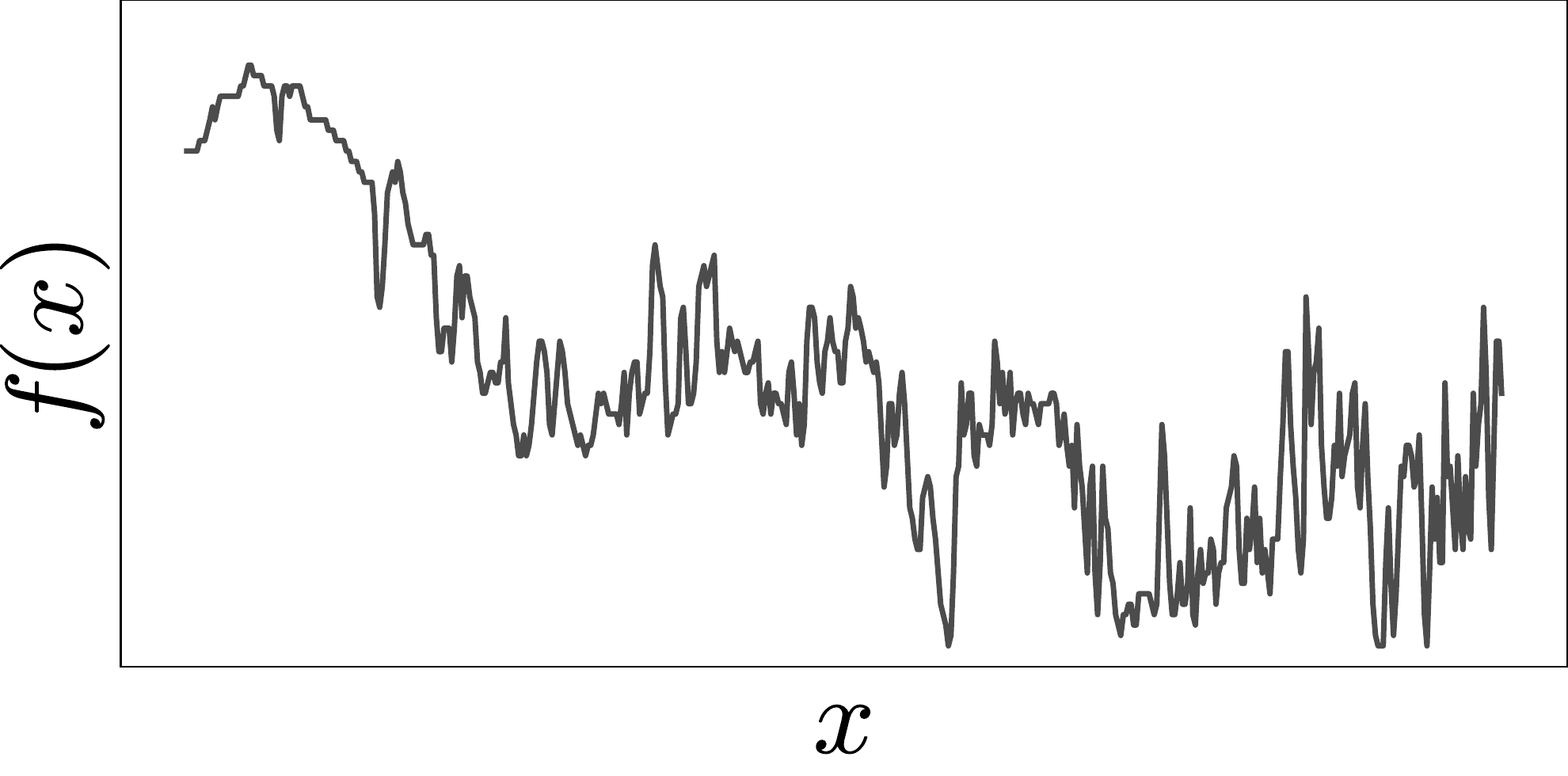} 
\end{subfigure} 
\begin{subfigure}{0.24\textwidth}
\includegraphics[width=\linewidth]{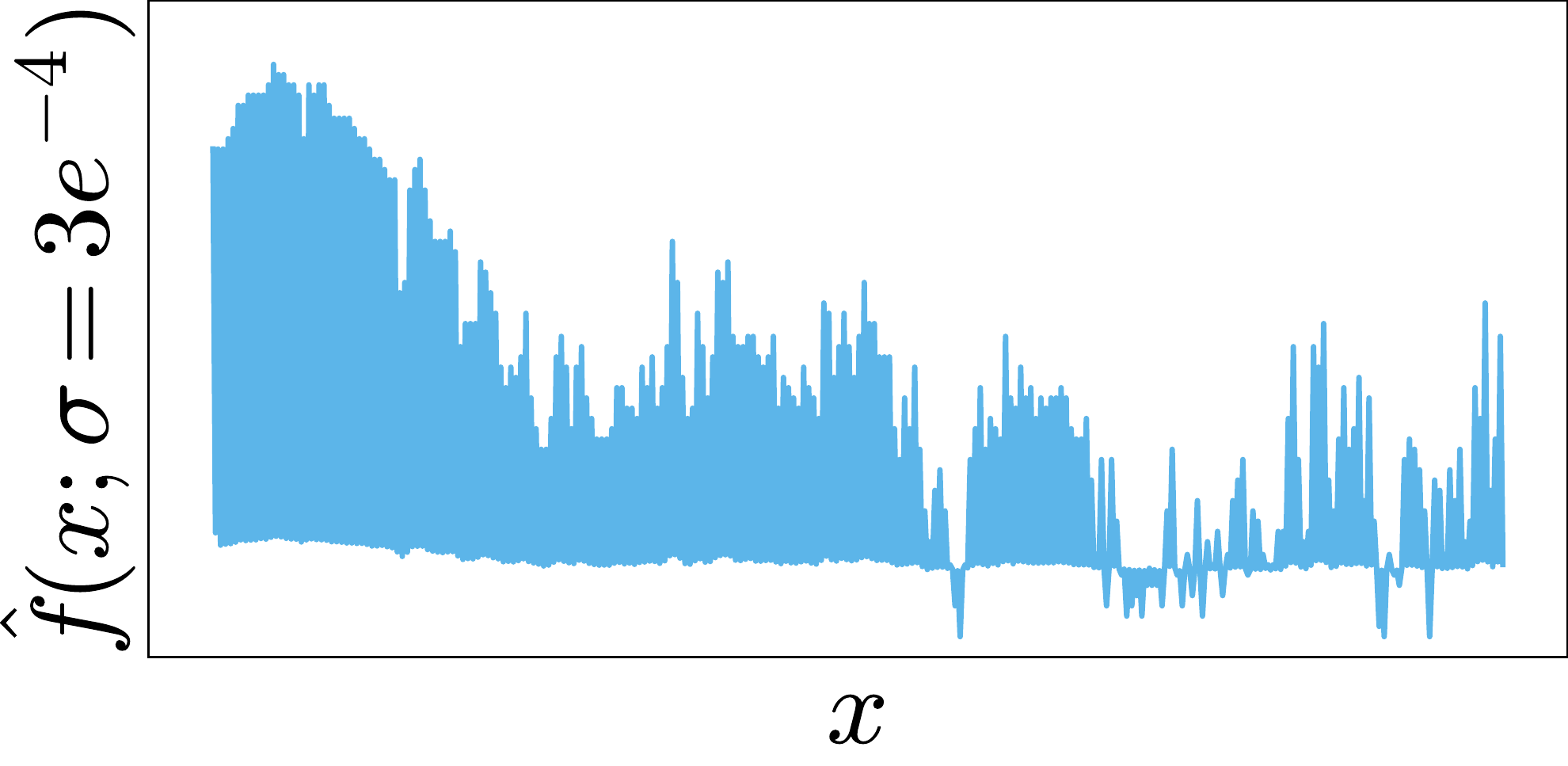} 
\end{subfigure} 
\begin{subfigure}{0.24\textwidth}
\includegraphics[width=\linewidth]{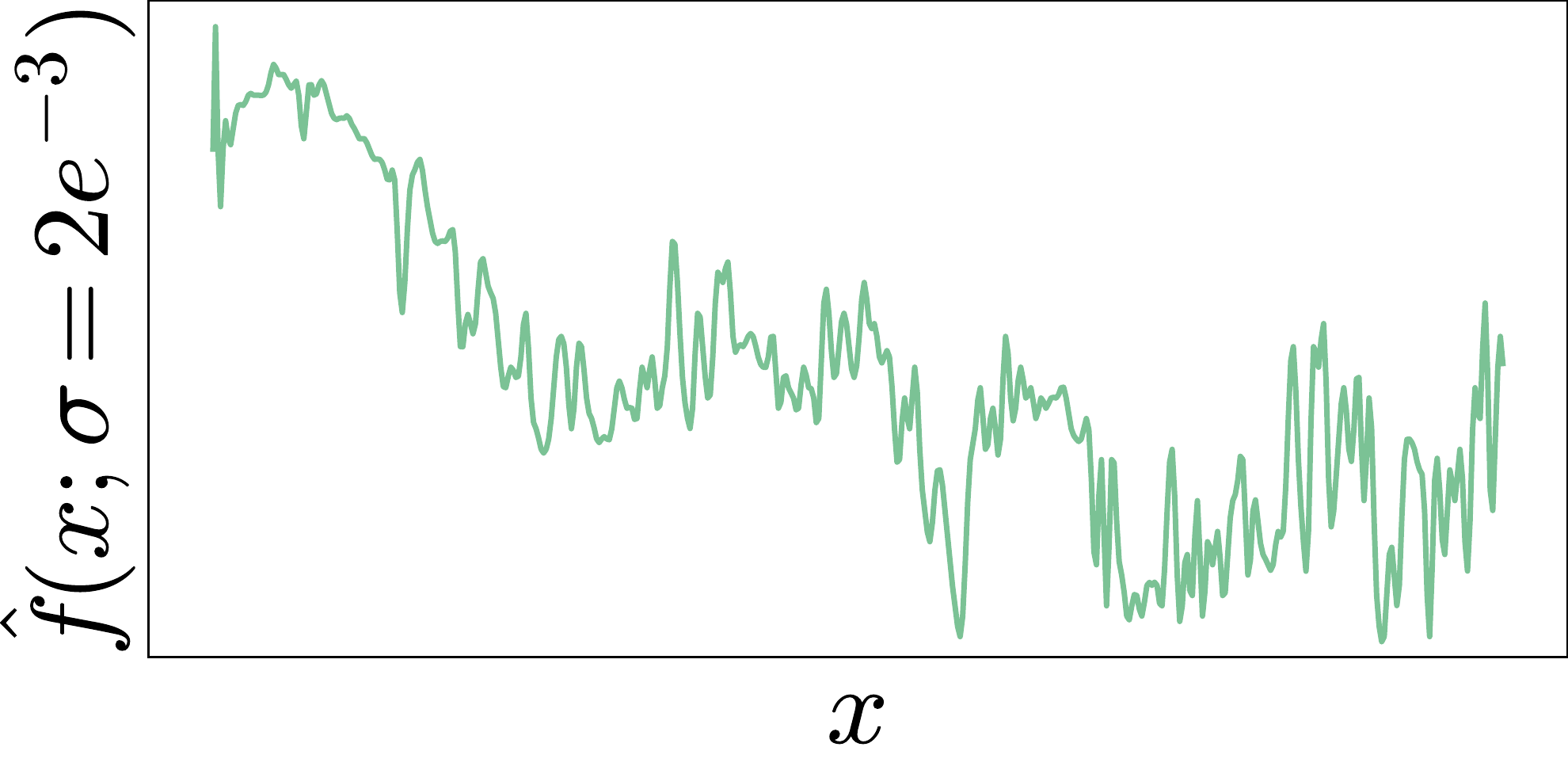} 
\end{subfigure} 
\begin{subfigure}{0.24\textwidth}
\includegraphics[width=\linewidth]{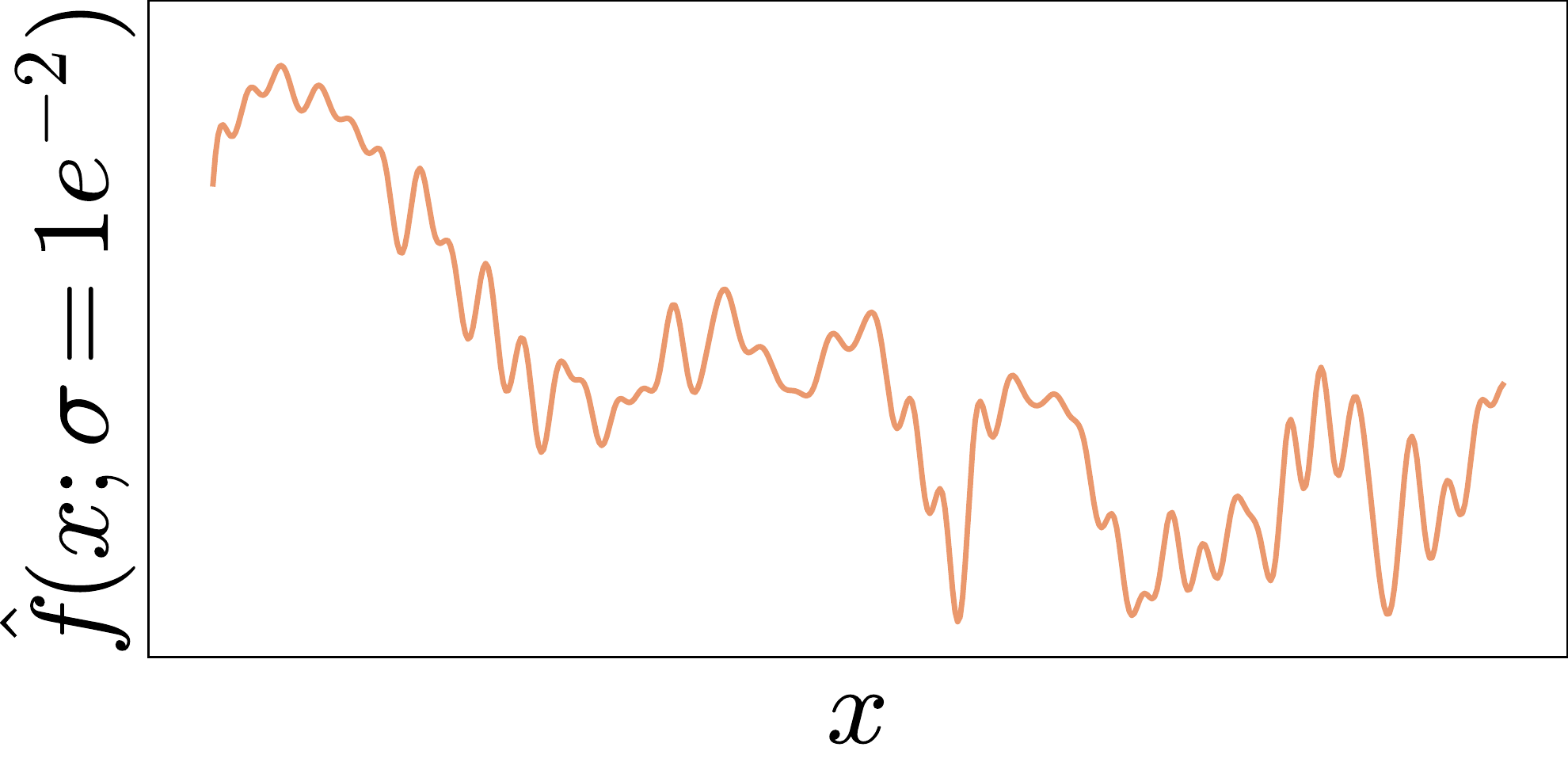} 
\end{subfigure} 

\begin{subfigure}{0.24\textwidth}
\includegraphics[width=\linewidth]{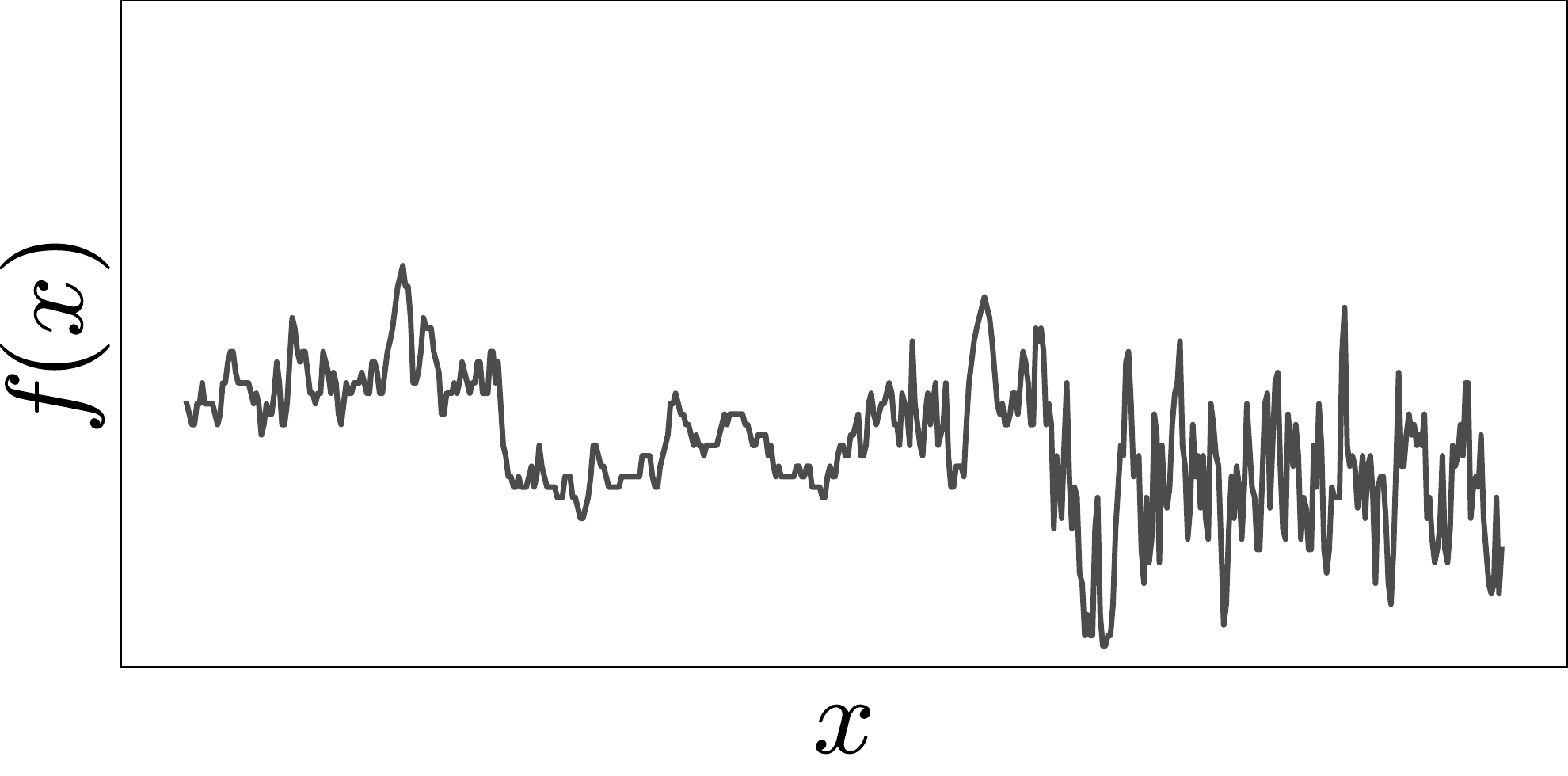} 
\caption{Ground truth}
\end{subfigure} 
\begin{subfigure}{0.24\textwidth}
\includegraphics[width=\linewidth]{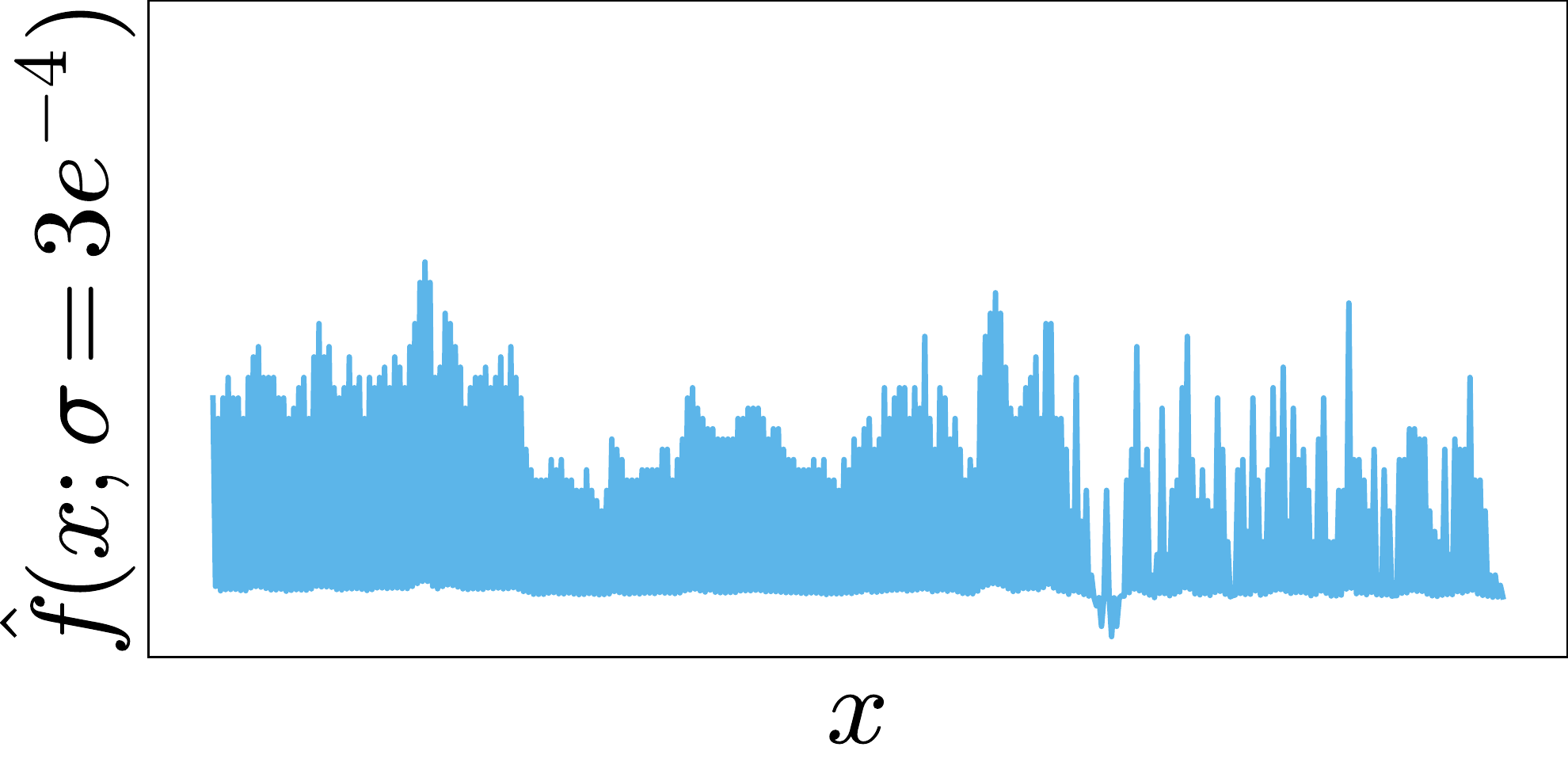} 
\caption{$\sigma = 0.0003$}
\end{subfigure} 
\begin{subfigure}{0.24\textwidth}
\includegraphics[width=\linewidth]{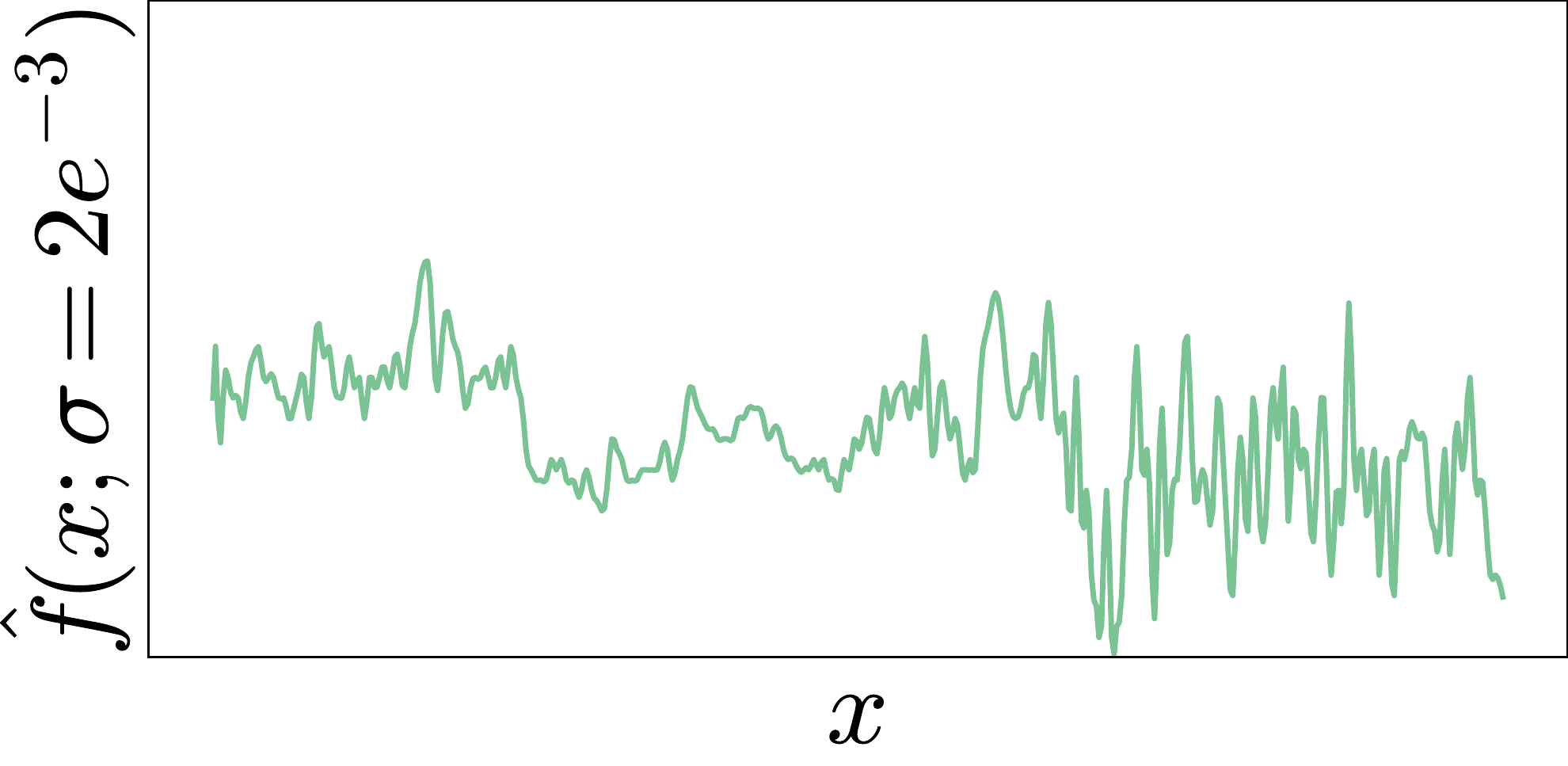} 
\caption{$\sigma = 0.002$}
\end{subfigure} 
\begin{subfigure}{0.24\textwidth}
\includegraphics[width=\linewidth]{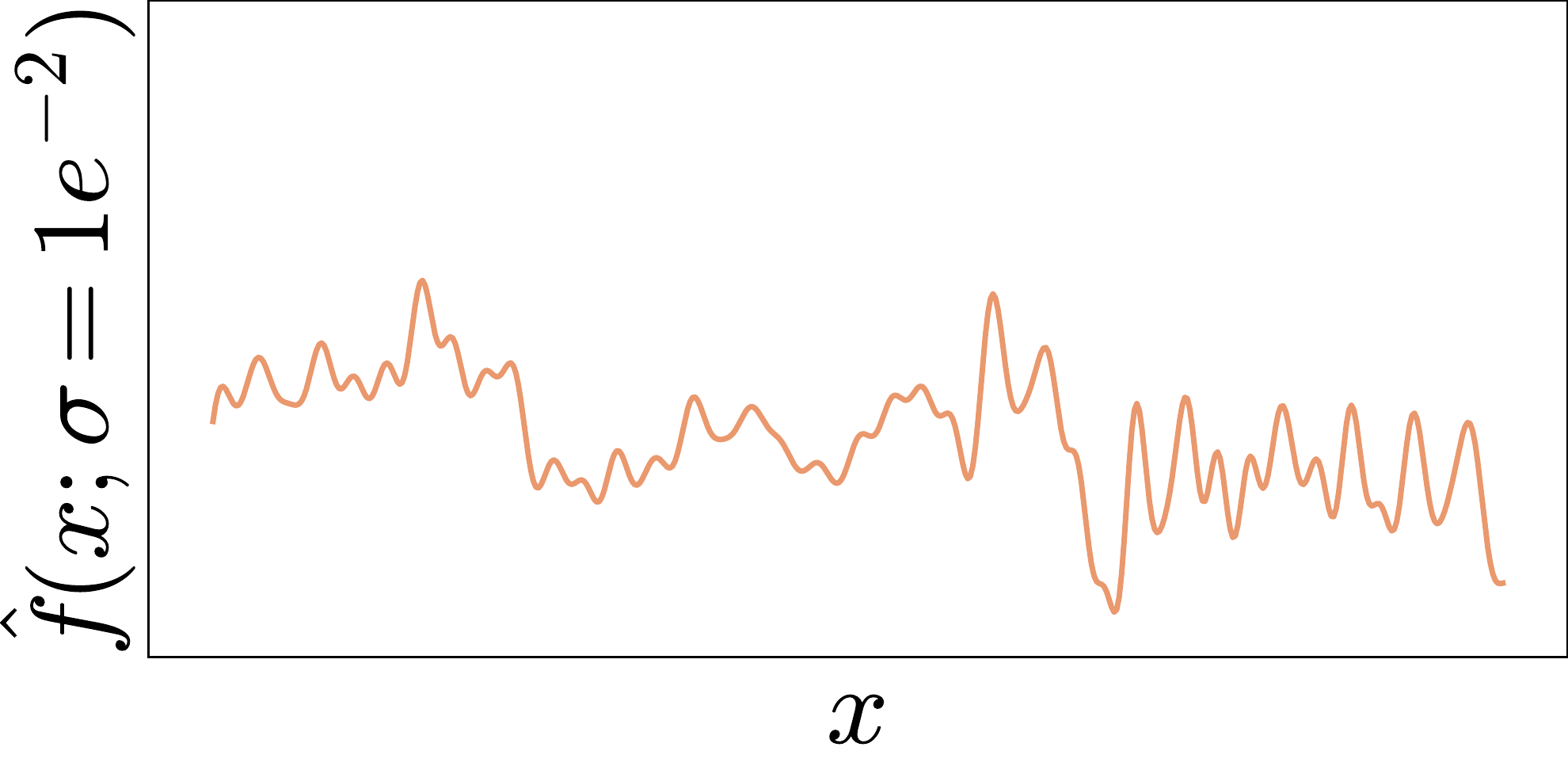} 
\caption{$\sigma = 0.01$}
\end{subfigure} 
\caption{Reconstruction ($\hat{f}(\cdot)$) of 1-D signals  ($f(\cdot)$) using the Gaussian embedder across different standard deviations. A too small $\sigma$ demonstrates poor generalizationd, a   too high $\sigma$ gives over-smooth generalizationd, while a mid-range $\sigma$ produces better results.}
\label{fig:1d_reconstruction}
\end{figure}

We use three random rows from an image to test the reconstruction ability of the Gaussian embedder for a 1-D signal. We use a one-layer linear network and train for $2000$ epochs. The results are shown in Fig.~\ref{fig:1d_reconstruction}. As illustrated, for a very large $\sigma$, the stable rank is too low, leading to poor memorization and overly smooth generalization. When $\sigma$ is too small, the memorization is better, but the generalization is poor. A mid-range $\sigma$ learns the mapping best.  A notable advantage of the Gaussian embedder is that it will output zeros outside the learning range instead of meaningless random numbers.

\subsection{Extension to 2-D signals}

A seemingly critical disadvantage of our embedding mechanism is that when working with higher dimensions, the embedding dimension should grow exponentially (in order to facilitate the dense sampling of embedders in higher dimensions). However, this can be alleviated using separable functions as the embedders. A straightforward example for this is the Gaussian embedder. Recall that high dimensional Gaussians with diagonal covariance matrices are separable along axes.  Therefore, we can employ 1-D embedders along each dimension and then concatenate the embedder outputs for each position.  As a result, the embedding dimension only increases linearly with the dimension of the input signal. However, there is an associated drawback with this method, which we will discuss next.

Consider sampling a 1-D separable embedder along $x$ and $y$ axes separately and concatenating them to obtain the embedding for each $(x,y)$ point of a 2-D signal. Also, denote the ground truth signal as $I(x,y)$. Then, using a linear model we have,


\begin{equation}
    I(x,y) \approx \mathbf{w}_x^T \Psi(\x) + \mathbf{w}_y^T \Psi(\y).
\end{equation}

The above formula can be written in the matrix form as,

\begin{equation}
    \begin{bmatrix}
    I(1,1) & \dots & I(N,1) \\
    \vdots & \ddots & \vdots \\
    I(1,N) & \dots & I(N,N)
    \end{bmatrix} \approx 
    \underbrace{\begin{bmatrix}
    \mathbf{w}_x^T \Psi(x_1) & \dots & \mathbf{w}_x^T \Psi(x_N) \\
    \vdots & \ddots & \vdots \\
    \mathbf{w}_x^T \Psi(x_1) & \dots & \mathbf{w}_x^T \Psi(x_N)
    \end{bmatrix}}_{\begin{matrix} \mathbf{A} \end{matrix}} +
    \underbrace{\begin{bmatrix}
    \mathbf{w}_y^T \Psi(y_1) & \dots & \mathbf{w}_y^T \Psi(y_1) \\
    \vdots & \ddots & \vdots \\
    \mathbf{w}_y^T \Psi(y_N) & \dots & \mathbf{w}_y^T \Psi(y_N)
    \end{bmatrix}}_{\begin{matrix} \mathbf{B} \end{matrix}}
\end{equation}

Clearly, $\mathbf{A}$ and $\mathbf{B}$ are rank $1$ matrices. Therefore, a linear network can only reconstruct an image signal with at most rank $2$. However, this drawback can be addressed in most practical cases using a non-linear MLP with a higher number of layers. Therefore, a vital insight to note here is that the advantage of using deeper and non-linear networks to encode signals becomes more significant as the dimensionality of the input signal increases. 

\begin{figure}[ht]
\captionsetup{size=small}
\centering 
\begin{subfigure}{0.2\textwidth}
\includegraphics[width=\linewidth]{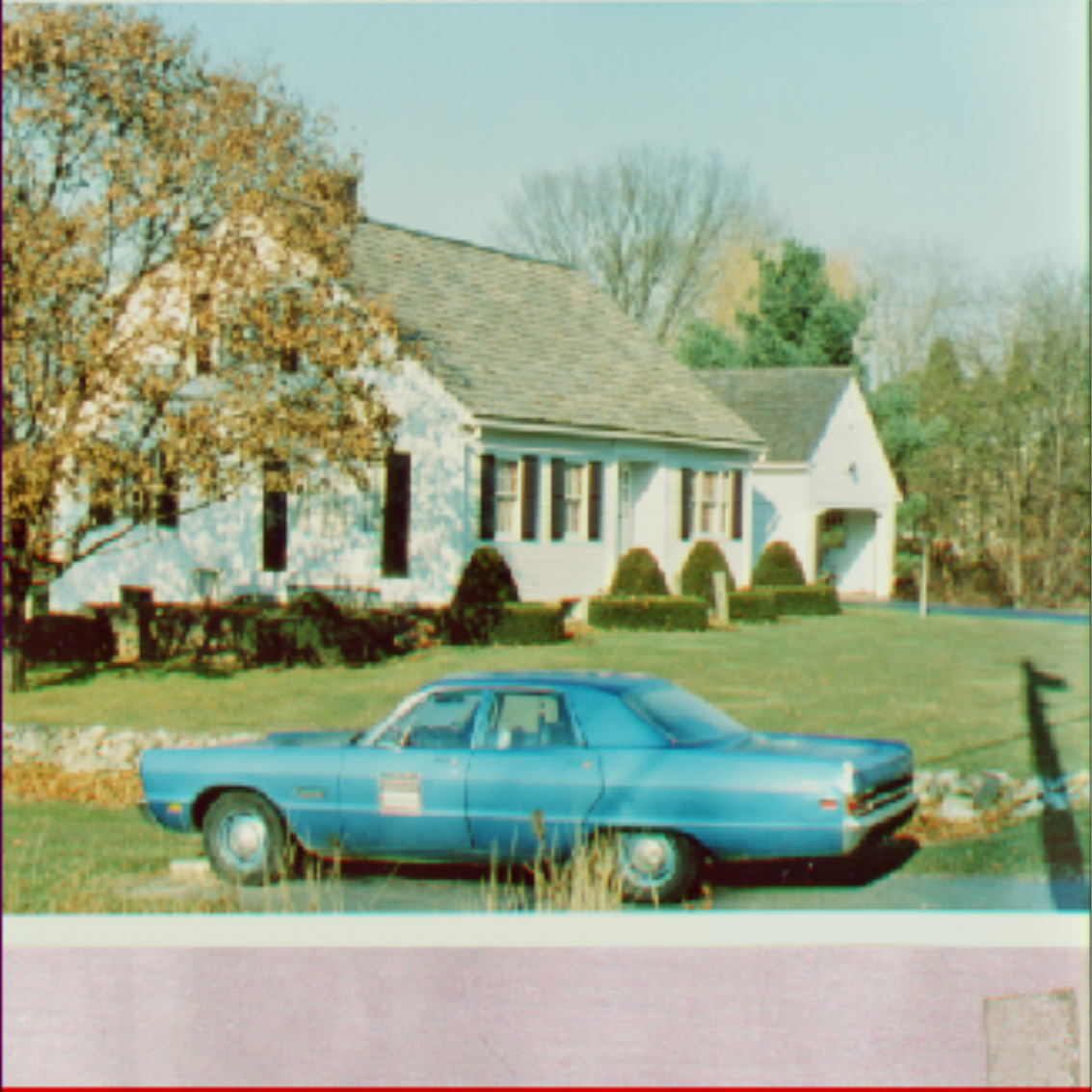} 
\end{subfigure} 
\begin{subfigure}{0.2\textwidth}
\includegraphics[width=\linewidth]{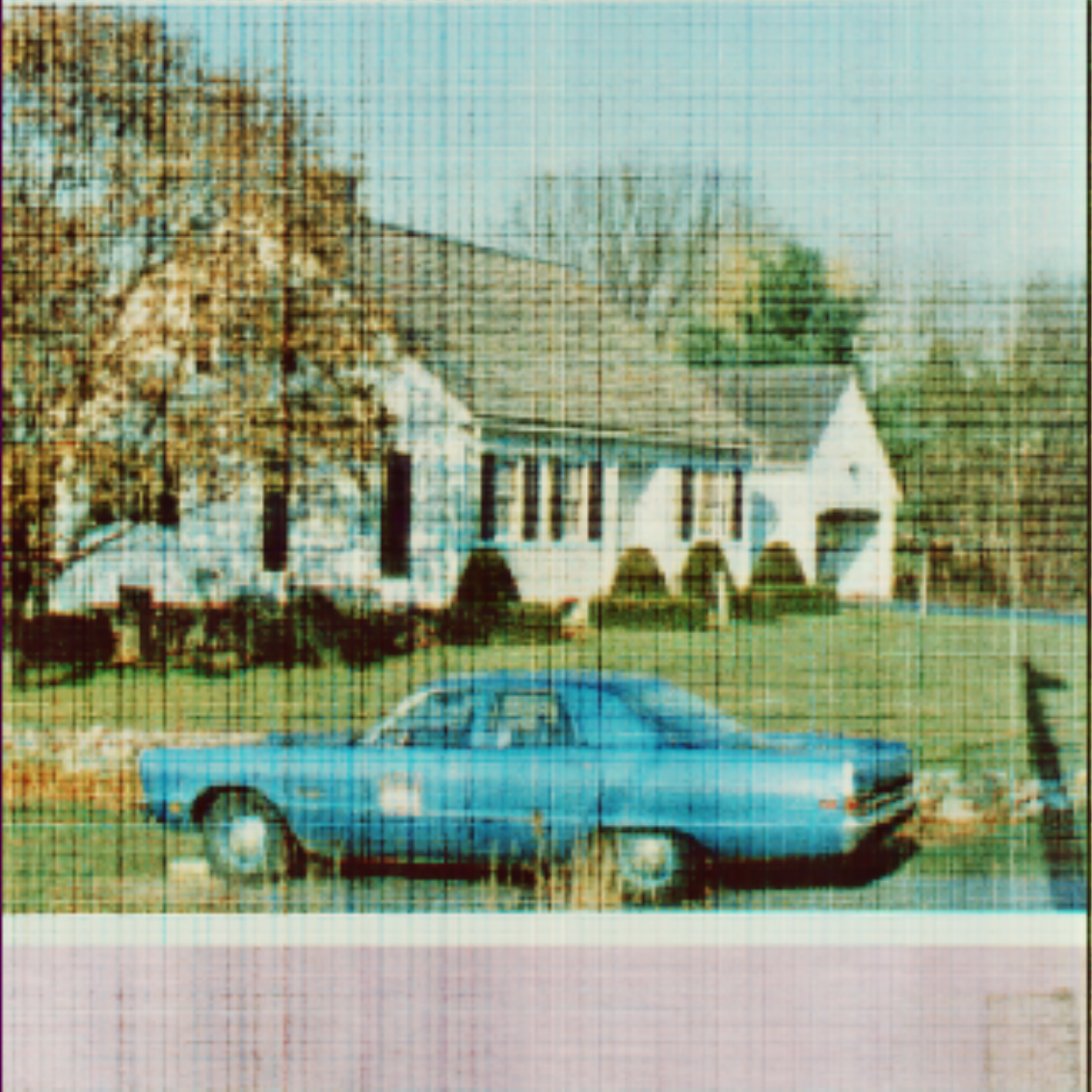} 
\end{subfigure} 
\begin{subfigure}{0.2\textwidth}
\includegraphics[width=\linewidth]{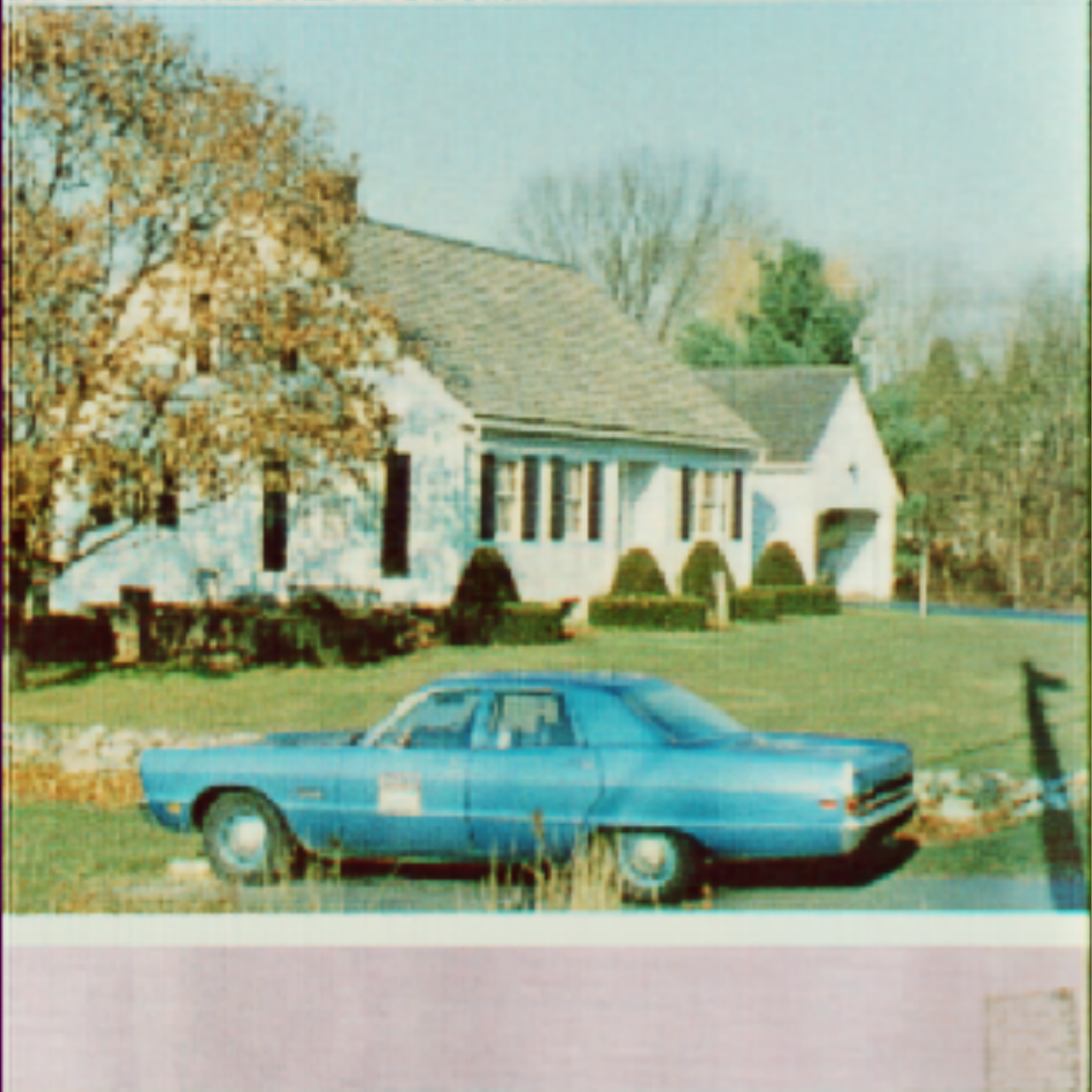} 
\end{subfigure} 
\begin{subfigure}{0.2\textwidth}
\includegraphics[width=\linewidth]{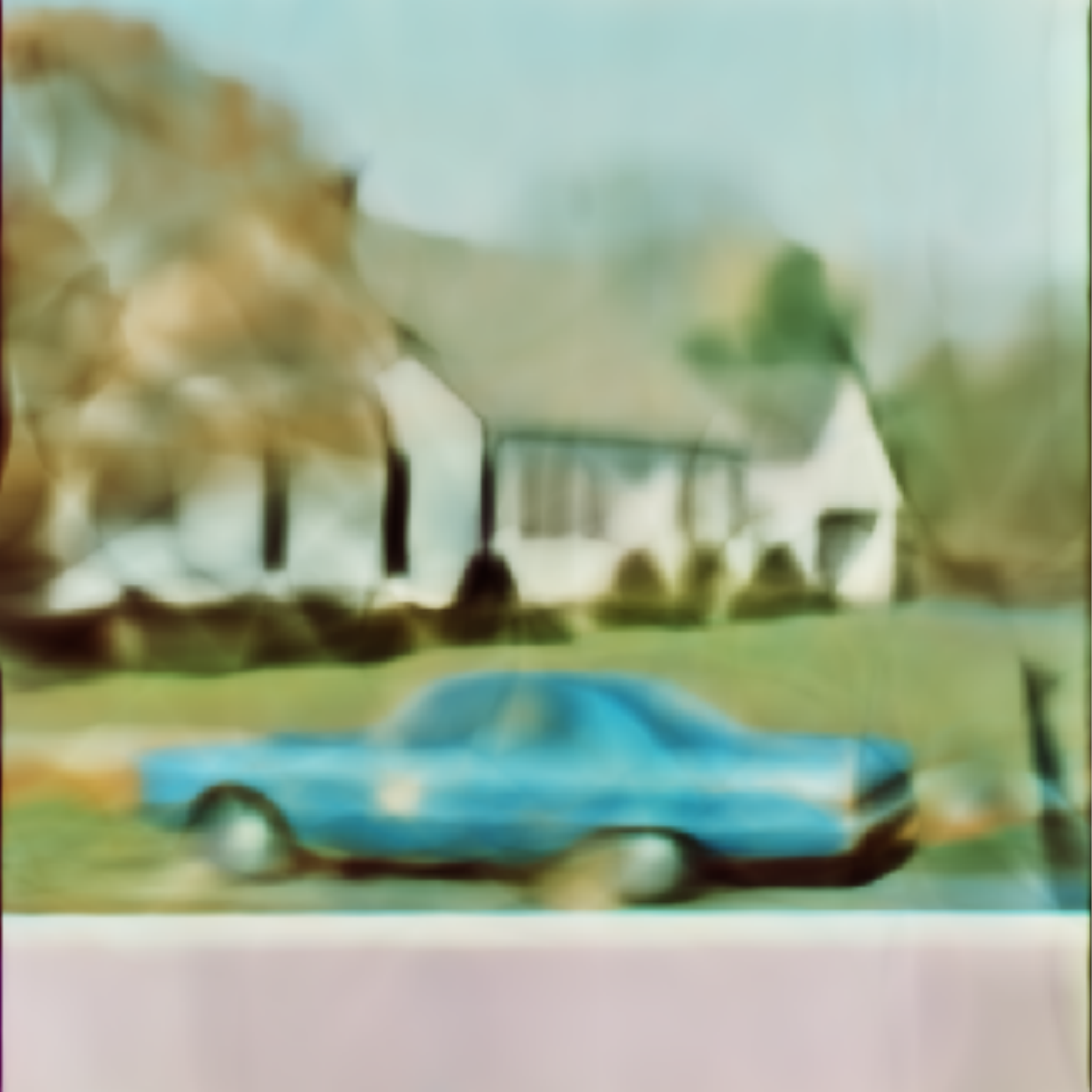} 
\end{subfigure} 

\begin{subfigure}{0.2\textwidth}
\includegraphics[width=\linewidth]{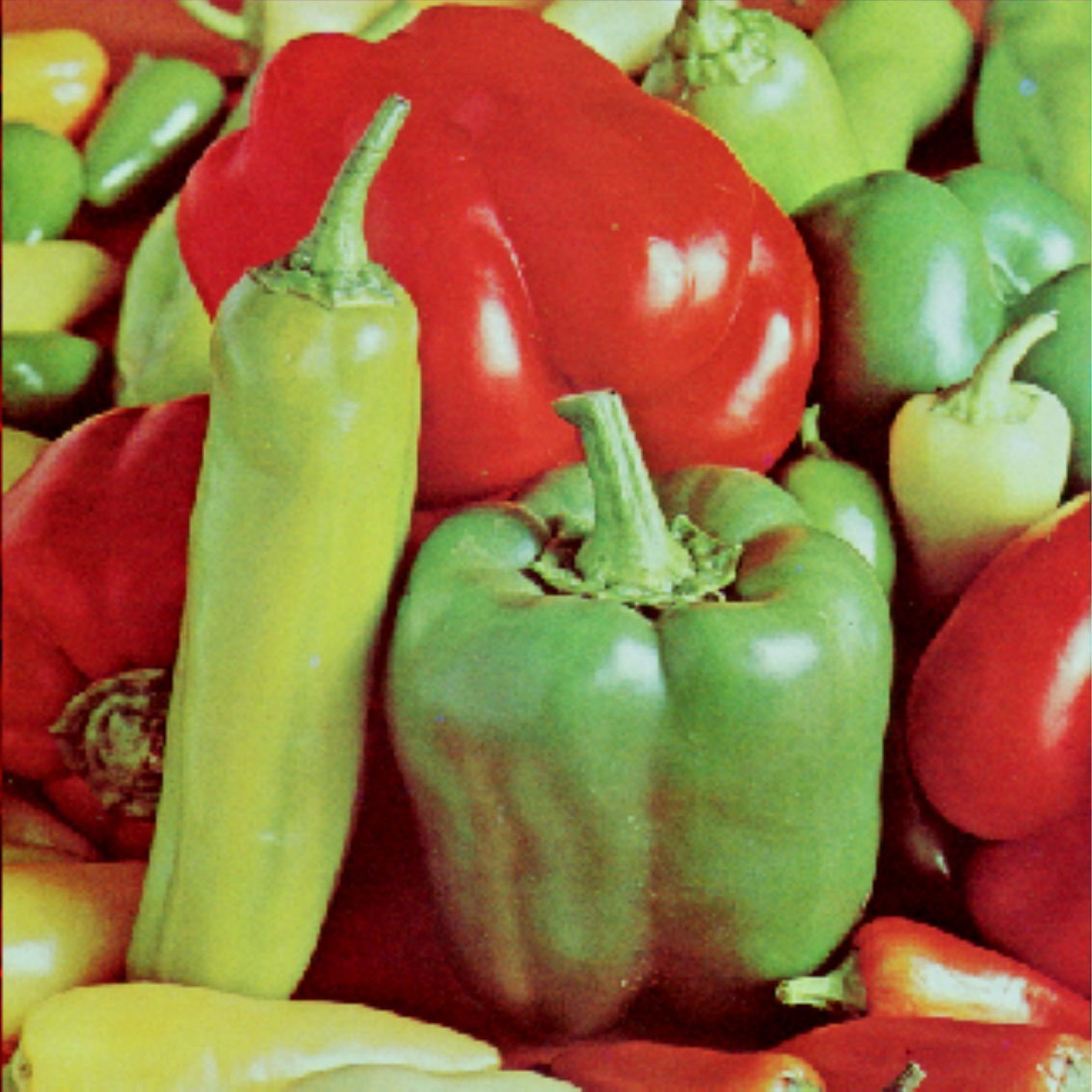} 
\end{subfigure} 
\begin{subfigure}{0.2\textwidth}
\includegraphics[width=\linewidth]{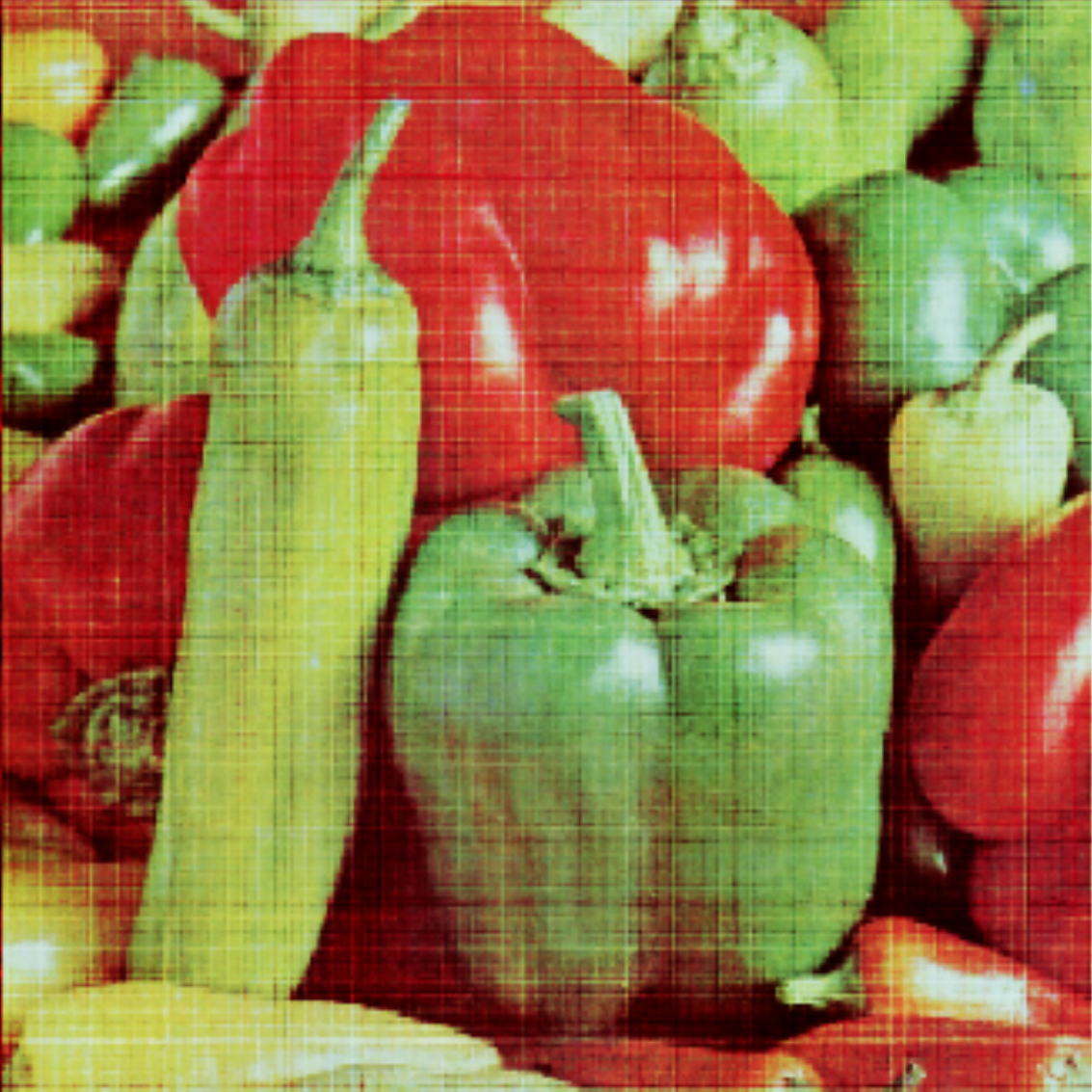} 
\end{subfigure} 
\begin{subfigure}{0.2\textwidth}
\includegraphics[width=\linewidth]{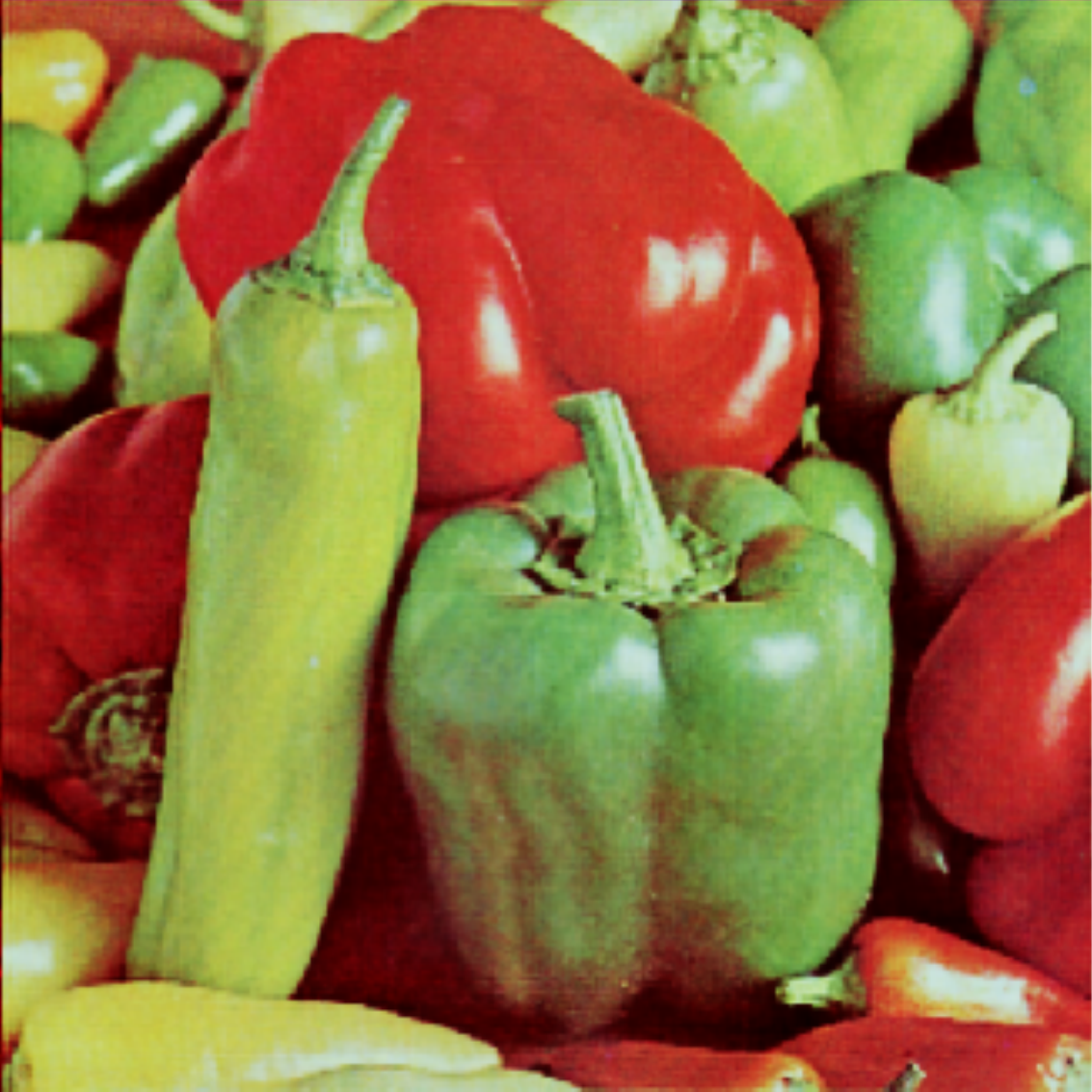} 
\end{subfigure} 
\begin{subfigure}{0.2\textwidth}
\includegraphics[width=\linewidth]{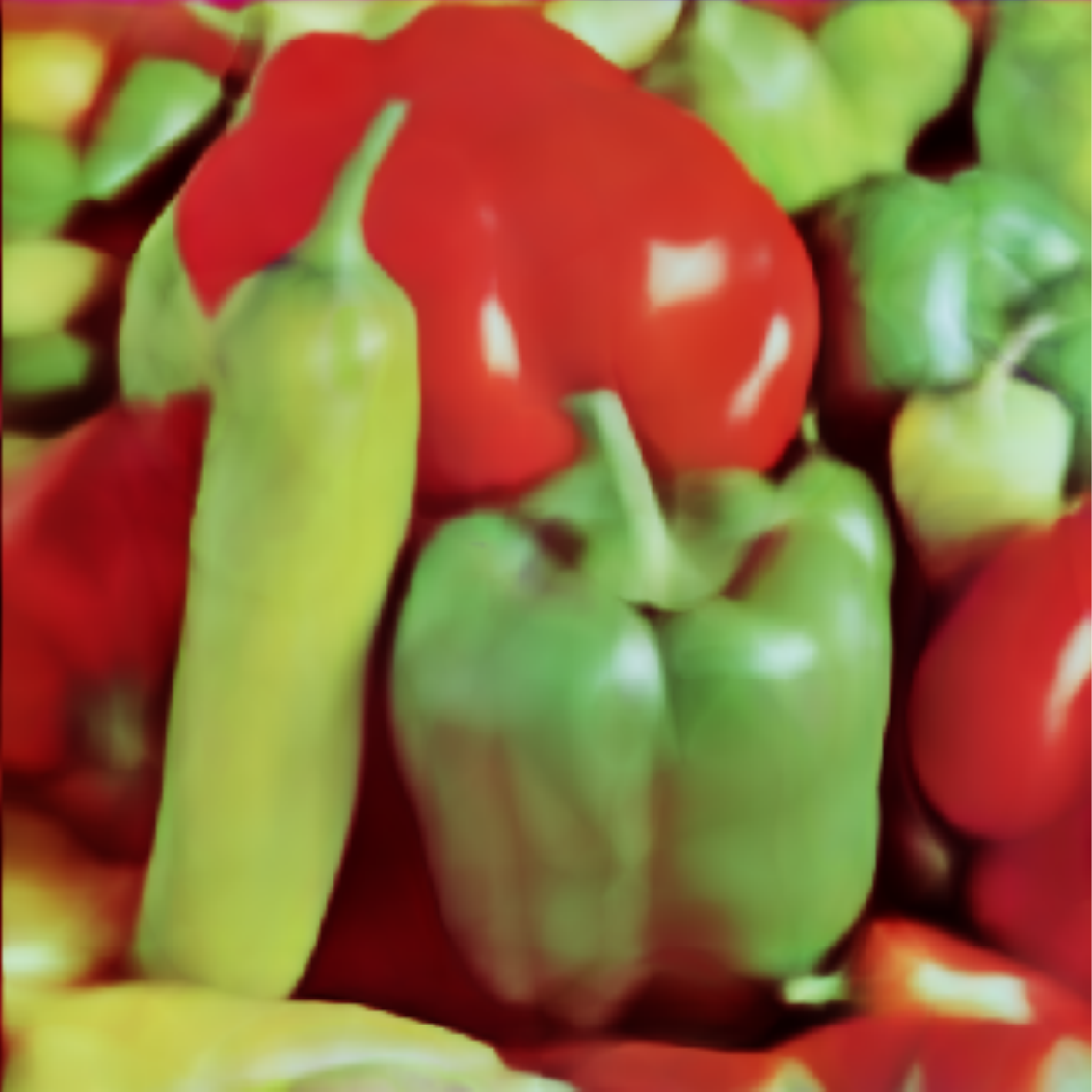} 
\end{subfigure}

\begin{subfigure}{0.2\textwidth}
\includegraphics[width=\linewidth]{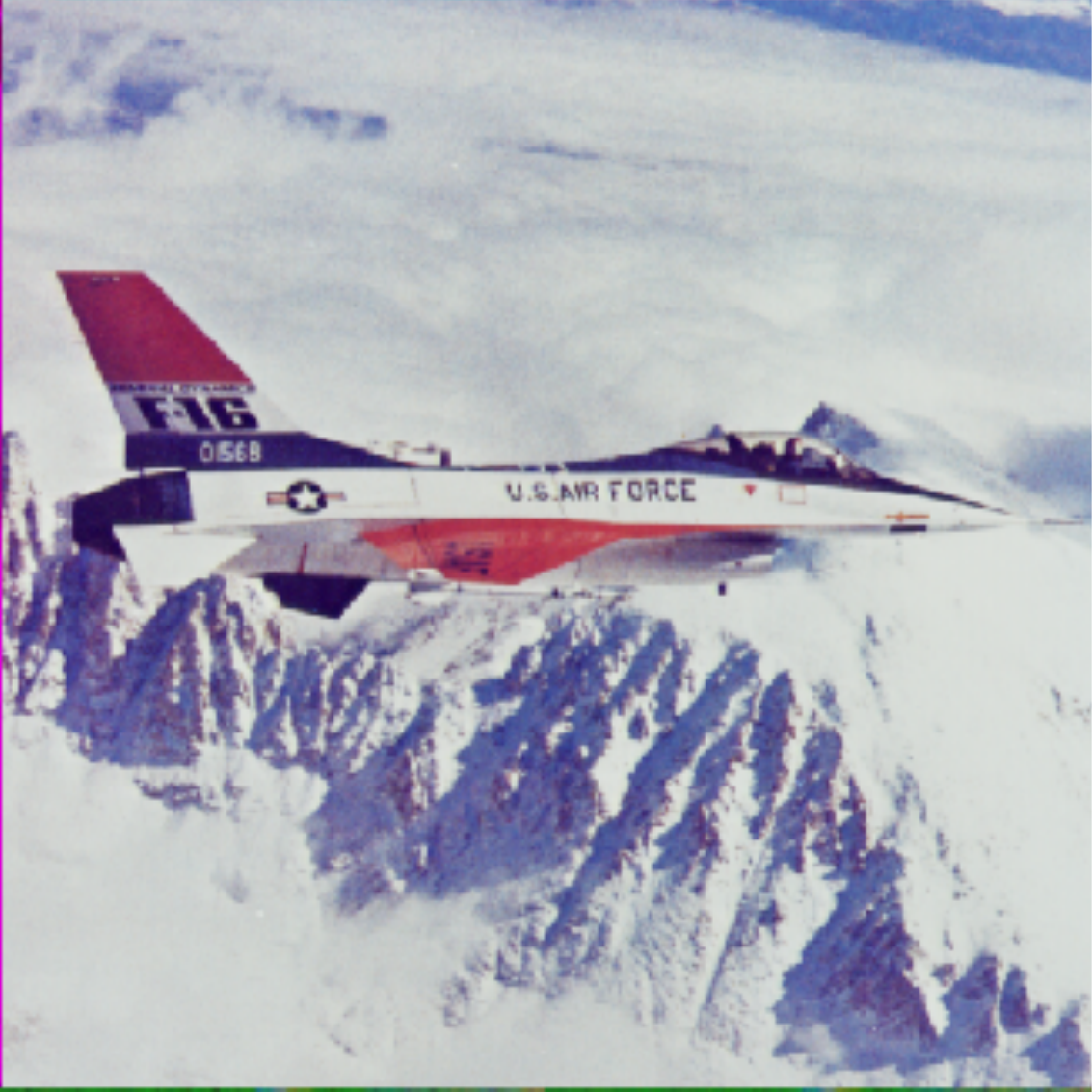} 
\caption{Ground truth}
\end{subfigure} 
\begin{subfigure}{0.2\textwidth}
\includegraphics[width=\linewidth]{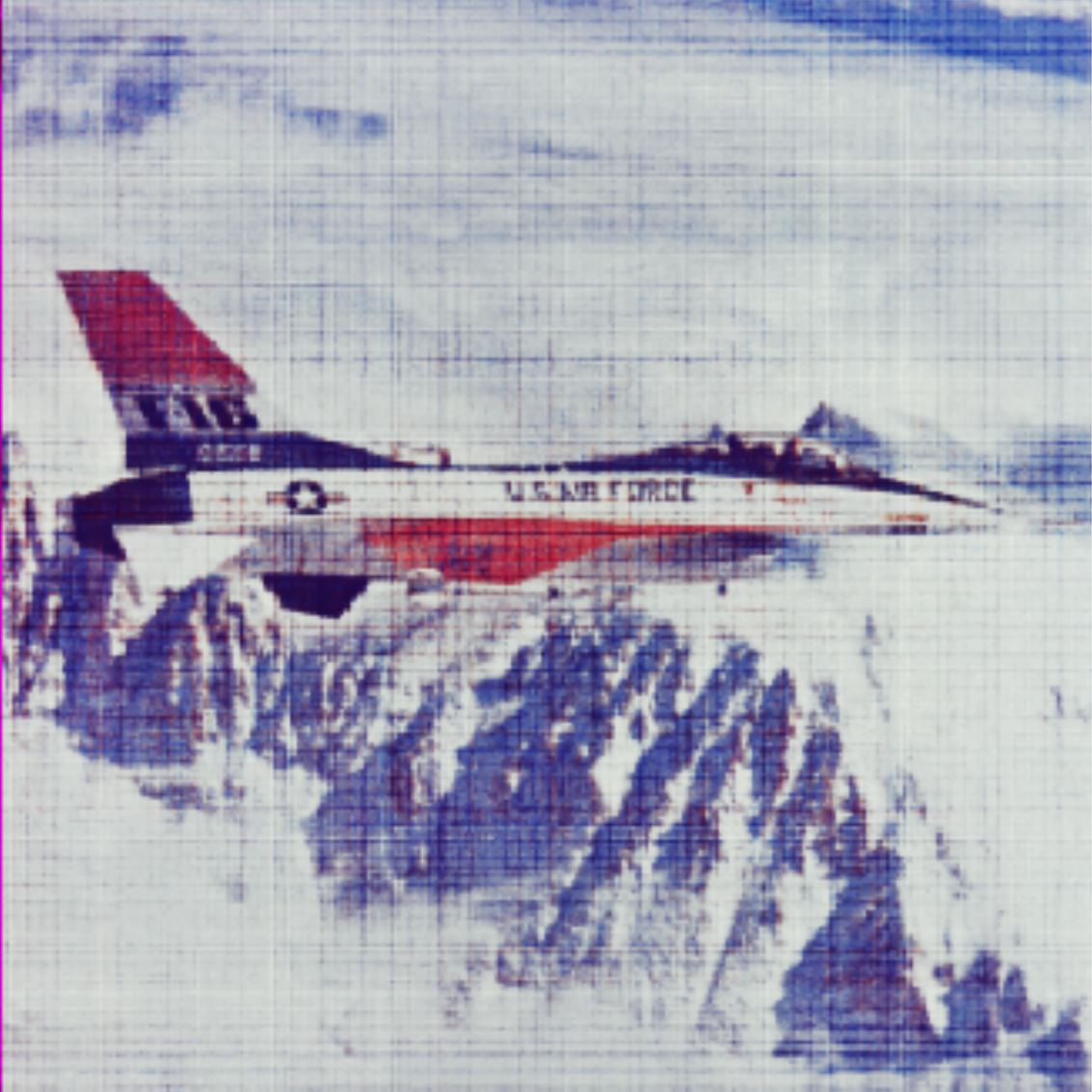} 
\caption{$\sigma = 0.001$}
\end{subfigure} 
\begin{subfigure}{0.2\textwidth}
\includegraphics[width=\linewidth]{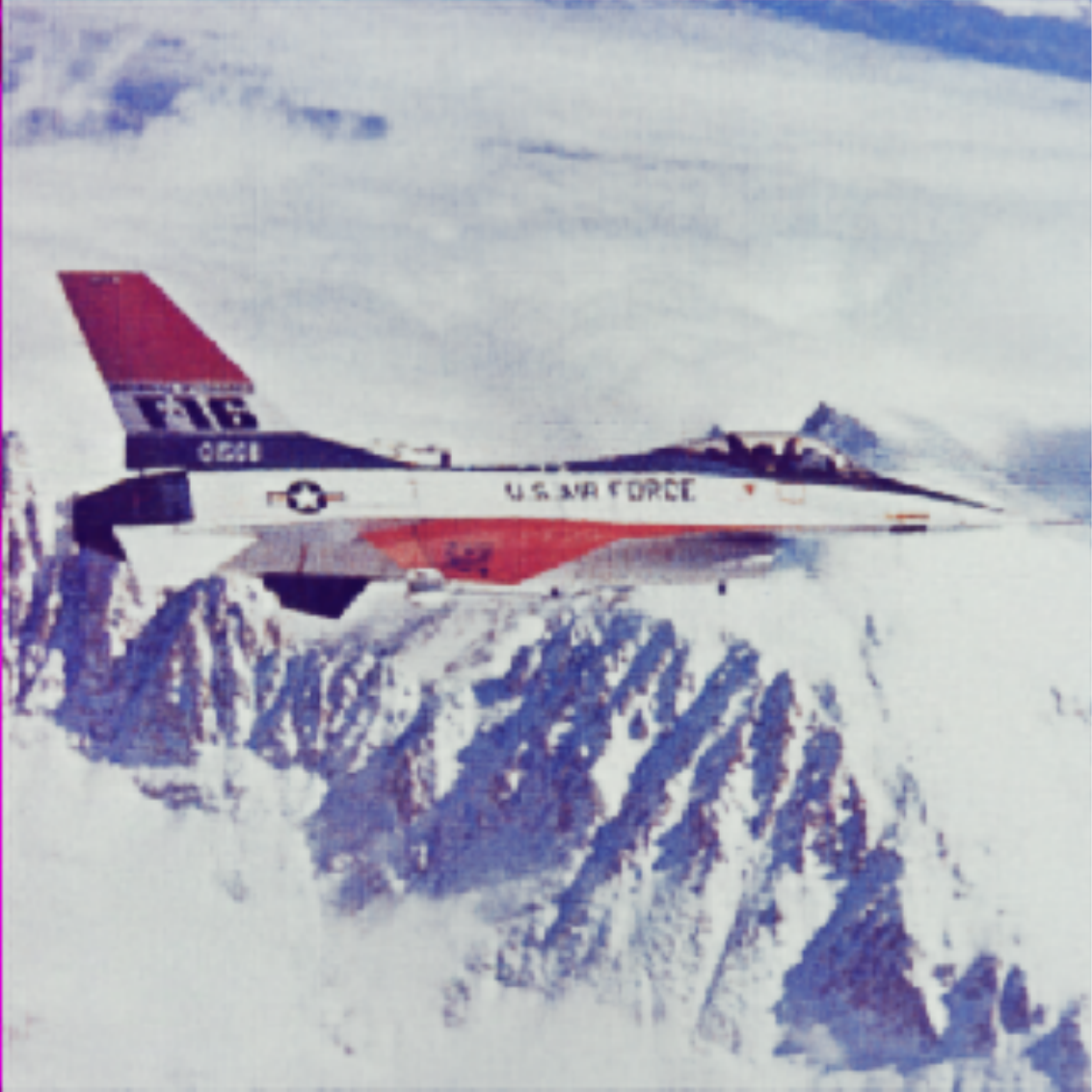} 
\caption{$\sigma = 0.003$}
\end{subfigure} 
\begin{subfigure}{0.2\textwidth}
\includegraphics[width=\linewidth]{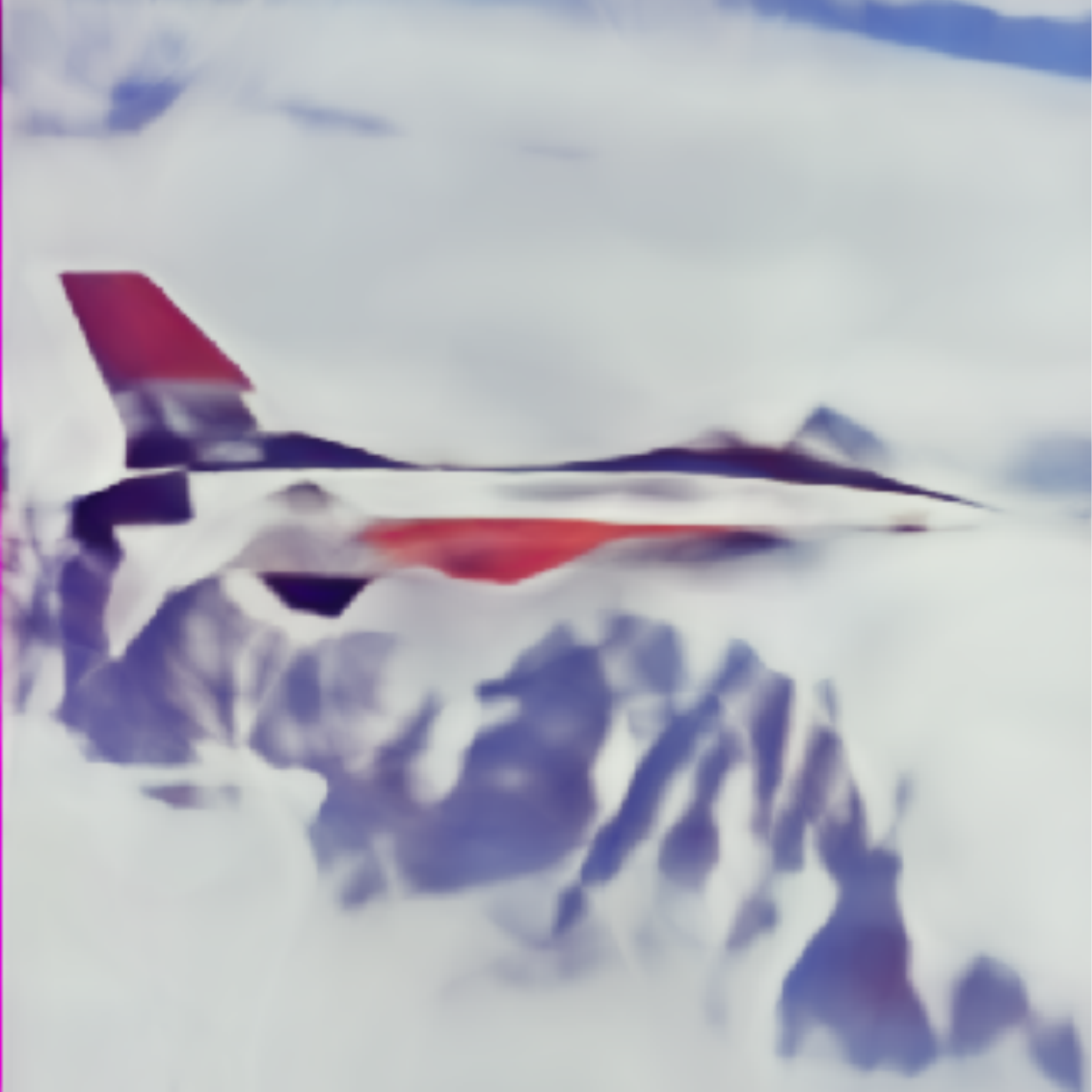} 
\caption{$\sigma = 0.07$}
\end{subfigure} 
\caption{Reconstruction of 2-D signals using the Gaussian embedder across different standard deviations. Here, we only sample along two directions. A too small $\sigma$ demonstrates poor generalization, a   too high $\sigma$ gives over smooth generalization, while a mid-range $\sigma$ produces better results.}
\label{fig:2d_reconstruction}
\end{figure}

\begin{figure}[ht]
\captionsetup{size=small}
\centering 
\begin{subfigure}{0.45\textwidth}
\includegraphics[width=\linewidth]{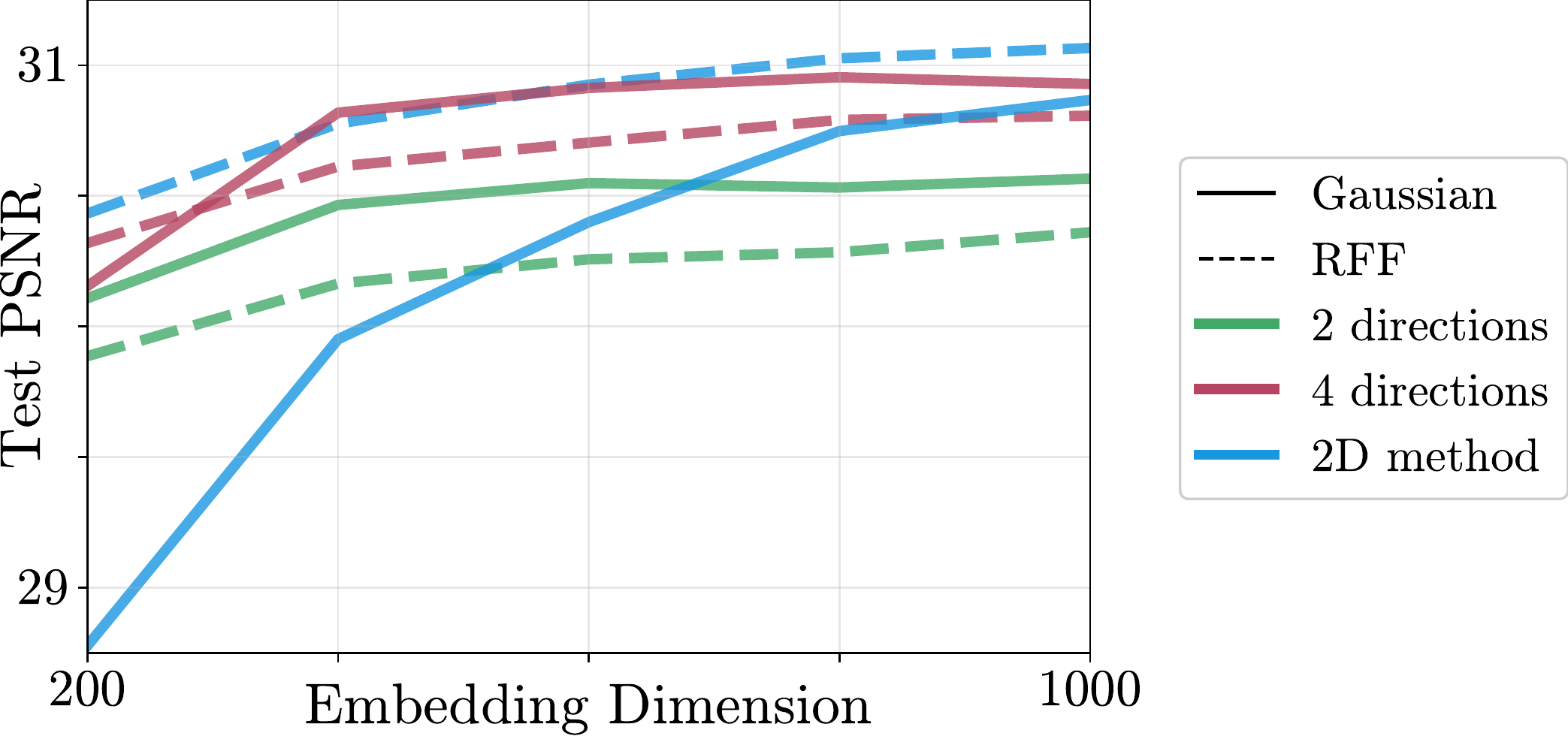} 
\end{subfigure} 
\begin{subfigure}{0.45\textwidth}
\includegraphics[width=\linewidth]{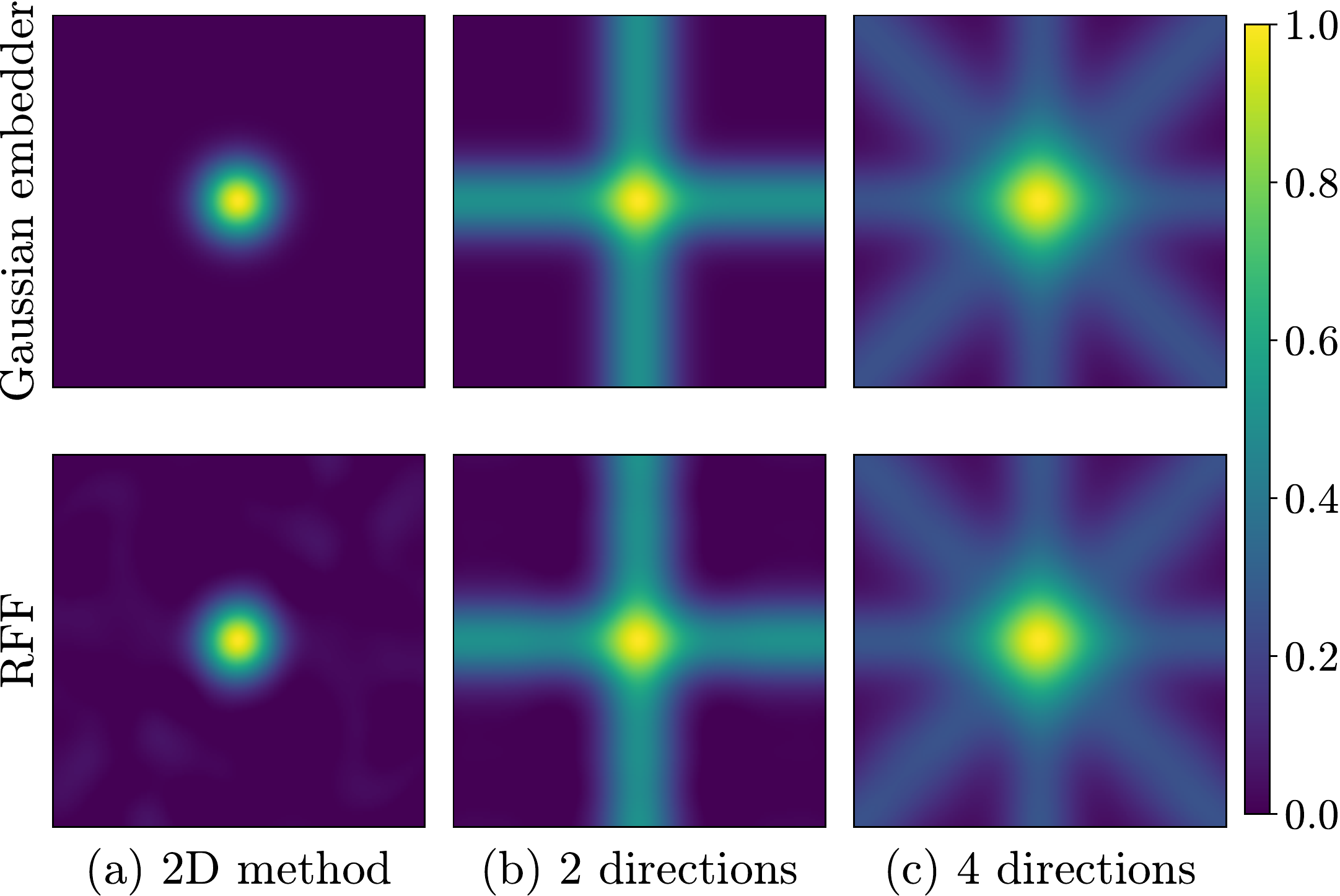} 
\end{subfigure} 
 \caption{Comparison of the test performance and distance preservation across different sampling methods in Gaussian embedding and RFF, when encoding 2-D signals. Note that in the context of RFF, sampling refers to sampling of the frequencies. \textit{Left}:  By sampling along four directions, the Gaussian embedder is able to achieve improved performance. When the embedding dimension is low, sampling in an equidistant grid on the $x-y$ plane (2D method)  results in sparse sampling, leading to sub-par performance. However, as the embedding dimension increases, this method demonstrates superior performance.  \textit{Right}: Distance preservation depends on the sampling method for both RFF and the Gaussian embedder.}%
    \label{fig:perf_gau}%
\end{figure}

An example is shown in Fig.~\ref{fig:2d_reconstruction}. For this experiment, we use a $4$-layer MLP with ReLU activation and only sample along $x$ and $y$ axes separately using a Gaussian embedder to obtain the embedding matrix. We employ $256$ neurons in each layer and train for $2000$ epochs. Similar to 1-D signal reconstruction, using a  mid-range standard deviation for the Gaussian embedder works best. In cases where there are restrictions with the used the MLP, one can always increase the rank of the embedding matrix by sampling the embedder along multiple directions (as opposed to only two directions). Fig. \ref{fig:perf_gau} depicts this. As shown, by sampling along four directions, the Gaussian embedder is able to obtain distance preservation along more directions and on-par results against the RFF embedder. In this experiment, for each embedding dimension and each sampling method in Gaussian embedding, we conduct a linear search for the standard deviation that gives the best test PSNR. For RFF, a constant standard deviation performed equally for all variants. For Gaussian embedder, regardless of the sampling method, the empirical best standard deviation is related to the number of samples along each direction, which is $\sigma = \frac{k}{4d\sqrt{\ln{10}}}$ where $d$ is the embedding dimension and $k=3.5$ from experimental results.

\section{Conclusion}

In this paper, we develop a novel perspective on positional encoding. In summary, we show that the performance of a positional embedding scheme is mainly governed by the stable rank of the embedding matrix and the distance preservation between the embedded coordinates. In light of this discovery, we propose a novel positional encoding mechanism that can incorporate arbitrary continuous signals as potential embedders, under certain constraints. This allows for a more interpretable and less restrictive way to encode positions that can be used in various computer vision tasks.


\bibliographystyle{plain}
\bibliography{main}

\newpage
\appendix

\setcounter{prop}{0}
\section*{\Large Appendix}
\section{Theoretical results}
\label{App:rank}

\begin{prop}
Consider a set of coordinates $\x = [x_1, x_2, \cdots, x_N]^T$, corresponding outputs $\y = [y_1, y_2, \cdots, y_N]^T$, and a $d$ dimensional embedding $\Psi:\mathbb{R} \to \mathbb{R}^d$. Assuming perfect convergence, the necessary and sufficient condition for a linear model to perfect memorize of the mapping between $\x$ and $\y$ is for $\X = [\Psi(x_1) \, \Psi(x_2) \, \dots \, \Psi(x_N)]$ to have full rank.
\end{prop}

\textbf{Proof:} Let us refer to the row vectors of  $\X$  as 
$[\p_{1},\ldots,\p_{d}]^{T}$. In order to perfectly reconstruct $\y$ using a linear learner with weights $\mathbf{w} = [w_1, w_2, \dots, w_d]$ as 
\begin{equation}
    \y = \sum_{i=1}^{d} w_{i} \p_{i} + b,
\end{equation}
one needs~$\X$~to be of rank~$N$ (since $\y$ needs to completely
span $\{ \p_{i} \}_{i=1}^{d}$). If~$d > N$ then there is no unique
solution to~$\{\mathbf{w}, b\}$ without some regularization. In the unlikely
scenario that the row vectors of $\X$ have zero mean, then $\X$ needs
to be of rank $N-1$ since the bias term~$b$ can account for that
missing linear basis. \qed

\begin{prop}
Let the Gaussian embedder be denoted as $\psi(t,x) = \exp (-\frac{\|t-x\|^2}{2\sigma^2})$. With a sufficient embedding dimension, the stable rank of the embedding matrix obtained using the Gaussian embedder is $\min(N, \frac{1}{2\sqrt{\pi}\sigma})$ where $N$ is the number of embedded coordinates. Under the same conditions, the embedded distance between two coordinates $x_1$ and $x_2$ is $\mathrm{D}(x_1,x_2) = \exp (-\frac{\|x_1-x_2\|^2}{4\sigma^2})$.
\end{prop}

\textbf{Proof:} Let us define the Gaussian embedder as $\psi(t,x) = \exp (-\frac{\|t-x\|^2}{2\sigma^2})$, where $\sigma$ is the standard deviation. Given $d$ samples points  $[t_{1},\ldots,t_{d}]$ and $N$ input coordinates $[x_{1},\ldots,x_{N}]$, the elements of the embedding matrix are
\begin{equation}
    \Psi_{i,j} = \psi(t_i,x_j).
\end{equation}

To make sure the stable rank is saturated, we assume that $d$ and $N$ is large enough. Then, $\Psi$ is approximately a circulant matrix. We know that the singular value decomposition of a circulant matrix $C$, whose first row is $c$, can be written as
\begin{equation}
    C =\frac{1}{n}F_n^{-1} diag(F_n c)F_n,
\end{equation}
where $F_n$ is the Fourier transform matrix. This means the singular values of a circulant matrix is the Fourier transform of first row. When $N$ is large enough, we can approximate the first row of $\Psi$ as  a continuous signal, which is $\psi(x, t=0) = \exp (-\frac{\|x\|^2}{2\sigma^2})$, so the singular values are 
\begin{equation}
    s(\xi) =\mathcal{F}(\psi(x;t=0)) = \sqrt{2\pi}\sigma\exp(-2\sigma^2\|\pi\xi\|^2).
\end{equation}
Therefore, we can calculate stable rank directly from the definition,
\begin{equation}
    \textrm{Stable Rank}(\Psi) = \sum_{i=1}^N{\frac{s_i^2}{s_1^2}} = \int_{-\infty}^{+\infty}\frac{s(\xi)}{s(0)}^2d\xi=\int_{-\infty}^{+\infty}\exp(-4\sigma^2\|\pi\xi\|^2)d\xi=\frac{1}{2\sqrt{\pi}\sigma}.
\end{equation}
Considering the general case, where $N$ might not be large enough, the stable rank will be $\min(N, \frac{1}{2\sqrt{\pi}\sigma})$.

The distance (or similarity) between two embedded coordinates can be obtained via the inner product:
\begin{equation}
\begin{aligned}
    \mathrm{D}(x_1,x_2) &= \int_{-\infty}^{+\infty} \psi(t,x_1)\psi(t,x_2)dt
    \\ &=\int_{-\infty}^{+\infty} e^{-\frac{(t-x_1)^2}{2\sigma^2}}e^{-\frac{(t-x_2)^2}{2\sigma^2}}dt
    \\ &=\int_{-\infty}^{+\infty} e^{-\frac{(t-x_1)^2+(t-x_2)^2}{2\sigma^2}}dt
    \\ &=\int_{-\infty}^{+\infty} e^{-\frac{t^2-2x_1t+x_1^2+t^2-2x_2t+x_2^2}{2\sigma^2}}dt
    \\ &=\int_{-\infty}^{+\infty} e^{-\frac{2t^2-2(x_1+x_2)t+\frac{(x_1+x_2)^2}{2}+\frac{(x_1-x_2)^2}{2}}{2\sigma^2}}dt
    \\ &=\int_{-\infty}^{+\infty} e^{-\frac{(t-\frac{x_1+x_2}{2})^2}{\sigma^2}}e^{-\frac{(x_1-x_2)^2}{4\sigma^2}}dt
    \\ &=e^{-\frac{(x_1-x_2)^2}{4\sigma^2}}\int_{-\infty}^{+\infty} e^{-\frac{(t-\frac{x_1+x_2}{2})^2}{\sigma^2}}dt
    \\ &=\sqrt{\pi}\sigma e^{-\frac{(x_1-x_2)^2}{4\sigma^2}}.
\end{aligned}
\end{equation}
which is also a Gaussian with a standard deviation of  $\sqrt{2}\sigma$. We can empirically define that the distance between two embedded coordinates $x_1$ and $x_2$ is preserved if $D(x_1,x_2)\ge 10^{-k}$, for an interval $x_1-x_2\leq l$, where $k$ is a threshold. In the Gaussian embedder, we can analytically obtain a $\sigma$ for an arbitrary $l$ using the relationship  $\sigma = \frac{l}{2\sqrt{k\ln{10}}}$. \qed

\begin{prop}
Let the RFF embedding be denoted as $\gamma(x)=[\cos{\mathbf{b}x},\sin{\mathbf{b}x}]$, where $\mathbf{b}$ are sampled from a Gaussian distribution. When the embedding dimension is large enough, the stable rank of RFF will be $\min(N, \sqrt{2\pi}\sigma)$, where $N$ is the numnber of embedded coordinates. Under the same conditions, the embedded distance between two coordinates $x_1$ and $x_2$ is $\mathrm{D}(x_1,x_2) = \sum_j \cos{b_j(x_1-x_2)}$.
\end{prop}

\textbf{Proof:} Given $\frac{d}{2}$ samples for $\mathbf{b}$ as $[b_{1},\ldots,b_{\frac{d}{2}}]$ from a Gaussian distribution with a standard deviation $\sigma$ and $N$ input coordinates $[x_{1},\ldots,x_{N}]$,  RFF embedding is defined as $\gamma(x)=[\cos{2\pi\mathbf{b}x_i},\sin{2\pi\mathbf{b}x_i}]$. 

To make sure the stable rank is saturated, we assume that the $d$ and $N$ is large enough. Although RFF embedding matrix is not circulant, it is naturally frequency based so we already know its spectrum, which is its singular value distribution
\begin{equation}
    s(\xi) = \frac{1}{\sqrt{2\pi}\sigma}\exp(-\frac{\xi^2}{2\sigma^2}).
\end{equation}
Similarly,
\begin{equation}
    \textrm{Stable Rank}(\gamma) = \sum_{i=1}^N{\frac{s_i^2}{s_1^2}} = \int_{-\infty}^{+\infty}\frac{s(\xi)}{s(0)}^2d\xi=\int_{-\infty}^{+\infty}\exp(-\frac{\xi^2}{2\sigma^2})d\xi=\sqrt{2\pi}\sigma,
\end{equation}
Considering the general case, the stable rank is $\min(N, \sqrt{2\pi}\sigma)$.

From basic trigonometry, it can be easily deduced that $\mathrm{D}(x_1,x_2) = \sum_j \cos{2\pi b_j(x_1-x_2)}$. When $d$ is extremely large it can be considered as $f(\xi) = \cos{2\pi\xi(x_1-x_2)}$ where $\xi$ is a Gaussian random variable with standard deviation $\sigma$. Then the above sum can be replaced with the integral,
\begin{equation}
\begin{aligned}
    \mathrm{D}(x_1,x_2) &= \int_{-\infty}^{+\infty} e^{-\frac{\xi^2}{2\sigma^2}}\cos{2\pi\xi(x_1-x_2)}d\xi
    \\ &= 2\int_{0}^{+\infty} e^{-\frac{\xi^2}{2\sigma^2}}\cos{2\pi\xi(x_1-x_2)}d\xi
    \\ &= 2\int_{0}^{+\infty} e^{-\frac{\xi^2}{2\sigma^2}}\frac{1}{2}(e^{i2\pi(x_1-x_2)\xi}+e^{-i2\pi(x_1-x_2)\xi})d\xi
    \\ &= \int_{0}^{+\infty} e^{-\frac{\xi^2}{2\sigma^2}+i2\pi(x_1-x_2)\xi}+e^{-\frac{\xi^2}{2\sigma^2}-i2\pi(x_1-x_2)\xi}d\xi.
\end{aligned}
\end{equation}
Further,
\begin{equation}
    \int_{0}^{+\infty} e^{-ax^2+bx}dx= e^{-\frac{b^2}{4a}}\int_{0}^{+\infty} e^{-a(x-i\frac{b}{2a})^2}dx=\frac{1}{2}(1+\mathrm{erfi}(\frac{b}{2\sqrt{a}}))\sqrt{\frac{\pi}{a}}e^{-\frac{b^2}{4a}}.
\end{equation}
Let $a = \frac{1}{2\sigma^2}$ and $b=\pm2\pi(x_1-x_2)$. Then, we have
\begin{equation}
    \mathrm{D}(x_1,x_2)=\sqrt{2\pi}\sigma e^{-2\pi^2\sigma^2(x_1-x_2)^2}.
\end{equation} \qed

\end{document}